\documentclass{article}

\usepackage{microtype}
\usepackage{graphicx}
\usepackage{graphicx}
\usepackage{subcaption}
\usepackage{booktabs} % for professional tables

\usepackage{hyperref}

\usepackage[accepted]{icml2025}
\usepackage{amsmath}
\usepackage{amssymb}
\usepackage{mathtools}
\usepackage{amsthm}

\usepackage[capitalize,noabbrev]{cleveref}

\theoremstyle{plain}

\theoremstyle{definition}

\theoremstyle{remark}

\usepackage[textsize=tiny]{todonotes}

\usepackage{float}
\crefformat{section}{\S#2#1#3}
\crefformat{subsection}{\S#2#1#3}
\crefformat{subsubsection}{\S#2#1#3}
\crefrangeformat{section}{\S\S#3#1#4 to~#5#2#6}
\crefmultiformat{section}{\S\S#2#1#3}{ and~#2#1#3}{, #2#1#3}{ and~#2#1#3}

\usepackage{url}
\usepackage{booktabs}       % professional-quality tables
\usepackage{amsfonts}       % blackboard math symbols
\usepackage{nicefrac}       % compact symbols for 1/2, etc.
\usepackage{microtype}      % microtypography
\usepackage{xcolor}         % colors
\usepackage{enumitem,kantlipsum}
\crefname{section}{$\mathsection$}{$\mathsection\mathsection$}
\Crefname{section}{$\mathsection$}{$\mathsection\mathsection$}
\usepackage{pifont}
\usepackage{xspace}
\usepackage{enumitem}
\usepackage{wrapfig}
\usepackage{subcaption}
\usepackage{enumitem,kantlipsum}
\usepackage{multirow}
\usepackage{tikz}
\usepackage{longtable}
\usepackage{array}
\newcommand{\outputVar}{\mathrm{Y}}
\newcommand{\outputval}{y}

\newcommand{\inputval}{x}
\newcommand{\circone}{\ding{172}\xspace}
\newcommand{\circtwo}{\ding{173}\xspace}
\newcommand{\circthree}{\ding{174}\xspace}
\newcommand{\circfour}{\ding{175}\xspace}
\newcommand{\circfive}{\ding{176}\xspace}
\usepackage{graphicx}

\usepackage{xspace}
\usepackage{caption}
\usepackage{wrapfig}
\usepackage[percent]{overpic}
\usepackage{float}
\usepackage{booktabs} 
\usepackage{collcell}
\newcommand{\mytexttt}[1]{\raggedright\texttt{#1}}
\newcolumntype{M}[1]{>{\collectcell\mytexttt}p{#1}<{\endcollectcell}}
\usepackage{inconsolata}
\usepackage[most]{tcolorbox}
\usepackage{listings}
\usepackage{fontawesome5} 
\usepackage[svgnames]{xcolor}
\usepackage{hyperref}

\usepackage{subcaption}

\usepackage{algorithm}
\usepackage{algorithmic}

\usepackage{xcolor}
\usepackage{longtable}

\newcommand{\eg}{e.g.,}
\newcommand{\ie}{i.e.,}
\newcommand{\etc}{etc.}

\newcommand\appref[1]{Appendix~\ref{#1}}
\newcommand{\fakeparagraph}[1]{\vspace{1mm}\noindent\textbf{#1}}

\newcommand{\name}{\textsc{BACo}}
\newcommand{\problem}{diversity-quality trade-off}

\usepackage{amsmath,amsfonts,bm}

\def\Figref#1{Figure~\ref{#1}}

\def\Tabref#1{Table~\ref{#1}}

\def\Secref#1{Section~\ref{#1}}
\def\eqref#1{equation~\ref{#1}}
\def\1{\bm{1}}

\DeclareMathAlphabet{\mathsfit}{\encodingdefault}{\sfdefault}{m}{sl}
\SetMathAlphabet{\mathsfit}{bold}{\encodingdefault}{\sfdefault}{bx}{n}

\hypersetup{
    colorlinks=true,
    citecolor=cyan,
    linkcolor=blue
}
\usepackage{cleveref}

\renewcommand{\cite}{\citet}

\begin{document}
\setlength{\abovecaptionskip}{10pt}
\twocolumn[
\icmltitle{Optimizing Diversity and Quality through Base-Aligned Model Collaboration}



\icmlsetsymbol{equal}{*}

\begin{icmlauthorlist}
\icmlauthor{Yichen Wang}{equal,uchi}
\icmlauthor{Chenghao Yang}{equal,uchi}
\icmlauthor{Tenghao Huang}{equal,usc}
\icmlauthor{Muhao Chen}{ucd}
\icmlauthor{Jonathan May}{usc}
\icmlauthor{Mina Lee}{uchi}
\end{icmlauthorlist}

\icmlaffiliation{uchi}{University of Chicago}
\icmlaffiliation{usc}{University of Southern California, Information Sciences Institute}
\icmlaffiliation{ucd}{University of California, Davis}

\icmlcorrespondingauthor{Yichen Wang}{yichenzw@uchicago.edu}

% \author{
% Yichen Wang\textsuperscript{1*},
% Chenghao Yang\textsuperscript{1*},
% Tenghao Huang\textsuperscript{2*}, \\
% ~\textbf{Muhao Chen\textsuperscript{3},}
% \textbf{Jonathan May\textsuperscript{2},}
% \textbf{Mina Lee\textsuperscript{1}} \\
% \textsuperscript{1}University of Chicago ~
% \textsuperscript{2}University of Southern California ~
% \textsuperscript{3}University of California, Davis \\
% \texttt{\{yichenzw, chenghao, mnlee\}@uchicago.edu}, 
% \texttt{tenghaoh@usc.edu} \\
% \textsuperscript{*}equal contribution
% }

\icmlkeywords{Machine Learning, ICML}

\vskip 0.3in
]



\printAffiliationsAndNotice{\icmlEqualContribution} % otherwise use the standard text.

\begin{abstract}
% Large language model (LLM) alignment has demonstrated impressive performance across a range of downstream tasks. However, recent studies have uncovered a key trade-off: while alignment improves quality and adherence to instructions, it often leads to significant output homogenization. This loss of diversity impacts performance in open-ended tasks such as instruction following and creative writing, where varied and original outputs are critical ...
% \todo{}

Alignment has greatly improved large language models (LLMs)' output \emph{quality} at the cost of \emph{diversity}, yielding highly similar outputs across generations, especially in open-ended generation tasks.
% \yc{resampling?} \ych{Just sampling, as ``re-sampling'' plus ``repeated'' are weird combination} from the same prompt.
% Large Language Model (LLM) alignment has significantly advanced the instruction following capability and generation quality.
% But the performance gain is often at the cost of diversity.
%, \ie{} the variation across responses to the same prompts. 
% Aligned models tend to collapse to generate a group of homogenized responses.
% For example, generating similar stories given the title when resampling, \mina{use the example in Figure 1 (e.g., names)}
% which could limit their utility in open-ended tasks like story generation. 
% \ych{merge the above two sentences together into one to make a quick introduction}
% Alignment Tuning enhances the instruction-following capabilities of Large Language Models (LLMs) but often at the expense of response diversity, leading to homogenized and less creative outputs. 
%
%
% To address the \problem{}, we introduce \name{}, an inference-time collaborative framework that combines a base and an aligned model through token-level routing. 
We propose Base-Aligned Model Collaboration (\name{}), an inference-time token-level model collaboration framework that dynamically combines a base LLM with its aligned counterpart to optimize diversity and quality.
% generalizing nudging \citep{fei2025nudging} into a full bidirectional model collaboration.
% To address the \problem{}, we introduce \name{} \mina{explain the acronym/initial}, an inference-time token-level collaborative framework that orchestrates a collaboration between an aligned LLM and its base version within a single decoding pass \mina{``within a single decoding pass'' - do we need this? doesn't ``token-level'' already mean this?}. Intuitively, the base model contributes diversity while the aligned model contributes quality and instruction-following.
% Inspired by nudging \citep{fei2025nudging} that small incorporation of aligned tokens can improve base model quality to achieve inference-time alignment, we generalize this idea: instead of occasional incorporations, \name{} uses a token-level routing mechanism that dynamically decides at each step which model to use, achieving the best of both worlds of quality and diversity during inference.
%In addition, we design
% Building upon prior work~\citep{fei2025nudging}, \name{} employs \emph{routing strategies} based on next-token prediction uncertainty and semantic role heuristics to determine which model to decode from at each step.
% Inspired by prior work~\citep{fei2025nudging}, 
% \name{} employs \emph{routing strategies} that determine, at each token, from which model to decode based on next-token prediction uncertainty and predicted contents' semantic role.
Using uncertainty and content-based signals, \name{} employs \textit{routing strategies} to determine, at each token, which model to decode from.
% \yc{Unlike the prior, which targets quality preservation, our design explicitly aims to enhance both diversity and quality.}
% reaching regions of the quality-diversity trade-off space inaccessible to either model alone.
% This is achieved through a token-level routing mechanism, which dynamically decides at each step which model to follow, inspired by nudging work showing that small aligned substitutions can boost base quality.
% Unlike prior approaches via retraining, prompt engineering, or multiple sampling passes, \name{} can generate diverse yet high-quality outputs and offers strong controllability via a family of routing strategies.
Prior diversity-promoting methods 
% such as retraining, prompt engineering, and multi-sampling methods, 
often improve diversity at the expense of quality or require expensive decoding or post-training.
In contrast, \name{} achieves both high diversity and quality post hoc within a single pass, while offering strong controllability.
We introduce a family of effective routing strategies and evaluate them across three open-ended generation tasks with 13 diversity and quality metrics. \name{} consistently surpasses state-of-the-art inference-time baselines.
% \mina{need to introduce ``routing'' first}
% We evaluate \name{} across three base-aligned model pairs and three open-ended generation tasks across 16 automatic metrics. 
% Results show that even a naive random router achieves a 12\% higher coverage of the trade-off space (\ie{} larger area under the trade-off curve) compared with all baselines, reflecting mitigated trade-off and broader flexibility. And that curated SOTA router of \name{} achieves superior quality-diversity trade-offs, outperforming baselines in 62\% of cases on average
% \ych{merge the above two sentences to make it more precise.}
% \name{} demonstrates that base-aligned collaboration mitigates \problem{}.
%We evaluate \name{} on three open-ended generation tasks across 13 automatic metrics for diversity and quality.
% across three base-aligned model pairs  
% Across three open-ended generation tasks and 13 metrics covering diversity and quality, \name{} consistently surpasses state-of-the-art baselines.
% , including single-model variants (base or aligned alone), decoding-based, prompting-based, and ensemble methods.
% bi-dimensional evaluation encompassing both quality and diversity metrics.
% \mina{state that these metrics are composed of both quality and diversity metrics}
% Even a simple random router expands the achievable quality-diversity trade-off space \mina{we haven't explained what ``space'' is. probably better to cut from abstract and introduce it in intro or briefly describe what it is} by 12\%
% Even a simple random router jointly enhances quality and diversity by 19\%, 
With our best router, \name{} achieves a 21.3\% joint improvement in diversity and quality, which is further supported by human evaluations.
% Human evaluations also mirror these improvements.
% \mina{add one sentence about implication of this work}
% The results suggest that collaboration between models at different alignment levels can improve and control diversity and quality.
Overall, our results demonstrate that collaboration between base and aligned models provides an effective and controllable mechanism for optimizing the diversity-quality trade-off.
% Code, data, and demo are available at.

{
\small
\vspace{2pt}
\textbf{\faLink~~}\href{https://yichenzw.com/base-aligned-collab.github.io/}{\textbf{Project Website}}

\textbf{\faGithub~~}\href{https://github.com/YichenZW/base-align-collab}{\textbf{Code}} ~~
{\raisebox{-0.34em}{\includegraphics[height=1.35em]{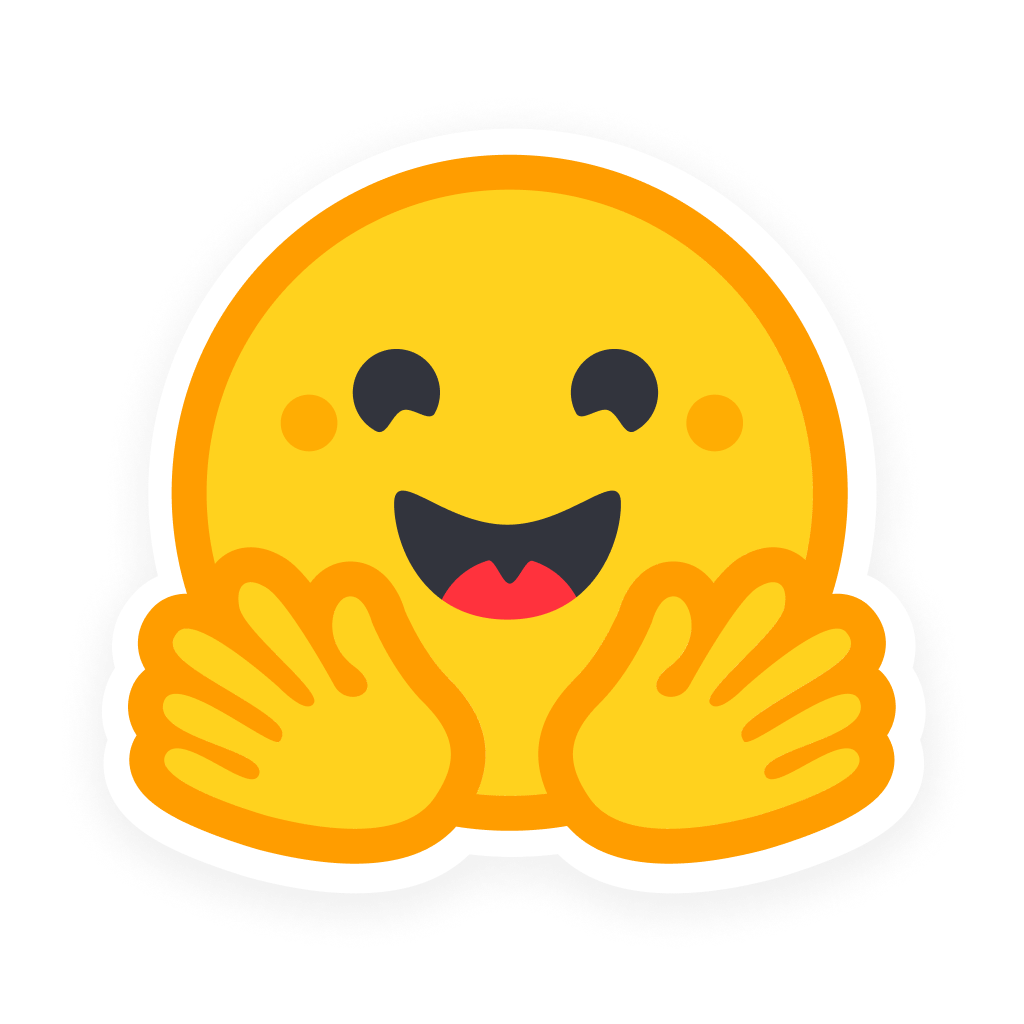}}}~\href{https://huggingface.co/datasets/ZachW/base-align-collab}{\textbf{Data}} ~~
\vspace{0.4mm}
{\raisebox{-0.34em}{\includegraphics[height=1.25em]{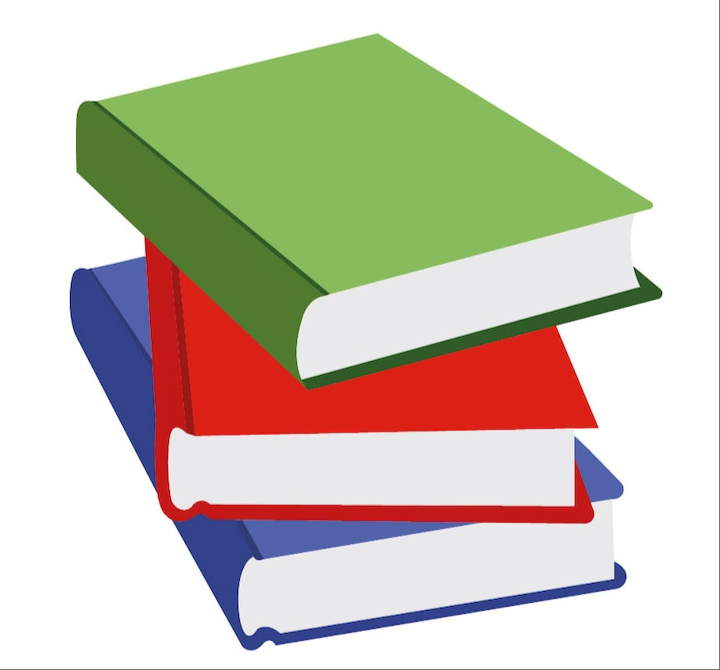}}}~\href{https://github.com/YichenZW/awesome-llm-diversity}{\textbf{Reading list}}
}
\end{abstract}

% \begin{center}
% \textbf{\faGithub~~Code:} \href{https://github.com/YichenZW/base-align-collab}{github.com/YichenZW/base-align-collab} \\
% {\raisebox{-0.34em}{\includegraphics[height=1.35em]{figures/hf-logo.png}}}
% ~\textbf{Data:} \href{https://huggingface.co/datasets/ZachW/base-align-collab}{huggingface.co/datasets/ZachW/base-align-collab} \\
% \vspace{0.4mm}
% {\raisebox{-0.34em}{\includegraphics[height=1.25em]{figures/books-emoji.png}}}
% \textbf{~Reading list:} \href{https://github.com/YichenZW/awesome-llm-diversity}{github.com/YichenZW/awesome-llm-diversity}
% \end{center}

\section{Introduction}

\begin{figure*}
    \centering
        \begin{overpic}[width=0.9\textwidth]{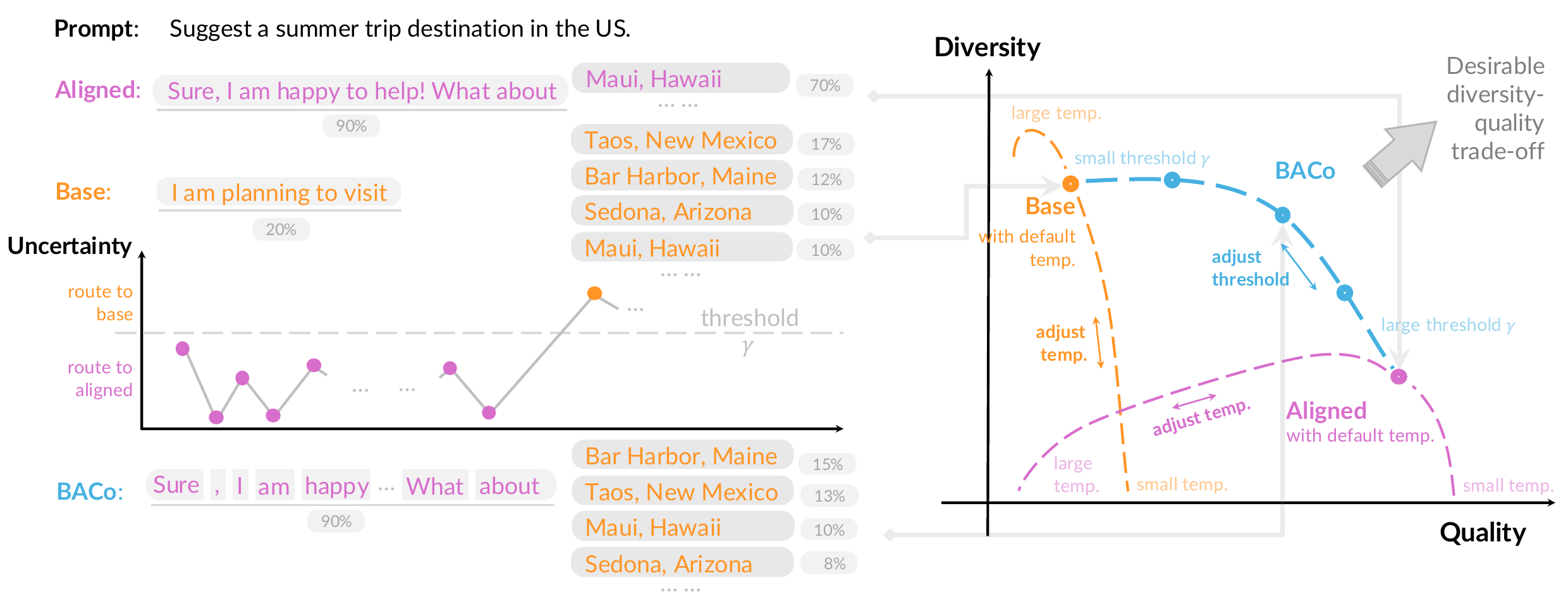}
        % adjust coordinates (x,y) until the labels align
        \put(0, -3.7){{(A) Example outputs from base, aligned, and \name{} models.}}
        % \mina{why is BACo output ("Sure, I am ... ... What about") in separate gray blocks?}}
        
        \put(62, -3.7){{(B) The \problem{} space.}}
    \end{overpic}
    \vspace{15pt}
    \caption{
    \name{} is an inference-time token-level model collaboration framework that combines a base model's diversity with its aligned counterpart's quality. 
    % (A) An example of sampled responses to an open-ended prompt from a base model, an aligned model, and \name{}. The aligned model tends to generate homogenized responses, though each response is high quality. The base model generates a group of diverse responses but with lower quality. \name{}, by routing between the two models based on next-token prediction uncertainty\ych{It is hard to plot the concept of ``uncertainty'' here in this figure, and we are not ready to introduce this concept here. Better leave this detail later.}, unifies their complementary strengths at inference time.
    (A) A comparison of generated outputs. 
    % \name{} combines a base and an aligned model at the token level. 
    The aligned model produces high-quality but low-diversity outputs, while the base model produces high-diversity but low-quality outputs. 
    \name{} optimizes both diversity and quality by dynamically \emph{routing} between them.
    The probabilities of token(s) are in grey next to text boxes. 
    % (B) Illustration of the \problem{} space in evaluation. The goal is to achieve diverse and high-quality generations (the top-right region). For both the base model and the aligned model, increasing temperature may improve diversity but often at a large cost to quality. \name{}, through collaboration between the base and aligned models, achieves a better trade-off. It also offers controllability at inference time. By adjusting the threshold, \name{} enables flexible traversal of the quality-diversity trade-off.
    % \mina{I think this figure can be simplified by a lot. what do you think about these ideas?
    % - only show tokens w/ prob >= 15\%
    % - this will make Aligned to have 1 destination (Havre de Grace)
    % - this will also make it much less confusing because currently Havasupai Falls come from both models
    % - I'd add a lot more outputs for Base to make the point that Base model is all about diversity
    % - align outputs of travel destinations for both models so that it's easy to see that there is 1 destination from Aligned and multiple from Base
    % - for BACo's travel destinations, why not match prob with Base?
    % - also, it's very confusing to see new options from neither Aligned and Base (e.g., Grand Canyon)
    % - add a label for "router" to treat it as a separate entity
    % }
    (B) Illustration of the \problem{} space. Single models face a steep trade-off, where improving diversity by adjusting configuration (e.g., by increasing temperature) degrades quality. \name{} achieves a better Pareto curve and allows for easy traversal across this frontier by adjusting the router's threshold.
    % The examples in this figure are modified for simplicity.
    % \mina{
    % - instead of ``larger temp.'' why not a bidirectional arrow with ``adjust temperature'', so that it's comparable to BACo's adjust threshold?
    % - ``Desirable diversity-quality trade-off''
    % - don't capitalize T in trade-off
    % }
    % \thh{the BACO line and label should be bolded to give readers a focus.}
    }  % And for many of them, a routing threshold serves as a tunable control parameter ($\gamma$).
% By adjusting this threshold, \name{} enables flexible traversal of the quality-diversity trade-off at inference time..}
    \vspace{-8pt}
    \label{fig:intro}
\end{figure*}

While alignment greatly improved large language models (LLMs)' output \emph{quality} in terms of instruction following and downstream task performance \citep{ouyang2022training}, it results in a stark reduction in output \emph{diversity} \citep{kirk2023understanding,zhang2025noveltybench,west2025base, spangher-etal-2025-creative, huang2025teaching}. 
% Large Language Model (LLM) alignment has led to impressive improvements across a wide range of downstream tasks, including instruction following and reasoning
% % human preference modeling, reasoning, coding, and agent-based applications 
% % \ych{almost everyone knows aligned LLMs can do many things, so I would suggest truncating this exhaustive list of downstream applications. }
% \citep{ouyang2022training}. 
% Yet, recent studies \citep{kirk2023understanding,zhang2025noveltybench,west2025base, spangher-etal-2025-creative} have highlighted a critical trade-off: although alignment improves task-specific performance, it often comes at the cost of a substantial reduction in the diversity of outputs. 
% Specifically, aligned models tend to produce more homogenized responses than their base models.
% \ych{remove this ``specifically...'' -- I think the previous two sentences almost read the same. }
%%%
% For example, when prompted for travel suggestions, base models usually yield diverse responses, whereas aligned models often converge on a single popular response (\Cref{fig:intro}A).
%%%
% \yc{Alignment produces an aligned model, trained to follow human preferences, from a base model that is only pretrained on next-token prediction.
Across repeated sampling, a model after alignment (\ie{} aligned model) tends to produce highly similar outputs, whereas a model before alignment (\ie{} base model)
% \mina{replace with ``a model after alignment (i.e., aligned model) tends to ..., whereas a model before alignment (i.e., base model)''} 
yields diverse outputs. 
For example, when prompted with ``suggest a summer trip destination in the US,'' a base model may produce diverse destinations across generations, while the aligned model often converges on a single dominant one (\Cref{fig:intro}A).
This \emph{\problem{}} 
% homogenization 
undermines utility in open-ended %applications. The consequences are multi-faceted:
generation tasks (\eg{} creative writing and dialogue) by 
%on a linguistic level, homogenization encourages
% encouraging formulaic language use and curbing creativity 
encouraging formulaic language use \citep{zhang2024lists,chakrabarty2025can}, diminishing creativity \citep{west2025base},
% \citep{chakrabarty2025can,zhang2024lists,west2025base}, %cognitive level, it limits
% narrowing the range of reasoning strategies \citep{chen2024not,yang2025alignment,ding2025dynamic,Xiao2025SubmodularOptimization}, %Critically, the impact of homogenization extends to human-AI interaction, where it has been shown to 
and suppressing ideation in human-AI interaction \citep{padmakumar2023does,Meincke2025,ashkinaze2025ai, spangher2025novel, spangher2025newsinterview}.
% On a linguistic level, this lack of diversity manifests through the loss of unique style \citep{chakrabarty2025can}, increased format similarity \citep{zhang2024lists}, and diminished creativity \citep{west2025base}. More fundamentally, it constrains models to a narrow repertoire of problem-solving strategies~\citep{yang2025alignment}, which reduces diversity in reasoning \citep{chen2024not, ding2025dynamic}, and 
% data synthesis \citep{kim2024evaluating, yang2025measuring}, 
% \ych{I do not think data synthesis is what the model is trained to do or should be viewed as a model capability -- you of course can argue this, but I feel this kind of debate is wasting your chance of attracting readers' attention and energy. }
% automatic research \citep{Xiao2025SubmodularOptimization}.
% , and gaming \citep{west2025base}. 
% The effects also extend to human-AI interaction, where diminished variation in model outputs has been shown to stifle human creativity and ideation \citep{padmakumar2023does, Meincke2025, ashkinaze2025ai}.
These findings motivate methods to improve diversity in aligned LLMs.

Prior diversity-promoting methods  attempt to address 
%\ych{Take care of ``easing'', ``lessen''..., you should use a consistent way to describe your action.} 
the \problem{} at both the training and inference stages (\Cref{sec:related_work}). The former \citep{lanchantin2025diverse, chung2025modifying, li2025jointlyreinforcingdiversityquality} incorporate explicit diversity objectives into preference optimization during reinforcement learning.
% (e.g., by measuring deviation among winning responses or comparing alternative response groups).
While effective at improving diversity, such methods require modifying the model's output distribution, which can compromise desirable alignment properties such as safety and helpfulness \citep{qi2023fine}.
The latter consists of decoding-based techniques, such as adjusting temperature and beam search \citep{vijayakumar2016diverse}, as well as prompt-based techniques, including in-context learning \citep{meyerson2024language}, prompt paraphrasing \citep{zhang2025noveltybench}, and multilingual back-translation \citep{wang2025multilingual}.
However, these inference-time techniques typically require multiple decoding passes or long-horizon planning to improve diversity, and may still disproportionately degrade generation quality \citep{peeperkorn2024temperature}.
% , \etc{} 
% \ych{Minor: Using ``including A, B, and C.'' instead of ``A, B, C, etc.'' -- it makes you look intentionally ignoring sth, or hide the complexity, which weakens your tone.  }
% Despite their promise, these approaches face key limitations:
% Training-time methods require altering the model's output distribution, which risks undermining desirable alignment properties, such as safety and helpfulness, in pursuit of diversity.
% Inference-time methods often demand multiple decoding passes or long-horizon planning to improve diversity, and may still disproportionately degrade generation quality \citep{peeperkorn2024temperature}.
This reveals a fundamental limitation of forcing a single model to excel at both diversity and quality. 

To overcome this limitation, we introduce Base-Aligned Model Collaboration (\name{}), an inference-time token-level model collaboration framework (\Cref{sec:method-main}) that combines the complementary strengths of a base model for diversity and its
aligned counterpart for quality.
% a base model's diversity with its aligned counterpart's quality. 
\name{} operates via a lightweight, token-level \emph{routing strategy} that dynamically switches between the two models in a single decoding pass, requiring no fine-tuning or prompting (\Cref{fig:intro}A). 
% Unlike prior methods like nudging \citep{fei2025nudging}, which are performance-capped by the aligned model,
% \yc{\citet{fei2025nudging} provide evidence that base and aligned models largely agree on most tokens (namely \textit{superficial alignment hypothesis}) and are practically possible for them to collaboratively generate. However, they adopt the method to infer time-aligned large models based on small models, solely focusing on quality, where the performance of collaboration is capped by the aligned model.}
This design is motivated by recent findings suggesting that collaboration between base and aligned models is feasible.
Specifically, \citet{fei2025nudging} demonstrate that base and aligned models largely agree on next-token predictions, 
a phenomenon known as \emph{superficial alignment} \citep{lin2023unlocking}.
% Recently, \citet{fei2025nudging} demonstrated that base and aligned models largely agree on next-token predictions, a phenomenon known as the superficial alignment \citep{lin2023unlocking}, suggesting that collaboration between them is feasible.
% Their method, \textsc{nudging}, employs a small aligned model to guide a larger base model during decoding, effectively improving its quality but without considering diversity. 
% However, their approach uses small aligned models to guide larger base models, focusing solely on quality; as a result, the performance of the collaboration remains bounded by that of the aligned model.
% Our approach produces generations superior to either base or aligned model alone, jointly on diversity and quality (\Cref{fig:intro}B). 
% \mina{misleading to call out ``the performance of the collaboration remains bounded by that of the aligned model'' as the main limitation and say we are superior because our method's quality alone is not superior}
In contrast to the prior work that leverages this phenomenon to improve
base-model quality, our goal is to jointly optimize diversity and quality,
surpassing either model alone through collaboration (\Cref{fig:intro}B).
% In contrast, our goal is to jointly optimize diversity and quality  (\Cref{fig:intro}B).
% by a base and aligned model collaborate. 

% Because diversity often trades off with quality, the ideal balance between them varies across tasks. 
% \ych{This cause-effect reads weird -- why trade-off brings task-varied balance? I think the missing link is that different task brings different requirement, thus ... [Suggestion: "The ideal balance between diversity and quality is task-dependent. Because these two properties are often in a trade-off, systems must be tuned to prioritize diversity for creative tasks and quality for factual tasks."]} 
% The ideal balance between diversity and quality is task-dependent. Because they are often in a trade-off, systems must be tuned to prioritize diversity for creative tasks and quality for factual tasks.
% The ideal balance between diversity and quality is task-dependent.
% Since the two are often in trade-off, xxx
% Given the trade-off between diversity and quality, the ideal balance varies across tasks.
Importantly, the diversity-quality trade-off cannot be resolved by a fixed design.
% Different tasks naturally favor different points along the diversity-quality spectrum.
Different tasks naturally favor different operating points along this spectrum, and user preferences may further vary based on context.
% For example, argumentative writing tasks often demand high-quality (\ie{} accurate and factual) outputs with less emphasis on diversity (\eg{} multiple valid expressions), whereas brainstorming tasks benefit more from high-diversity (\ie{} broader exploration and novelty), even if some outputs are lower quality (\eg{} less polished in writing). 
% Given the diversity-quality trade-off, since we can achieve the disable high-diversity, high-quality region, different tasks may require difference ballance the diversity level and quality level.
% For example, reasoning tasks could have a high requirments (lower bound) of the quality to ensure facutuality (80\% quality + 20\% diversity), while brainstorms focus more diversity for rare but novel ideas and more acceptable for average quality to be lower (20\% quality + 80\% diversity).
% Moreover, different people may prefer different trade-offs depending on their preferences or intent.
% Therefore, a practical framework should enable \textit{controllability}, the ability to adjust along the diversity-quality spectrum according to task or preference.
A practical framework, therefore, requires \textit{controllability}, the ability to adjust generation along the diversity-quality spectrum on demand.
% A method well suit for diverse tasks request controllability on the diversity-quality balance  
% Moreover, \name{} design with a adjustable threshold (\Cref{fig:intro}) to control the ballance.
\name{} provides such controllability through an adjustable routing threshold (\Cref{fig:intro}), which continuously tunes the contribution of the base
and aligned models during decoding.
% that tunes the contribution of base and aligned model.
% Beyond this performance gain, the routing strategy itself is highly controllable. 
% Moreover, since a single routing strategy does not guarantee diversity across all aspects, 
% We also introduce a family of strategies, including logit-based, content-based, and their combinations, within a proposed design space (\cref{sec:method:router}). 
In addition, we introduce a family of routing strategies within a unified design space (\cref{sec:method:router}) that consistently improve the
diversity-quality trade-off. 
These strategies capture complementary signals and exhibit different strengths across diversity dimensions, providing an additional axis of control beyond the routing threshold.
We validate our approach across three open-ended generation tasks: instruction following, dialogue, and creative writing (\Cref{sec:experiments}).
In open-ended generation, both diversity and quality can be defined in numerous ways, and no single metric provides a complete evaluation.
Therefore, rather than treating evaluation as a single objective, we formulate it as a collection of bi-dimensional diversity-quality trade-off spaces, each defined by a specific pair of metrics.
Focusing on diversity, we evaluate 11 diversity and 2 quality metrics, resulting in $11 \times 2$ spaces, to assess \name{}'s consistent
improvements. 
In addition to instruction-following and dialogue tasks which have relatively short outputs, we also evaluate long-form generation, where we measure long-form diversity at the discourse level in terms of plot structure and emotional flow \citep{tian2024large}. %,   
% as a high-level narrative abstraction \citep{tian2024large}.
% Additionally, we extend evaluation for long-form generation by story arcs \citep{tian2024large} as a high-level narrative abstraction.
% For creative writing, we additionally evaluate long-form diversity 
% for evaluating the narrative diversity of this long generation.
% In addition to diversity metrics, we introduce creativity as a high-level evaluation objective. Creativity is particularly important in open-ended generation tasks, where the goal is not only to vary surface forms but also to produce novel and compelling ideas. 
% While diversity focuses on differences across outputs, creativity measures whether those differences are meaningful, imaginative, or original.
% Beyond diversity, we assess creativity as a higher-level objective, measuring the novelty and originality of the generated content.
% \todo{We apply two creativity metrics...} 
% We complement automatic evaluations with human evaluations of diversity and quality, providing human-centered validation and comparison with automatic results.
Our evaluation focuses on open-ended generation settings where diversity is a first-class objective, for which benchmarks designed primarily for single-answer
accuracy (e.g., math and code reasoning) are less suitable.
% Benchmarks designed primarily for single-answer accuracy, such as math and code reasoning, are therefore not well suited for assessing this trade-off.
Nevertheless, as a cross-check, we additionally evaluate \name{} on tasks with verifiable quality criteria, including mathematical reasoning and verifiable instruction following, to ensure that diversity improvements do not
come at the cost of severe quality degradation.
% Nevertheless, to validate that the observed gains are not artifacts of open-ended evaluation metrics, we additionally evaluate \name{} on mathematical reasoning tasks and verifiable instruction following.
% However, to verify that our gains are not driven by overfitting to open-ended evaluation metrics, we additionally evaluate \name{} on verifiable instruction following and mathematical reasoning.
We complement automatic evaluations with human evaluations of diversity and quality.

In our experiments (\Cref{sec:results}), we demonstrate that \name{} achieves a new state-of-the-art in optimizing the \problem{} in inference time. 
% Even a simple router that routes randomly yields a 19.0\% joint improvement in diversity and quality across tasks and metrics, underscoring the effectiveness of the framework.
% Even a simple random router yields a 19.0\% joint improvement, indicating that the base-aligned model collaboration framework itself is effective, while informed routing further amplifies these gains.
Overall, \name{} with our best router achieves a 21.3\% improvement over the strongest
inference-time baselines, with even larger gains on semantic diversity metrics.
% \mina{give one or two easy to understand examples of baselines, e.g., base/aligned model alone with varying temperature values}
% the Pareto frontier, evidencing clear dominance .
These improvements are consistent across tasks and evaluation metrics, and are
further supported by human evaluations.
% These results are further supported by human evaluations.
% Furthermore, we analyze model collaboration patterns under \name{} and human-perceived creativity of its outputs in \Cref{sec:analysis}. 
% View the base and aligned model as demonstration of checkpoints with different alignment levels during the alignment process, we suggest that collaboration between models at different alignment levels can improve and control diversity and quality.
By viewing the base and aligned models as two checkpoints during training, our results suggest that collaboration across checkpoints can enable effective control over diversity and quality.
Overall, \name{} provides a simple framework for base-aligned model collaboration, effectively improving both diversity and quality. 

In summary, our contributions are threefold:
\begin{enumerate}[wide, labelwidth=!, labelindent=0pt]
\item[\circone] We propose \name{}, an inference-time token-level model collaboration framework that combines a base model and its aligned counterpart, along with a family of lightweight routing strategies, to produce high-diversity and high-quality outputs across generations. 
% , and introduce a family of effective routing strategies.
% We further propose a family of routing strategies fit into the framework. % under a design space.
% , , while allowing fine-grained control over \problem{}.
% \item[\circone] We propose \name{}, a lightweight, inference-time framework that synergizes base and aligned models. Through a family of routing strategies, it produces high-quality, diverse outputs while offering users fine-grained control over \problem{}.
\item[\circtwo] We formulate the diversity-quality trade-off as a collection of bi-dimensional evaluation spaces and conduct a comprehensive evaluation across $11 \times 2$ metric pairs, including long-form 
diversity and human evaluation.
% We evaluate the diversity-quality trade-off across $11\times2$ evaluation spaces, extend automatic metrics to long-form generation, and conduct human evaluation.
% We evaluate the \problem{} across 11 $\times$ 2 evaluation spaces, and we also extend automatic diversity evaluation to long-form evaluation and conduct human evaluation. 
% 
% \item[\circtwo] We develop an evaluation protocol tailored to the \problem{}, repurposing existing open-ended generation datasets while introducing new metrics that measure controllability (the extent to follow user preferences in balancing quality and diversity) and dominance 
% (how often a method expands the diversity-quality Pareto frontier).
% \item[\circtwo] We introduce a new evaluation benchmark tailored for \problem{}. The benchmark includes novel metrics to rigorously quantify a method's controllability (its ability to follow user preferences on balancing diversity and quality) and its attribution to the Pareto frontier. 
% \item[\circthree] Through extensive experiments, we show that \name{} outperforms powerful baselines. It demonstrates clear Pareto dominance, significantly outperforming baselines with a 14.2\% improvement in controllability and 32.7\% coverage of the combined frontier, with its effectiveness confirmed by human evaluation.
\item[\circthree] Through extensive experiments on three open-ended generation tasks (i.e., instruction following, dialogue, and
creative writing), we show that \name{} consistently outperforms strong 
baselines. %, achieving substantial improvements.
% Through extensive experiments across instruction-following, dialogue, and creative writing, we show that \name{} consistently outperforms strong baselines. It achieves a optimization on \problem{}. 
% Human evaluations further validate its advantages.
% in quality, diversity, and creativity.
\end{enumerate}

\section{Preliminary}
\fakeparagraph{Large Language Models (LLMs).} LLMs are typically trained to autoregressively predict the next token of the output $\outputval$ given a prompt $\inputval$.
The conditional probability is factorized as
% of a output $\outputval$ given input $\inputval$ and model parameter $\theta$ is 
% and the probability of output $P\left(\outputval_{1:N} | \inputval; \theta \right)$ carespn be decomposed as: 
% $P\left(\outputval_{1:N} | \inputval; \theta \right)=\Pi_{t=1}^{N}P\left(\outputval_t | [\inputval, \outputval_{1:t-1}]; \theta \right)$, where $\outputval_{1:t-1}$ denotes the generated output prefix up to position $t-1$. 
$P\left(\outputval | \inputval; \theta \right)=\Pi_{t}P\left(\outputval_t | [\inputval, \outputval_{<t}]; \theta \right)$, where $\outputval_{<t}$ denotes the output prefix generated up to position $t$-$1$, and $\theta$ denotes the model parameter.
% Modern LLMs go through multiple training stages.

Alignment is the process of fine-tuning an LLM to align its outputs with human intent, ethical principles, and desired behavioral norms, typically through instruction tuning or reinforcement learning from human feedback (RLHF) \citep{ouyang2022training, bai2022training}.
We use \textit{base models} to denote models without alignment tuning (e.g., \texttt{Llama-3-8B}) and \textit{aligned models} to denote those further optimized with alignment (e.g., \texttt{Llama-3-8B-Instruct}) \citep{dubey2024llama}.
% , such as instruction tuning or reinforcement learning from human feedback (RLHF) \citep{ouyang2022training, bai2022training}.
% Lastly, we define the entropy of each next-token distribution as $H(\outputVar_t | [x, \outputval_{<t}]; \theta) = \sum_{\outputval_t} P\left(\outputval_t | [\inputval, \outputval_{<t}]; \theta \right)\log P\left(\outputval_t | [\inputval, \outputval_{<t}]; \theta \right)$.
% \ych{This part serves as the ``Notation''. I would suggest separate this part and section 2.1 into an individual section called ``Preliminary'' to set up the stage. }

\fakeparagraph{Diversity and Quality Measurement.} 
In this paper, we measure diversity over a group of outputs independently generated from the same prompt $x$: $\mathcal{Y}(x) = \{ y^{(1)}, \dots, y^{(k)} \}, y^{(i)} \sim P(\cdot \mid x; \theta)$. This group-level diversity is denoted as $D(\mathcal{Y})$ \citep{kirk2023understanding, west2025base}  (e.g., the clustering-based approach in ~\citet{kuhn2023semantic}). 
% Hereafter, we call the group-level diversity diversity for simplicity.
% Quality is modeled as a task-specific objective \mina{?? what do you mean quality is objective} $Q(x,y)$ for each output, typically given by a reward model \mina{or human evaluator}, reflecting alignment \mina{do you mean quality is inherently about alignment?} objectives (\eg{} instruction-following) \citep{lambert2024rewardbench, zhang2025noveltybench}. 
Quality is modeled as $Q(y | x)$ for each output given the prompt, typically by a reward model or human evaluator, reflecting linguistic fluency and instruction-following \citep{lambert2024rewardbench, zhang2025noveltybench}. 
% reflecting alignment \mina{do you mean quality is inherently about alignment?} objectives (\eg{} instruction-following) \citep{lambert2024rewardbench, zhang2025noveltybench}. 
The group-level quality is then defined as the average quality across all outputs in a group: 
$Q(\mathcal{Y}) = \sum_{i=1}^k Q(y^{(i)} | x)$.
% (\ie{} $\mathcal{Y}(x)$).
For simplicity, we refer to group-level diversity and group-level quality as diversity and quality in this work.

\paragraph{Diversity-Quality Trade-off.}
\label{sec:problem}

Alignment, while effective at improving output quality, comes at the cost of reduced output diversity \citep{lu2025ai, west2025base, yang2025alignment}. 
% \mina{with these three works in citations, do we still need to show empirical evidence to show the existence of the trade off?} 
% \ych{Re: The point of this section is to demonstrate more clear what do we mean by \problem{} and how severe it could be under our main evaluation setting (NoveltyBench). I still think we should keep this section, as not everyone is that familiar with \problem{}, which is of central motivation for us. Also, we need a dedicated place to formalize the concept of ``diversity-quality space''. }
% The literature focuses more on token-level distribution shrinkage.
% As a pilot study, we empirically demonstrate also holds at the whole-output level. 
To demonstrate this trade-off and quantify its magnitude,
we run a preliminary experiment with \texttt{Llama-3} on a subset of WildChat~\citep{zhao2024wildchat}.
% As a pilot study to demonstrate the trade-off, 
We first situate the performance of the base and aligned models with
% \begin{wraptable}{r}{\linewidth}
\begin{table}[t]
\centering
\small
\renewcommand{\arraystretch}{1.2}
% \vspace{-5pt}
\begin{tabular}{lcc}
\toprule
& \textit{Diversity} & \textit{Quality} \\
\textit{Model} & \textit{(\#Clusters)} $\uparrow$ & \textit{(Reward)} $\uparrow$ \\
\midrule
\texttt{Llama-3-8B}     & \textbf{8.13} & 1.28 \\
\texttt{Llama-3-8B-Instruct} & 2.58 & \textbf{7.62} \\
\bottomrule
\end{tabular}
\vspace{7pt}
\caption{Diversity and quality of \texttt{Llama-3-8B}'s base and aligned models in our preliminary experiment. 
The results demonstrate a clear \problem{} in the two model's performance.
Diversity is measured by the number of semantic equivalent clusters of the output group, and quality is the average reward per output from another LLM. 
}
% \vspace{-15mm}
\label{fig:prelim}
% \end{wraptable}
\end{table}
default configuration (the two noted points in \Figref{fig:intro}B) within the diversity-quality space.
% , providing an absolute reference.
% for their respective trade-offs. 
% We extend previous analyses (mainly on synthetic games, novel/poem generation, and reasoning chains) by applying it to   dialogue.
% \ych{I feel this is not true for summarizing literature -- we cannot say previous works only does token-level study, that is not accurate. (Also whole-output reads weird) I think maybe you can say, previous works focuses more on synthetic games, novel or poem generations and reasoning chain generations, and here we extend previous work to include real-world user dialogue. You can rephrase to make it fit the context}
% on the \texttt{Llama-3} model family. 
Following the evaluation protocol of the diversity-focused benchmark NoveltyBench \citep{zhang2025noveltybench}, we evaluate \texttt{Llama-3}'s base and aligned models on an open-ended subset of WildChat \citep{zhao2024wildchat}.
Here, diversity is measured as the number of semantic equivalent classes of the output group via \cite{zhang2025noveltybench}'s clustering, and quality is measured as the average reward per output from \texttt{Skywork-Reward-Gemma-2-27B} \citep{liu2024skywork}.\footnote{These are two of many possible measurements introduced later in the paper. We use them here as representative examples for the pilot study, as they are among the widely adopted metrics.} We sample 10 outputs per prompt.
% , using the number of semantic equivalent clusters to measure diversity and a reward score to measure quality. 
% \mina{
% it might be worth cutting this empirical result to avoid confusion. for instance, 
% why are we using \#cluster for diversity and reward for quality?
% what about the other 11 diversity metrics and 1 metric for quality?
% what is this paragraph adding that wasn't known before or is not supported by prior work?}
% \ych{Re Mina: it is too much to introduce that many metrics here. Here our strategy is simply, we follow some previous proposed evaluation protocol proposed in NoveltyBench (using cluster for diversity and reward for quality). As I explained in previous note, I do not think we should assume general readers know \problem{} and read all these prior work, but knowing ``how severe the tension is'' is of key importance for our motivation part later. }
As illustrated in \Cref{fig:prelim}, the \problem{} is stark: the base model is \textbf{3.15x more diverse}, whereas the aligned model has \textbf{5.95x higher quality}. 
Inherently, this performance trade-off stems from alignment's tendency to reduce the entropy of the next-token prediction distribution, 
% ($H_\theta\left(\outputval_{t} | \inputval, \outputval_{<t} \right)$), 
concentrating probability mass on fewer, high-quality tokens, a phenomenon known as \textit{mode collapse} \citep{lin2023unlocking, Shumailov2024, hamilton2024detecting, yang2025alignment, cui2025entropy}.

% \thh{One more thing worth clarifying here is why dont we use base model? Is there any other alternative? e.g. why dont we train a creative model?}
% \ych{We previously plan to clarify this later -- this section is only used for motivating our method. But yeah perhaps I can use very brief sentence to highlight this. }

% This presents a dilemma: should we use a diverse but less reliable base model, or a high-quality but more uniformly aligned model?  
% This presents a dilemma: the selection we can either have a high-diversity, low-quality base model or a low-diversity, high-quality aligned model.
This presents a dilemma: one can either use a high-diversity but low-quality base model, or a high-quality but low-diversity aligned model.
The single-model paradigm is insufficient, as neither extreme is ideal for all applications.
% \mina{weak argument; unclear in what applications we want something in the middle though}
% {For example, story writing requires both sufficient diversity to explore rare but novel ideas and sufficient quality to ensure plot coherence; research ideation requires both sufficient diversity to think exploratively and sufficient quality to ensure proposal reasonableness.}
Hence, we argue that an ideal method is able to pursue the best of both worlds. 
To this end, we formalize the problem in a two-dimensional \emph{diversity-quality space} $S=\{(D, Q)\}$. 
 In this space, any given method under specific configurations $\gamma$ (\eg{} sampling parameters) is evaluated to be a single point.
 % including those models trained to promote diversity \citep{ismayilzada2025creative, lanchantin2025diverse}. 
 % An ideal system should not be a single point \thCRh{a system cannot be a point. We should rewrite this sentence to make it clear. I think what we are saying is an ideal system should have flexibility based on user's needs. how about we just say ``an ideal system should approximate pareto frontier''?} but should instead
 An ideal method, by adjusting $\gamma$, should approximate the \emph{Pareto frontier}: the set of optimal solutions where diversity cannot be improved without sacrificing quality, and vice versa. 
 This frontier represents the best possible trade-offs.

\section{\name{}: Base-Aligned Collaboration for  Diversity and Quality}
\label{sec:method-main}

Recent work has provided empirical evidence for the superficial alignment hypothesis \citep{zhou2023lima, lin2023unlocking}, which suggests that a base model and its aligned counterpart largely agree on next-token predictions.
Building on this observation, \citet{fei2025nudging} show that introducing only a small fraction of aligned-model tokens into a base model’s decoding can recover task-specific performance comparable to that of the aligned model. However, such
approaches primarily focus on improving the quality of the base model, without explicitly addressing the loss of diversity introduced by alignment.

% Recent work has shown empirical evidence on the \textit{superficial alignment hypothesis} \citep{zhou2023lima, lin2023unlocking}, which suggests that a base model and its aligned counterpart largely agree on next-token predictions.
% Building on this, \citet{fei2025nudging} demonstrate that their in-context alignment method, \textsc{nudging}, introduces only a small proportion of aligned-model tokens to a base model's decoding for it to achieve task-specific performance (including instruction-following) comparable to the aligned model.
% This suggests the feasibility of collaboration between them.
% models with different alignment levels.

Motivated by this gap, we hypothesize that
\begin{tcolorbox}[colback=gray!5!white, colframe=gray!60!black]
{Collaboration between a less-aligned,
higher-diversity model and a more-aligned, higher-quality model during inference
can better balance the diversity-quality trade-off than either model alone.}
% without additional training}. 
\end{tcolorbox}
% \yc{textbox}
Base and aligned models form a natural pair to test this hypothesis, as they are readily available off the shelf and exhibit complementary strengths.

Based on this hypothesis, we propose \textbf{\name{}}, an inference-time framework that orchestrates collaboration between a base model ($P_{\text{base}}$), serving as
a source of diversity, and its aligned counterpart ($P_{\text{aligned}}$), serving
as a source of quality, at the token level.\footnote{The token-by-token nature of LLM autoregressive decoding makes token-level control feasible.}

% Hence, based on our hypothesis, we propose \textbf{\name{}} that orchestrates a collaboration between a {base model} ($P_{\text{base}}$) for diversity and its {aligned counterpart} ($P_{\text{aligned}}$) for quality at the token level.

At the core of \name{} is a \textbf{router}, a lightweight decision module that determines, at each decoding step, which model should generate the next token. 
The router operates according to \textbf{routing strategy(ies)} $\mathcal{R}$, which selects between base and aligned models accordingly.\footnote{The router can be based on a single routing strategy or combine multiple ones.}
Intuitively, the router acts as a ``gatekeeper'': it routes the next token generation to the base model when diversity is desired and to the aligned model when quality is desired.

Formally, \name{} orchestrates the two models as:
% \begin{small}
% \begin{equation}
% \label{eq:moe_main}
%     P_{\text{\name{}}}\left(\outputval_t | c_t \right) = w_{\text{base}} \cdot P_{\text{base}}\left(\outputval_t | c_t; \theta_{\text{base}} \right) + (1-w_{\text{base}}) \cdot P_{\text{aligned}}\left(\outputval_t | c_t; \theta_{\text{aligned}} \right)
% \end{equation}
% \end{small}
\begin{small}
\begin{equation}
\label{eq:moe_main}
\begin{split}
    P_{\text{\name{}}}(\outputval_t | c_t) ={} & w_{\text{base}} \cdot P_{\text{base}}(\outputval_t | c_t; \theta_{\text{base}}) \\
    & + (1-w_{\text{base}}) \cdot P_{\text{aligned}}(\outputval_t | c_t; \theta_{\text{aligned}})
\end{split}
\end{equation}
\end{small}
where $c_t = [\inputval, \outputval_{<t}]$, and the gating weight $w_{\text{base}} \in \{0,1\}$ for each candidate token $\outputval_t$ is given by the router:
\begin{small}
\begin{equation}
\label{eq:moe_gate_corrected}
    w_{\text{base}} = \mathbb{I}\left[\mathcal{R}\left(\outputval_t | c_t, P_{\text{base}}, P_{\text{aligned}}\right)=\text{base}\right]
\end{equation}
\end{small}

In practice, since one word may consist of multiple tokens, we restrict switching to word boundaries to prevent erroneous generation when the two models use different tokenizations. 
% (implementation details in \appref{app: implementation_details}).
% Because the base and aligned models agree on most next-token predictions, only occasional model switching is required, making \name{} naturally less costly. 
% Because the base and aligned models agree on most next-token predictions, in practice,
% it incurs only a modest additional computational cost.
% More details and discussion on implementation is in  \appref{app: implementation_details} and \appref{sec:future_work}.
Full decoding pseudocode is provided as Algorithm~\ref{alg:baco} in \appref{app: implementation_details}; further implementation and discussion are in \appref{app: implementation_details} and \appref{app:analysis}.
\subsection{Routing Strategy Design}
\label{sec:method:router}

At the core, \name{} is a lightweight router that determines which model to route to at each decoding step. 
% Conceptually, designing a router involves choosing what information to use as prior for routing decisions, and which model to switch to given the information.
A routing strategy specifies (\textit{i}) what information to use as prior for routing decisions, and (\textit{ii}) which model to switch to given the information.
% A routing strategy includes a threshold parameter that controls the attribution to the two models, providing a means to continuously adjust the balance between diversity and quality.
Each strategy includes a \textit{threshold parameter} $\gamma$ that, analogously to decoding temperature, continuously adjusts the diversity-quality balance: larger $\gamma$ biases toward the base model (more diversity), smaller $\gamma$ biases toward the aligned model (more quality).

% that provide a means to continuously adjust the balance between diversity and quality.
% The effectiveness of the router is determined by its routing strategy(ies).
We design routing strategies based on two complementary signal categories: logit-based and content-based. Each category captures
a distinct 
yet widely applicable perspective.
% yet widely applicable intuitions, and are supported by our empirical results.
% We consider two broad categories of routing strategies, logit-based and content-based, primarily distinguished by what prior is used.
% These categories capture complementary intuitions and are broadly applicable across tasks, supported by our empirical results.

\textbf{Logit-Based.} 
% These strategies leverage the next-token probability distribution to infer model's uncertainty. 
% High uncertainty often signals an opportunity for diverse, open-ended generation. 
% This category is motivated by a \textit{model-centric} perspective. Tokens with low maximum probabilities or high entropy indicate positions where the model is uncertain about the next step. These points naturally allow greater flexibility for open-ended exploration because the model’s continuation is less predictable, making them suitable candidates for diversification \cite{yang2025alignment, wang2025beyond}.
Logit-based strategies leverage the next-token prediction distribution to infer the model
uncertainty, reflecting a \emph{model-centric} perspective \citep{fei2025nudging, zheng2024citer, leviathan2023fast}.
High uncertainty, indicated by low maximum probability or high entropy, suggests that
multiple continuations are plausible, making such positions natural opportunities for diversification \citep{yang2025alignment, wang2025beyond}.
% Heuristics can be built on simple statistics such as the maximum token probability, the average top-k token probability, or the entropy of the distribution. 
% Key strategies implemented under this category include:
Concretely, we implement:
% \footnote{Though we mainly use base model logits or entropy in the main text, we could also adopt aligned model's logits or both of them. More details could be found in \todo{appendix}.}
% Key examples \jon{examples of what? I can't resolve the antecedent} include:
% \jon{Don't understand}}
% \begin{enumerate}[wide, labelwidth=!, labelindent=0pt]
\circone \textbf{\name{}-\textsc{P}} routes to the base model when its maximum token probability falls below a threshold $\gamma$, i.e., $\max_{y_t} P_{\text{base}}(y_t \mid \cdot) < \gamma$;
\circtwo \textbf{\name{}-\textsc{H}} routes to the base model when its next-token
entropy exceeds, i.e., $H_{\text{base}}(\outputVar_t \mid \cdot) = \sum_{\outputval_t} P_{\text{base}}\left(\outputval_t | \cdot \right)\log P_{\text{base}}\left(\outputval_t | \cdot \right) > \gamma$; \etc
% i.e., $H_{\text{base}}(\outputVar_t \mid \cdot) = \sum_{\outputval_t} P\left(\outputval_t | [\inputval, \outputval_{<t}]; \theta \right)\log P\left(\outputval_t | [\inputval, \outputval_{<t}]; \theta \right) > \gamma$, 
% we define the entropy of each next-token distribution as $H(\outputVar_t | [x, \outputval_{<t}]; \theta) = \sum_{\outputval_t} P\left(\outputval_t | [\inputval, \outputval_{<t}]; \theta \right)\log P\left(\outputval_t | [\inputval, \outputval_{<t}]; \theta \right)$.
% \jon{you didn't really introduce entropy though $H$ is standard. But also this is written odd since $y_t$ is a specific choice and entropy should be over a distribution. Also not great to have the same hyperparameter variable name ($\gamma$) for both} 
% indicating high uncertainty.

\textbf{Content-Based.} 
Content-based strategies adopt a \emph{language-centric} perspective, making routing
decisions based on the semantic roles of predicted tokens.
The motivation is twofold.
% \jon{Keep using the circles for enumeration} 
First, linguistic structures such as content words often correspond to semantic or stylistic branch points where diversity is most perceptible to humans \citep{yao2019plan, sims2019literary}.
% First, linguistic features may provide cues for when diversity is desired (e.g., verbs are critical in narrative generation . Linguistic structures such as content words, punctuation, and other syntactic or semantic features serve as natural ``branch points'' in text. Triggering routing decisions at these points is likely to produce meaningful and human-perceived diversity.
% These strategies determine switching decisions based on the semantic roles of the predicted tokens themselves.
% linguistic categories are of importance in the planning of open-ended generation. 
% \mina{unnecessarily convoluted writing -- just say ``Linguistic features may provide important cues for when diversity is desired.''}
% For instance, verbs are critical for narrative generation \citep{yao2019plan, sims2019literary},
% and nouns are critical in question answering \citep{li2019entity}.
Second, \cite{lin2023unlocking, fei2025nudging} suggest that disagreements between base and aligned models often arise over stylistic tokens, such as formatting tokens (\eg{} `\textbackslash n') or function words (\eg{} `and', `if'). 
% {Note that one word can be composed of multiple tokens. Hence, we only switch at word boundaries in case models have different tokenization (detailed implementation see \appref{app:router:addrule}).
% \mina{is ``formative tokens'' correct term? quick google search doesn't give straightforward examples. again, ``formative'' has a very specific meaning (serving to form something, especially having a profound and lasting influence on a person's development.) so I recommend that we pick a different term; e.g., format-related tokens, formatting tokens}
% \mina{also, note that now we're mixing token and word without properly explaining it}
% Key strategies implemented under this category include: 
We therefore implement:
% These strategies make decisions based on the semantic or syntactic properties of the generated text, motivated by observations that model disagreements often concern stylistic tokens \citep{lin2023unlocking, fei2025nudging}.
% \begin{enumerate}[wide, labelwidth=!, labelindent=0pt]  
\circone \textbf{\name{}-\textsc{Punc}} routes to the aligned model when its top-ranked token is either a punctuation or formatting token;
\circtwo \textbf{\name{}-\textsc{FC}} routes to the aligned model for function words to preserve stylistic coherence.
Content-based strategies are also applicable to black-box models, as they do not require
access to logits.
In practice, combining logit-based and content-based strategies yields the strongest
performance, as they rely complementary signals.
% Unless otherwise specified, we report results using representative single-strategy routers
% and their combinations (e.g., \texttt{-P} and \texttt{-P-PUNC}). 
For example, \name{}-\textsc{P-FC} prioritizes function-word routing (-\textsc{FC}) before falling back to probability-based decisions (-\textsc{P}).
% For example, \name{}-\textsc{P-FC} applies the function-word strategy (-\textsc{FC}) before falling back to the probability-based strategy (-\textsc{P}), as always routes to the aligned models when it generates function words, and applies maximum token probability to judge the rest of the case.
% Under the above motivations, 
In the next section, we evaluate a wide range of routing strategies and find that many variants are effective. For clarity, we present representative strategies in the main paper, with additional variants and implementation details in Appendix~\ref{app:router}.
We focus on the two categories of routing strategies and leave other routers (e.g., learned routers) for future work.\footnote{We exclude learned routers because (1) simple heuristics already yield substantial gains; (2) diversity is inherently multi-dimensional and cannot be captured by a single metric, leading to conflicting objectives and unstable training when jointly optimized (\Cref{sec:auto_eval}); and (3) diversity evaluation requires group-level sampling, which would substantially increase training cost and complexity.}

\section{Experimental Setup}
\label{sec:exp-setup}
\label{sec:experiments}
% \mina{good place to remind what the goal of the experiments is}
We design our experiments to empirically validate \name{}'s central goal: to optimize the diversity-quality trade-off. 
% through collaboration between a base and an aligned model at inference time. 
% Specifically, we aim to examine (\textit{i}) whether \name{} jointly improves diversity and quality across metrics and tasks, and 
% (\textit{ii}) how different routers perform.
% Together, these experiments validate the \name{}'s eff
% Together, these experiments validate the \name{}'s performance improvement and controllability (\ie{} the ability to adjust along the diversity-quality spectrum according to task or preference).
% \ych{outperformance reads really weird. }
% Together, these experiments test whether collaborative decoding can achieve controllable, high-quality, and genuinely diverse generation without retraining.

% \mina{the section title and this preamble make me think that the experimental results will be presented in this section, which is not the case. revise to avoid the confusion; (e.g., ``Experimental Setup'' might be a good title for section 4 as opposed to 4.1 since 4.2 is also part of setup)}

% \subsection{Datasets and Methods}

\fakeparagraph{Datasets.} We mainly evaluate \name{} across three representative open-ended generation tasks: {NoveltyBench} \citep{zhang2025noveltybench} for instruction following, {WildChat} \citep{zhao2024wildchat} for dialogue, and {Narrative-Discourse} \citep{tian2024large} for creative writing. Together, these datasets cover both short- and long-form open-ended generation across varying levels of task complexity. For full dataset details, we refer readers to \appref{app:datasets}.

% \paragraph{Inference Setup.} Our study focuses on group-level diversity. For each prompt, we generate a group of $n=10$ responses. Unless otherwise specified, sampling is performed with a temperature of $1.0$ and nucleus sampling (top-p) with $p=0.9$, applied consistently across all methods. These settings serve as our default inference configuration, with exceptions only for some baseline methods\footnote{Exceptions on baselines that inherently require alternative decoding strategies (\eg{} diverse beam search only compatible with greedy decoding) or when varying temperature for the single model's performance.}.

% \fakeparagraph{Scope.} Our study focuses on inference-time methods applied to frozen models. 
% The core setting of interest is efficient, single-pass left-to-right decoding, where each response is generated independently from a given prompt without cross-sample interaction, iterative revision, or prompt optimization (with a few baselines excepted serve as stronger comparisons). This reflects the practical regime in which most deployed LLMs operate. 

\fakeparagraph{Baselines.} We compare \name{} with inference-time methods across five categories:
\circone~ {Single-model:} a base model or an aligned model, 
% \mina{sounds like two models together; use ``a'' and ``or''}, 
each sampled at varying temperatures.  
\circtwo~ {Prompting-based:}
% \mina{consistency - ``-based'' or ``baseline(s)''?}
% (\textit{a}) 
in-context resampling \citep{meyerson2024language, zhang2025noveltybench}, where $n$ outputs are generated sequentially within a single dialogue;
% (\textit{b}) 
paraphrase prompting \citep{jiang2020can, zhang2025noveltybench}, where paraphrased variants of the same instruction are used to increase output diversity.  
\circthree~ {Decoding-based:} Diverse Beam Search \citep{vijayakumar2016diverse}, where a diversity penalty was added to the beam search algorithm.  
% \mina{why are we not describing this one like others?} 
\circfour~ {Ensemble-based:}
% (\textit{a}) 
response ensemble, where $n/2$ outputs are sampled from the base model and its aligned counterpart, and pooled into a single group; 
% (\textit{b}) 
logit ensemble, which merges the next-token probability distributions of the two models before sampling.
\circfive~ {Collaboration-based:} 
% (\textit{a}) 
\textsc{nudging} \citep{fei2025nudging}, where an aligned model selectively introduces tokens during a base model's decoding.
% \mina{add a pointer to section 3 where you introduce nudging}
% aligned-aligned \name{}  \mina{is this sixth category or parallel with nudging? clarify}, which applies our framework to two distinct aligned models within the same family. 
% The \textbf{inference setups} is at \appref{app:exp_setup}.
% Some methods (\ie{} diverse beam search, 
% \mina{which ones?} 
Note that diverse beam search, paraphrase prompting, and in-context resampling require additional computation; in-context resampling does not perform parallel sampling.\footnote{A wall-clock runtime comparison across all inference-time methods is provided in Appendix~\ref{app: implementation_details}.}
% of each outputs, where later sampling depend on earlier outputs.
% In-context resampling adds further overhead with cross-sample coordination, a \textit{sequential} process where later sampling knows previous outputs; in contrast, \name{} samples each output in \textit{parallel}.
% \mina{``In-context resampling ... parallel'' - cut? not the right place to describe limitations of baselines; it should be discussed in related work instead}
% And in-context resampling additionally involves cross-sample coordination, where later sampling knows what former outputs are, and \name{} samples each output in parallel.
These methods therefore provide more competitive reference points.
% \mina{what is this? is ours also cross-sample coordination?}
% , and therefore provide competitive reference points
The {inference setups} and {experimental scope} are provided in \appref{app:exp_setup}.
% \mina{consistency - \Cref{app:exp_setup}? apply everywhere}

\fakeparagraph{\name{}.}
Our experiments leverage two open-weight model pairs: \texttt{Llama-3-8B} and \texttt{Llama-3-8B-Instruct} \citep{grattafiori2024llama}, and \texttt{Olmo2-7B} and \texttt{Olmo2-7B-Instruct} \citep{olmo20242} as they are widely used in literature (\eg{} \citet{fei2025nudging}).
We implement the single-strategy routers (\eg{} \textsc{-P} which is based on  maximum token probability) from \Secref{sec:method:router} 
% \mina{give a concrete example with the name here too so that it contrasts with multi-strategy more easily} 
and denote multi-strategy routers as ``\textsc{-X-Y}'', where strategy Y precedes X (e.g., \textsc{-P-Punc}, \textsc{-P-FC}, and \textsc{-H-Punc}). 
These implementations serve as representative examples that demonstrate the possible design space 
% \mina{again... overloading the term. is this flexibility the same as the metric flexibility? \textbf{[important]} need to decide on the terms for clarity}
of the \name{} framework. \name{} framework works well 
% (do not generate gibberish) \mina{``(do not generate gibberish)'' - cut? this seems like a really low bar to be considered as ``works well''} 
with a wide range of routers.
% which readily accommodates a wide range of curated  \mina{?? unclear what it means}.
% Two more routers from \mina{weird to say routers from appendix; say we include two basic routers as baselines and then point people to Appendix for more details} 
We include two basic routers as baselines:
\circone~\textsc{-Rand} routes to the base model by random chance $\gamma$;
\circtwo~{-\textsc{Judge}} employs an external model to evaluate candidate tokens and makes a routing decision.
Refer to \appref{app:router:notation} for more details.

% \mina{overall feedback on Section 5.1:
% - it improved a lot, but it's still very dense and sometimes hard to follow and understand why certain decisions are made.
% - add justification/explanation when applicable
% - improve clarity in writing
% - (I get that we can't explain everything in detail here, but for the things that we decide to explain, we need to make it clear and understandable to people!)
% }

% \begin{wrapfigure}{r}{0.3\linewidth}
\begin{figure}
% \vspace{-55pt}
    \centering
    \includegraphics[width=0.55\linewidth]{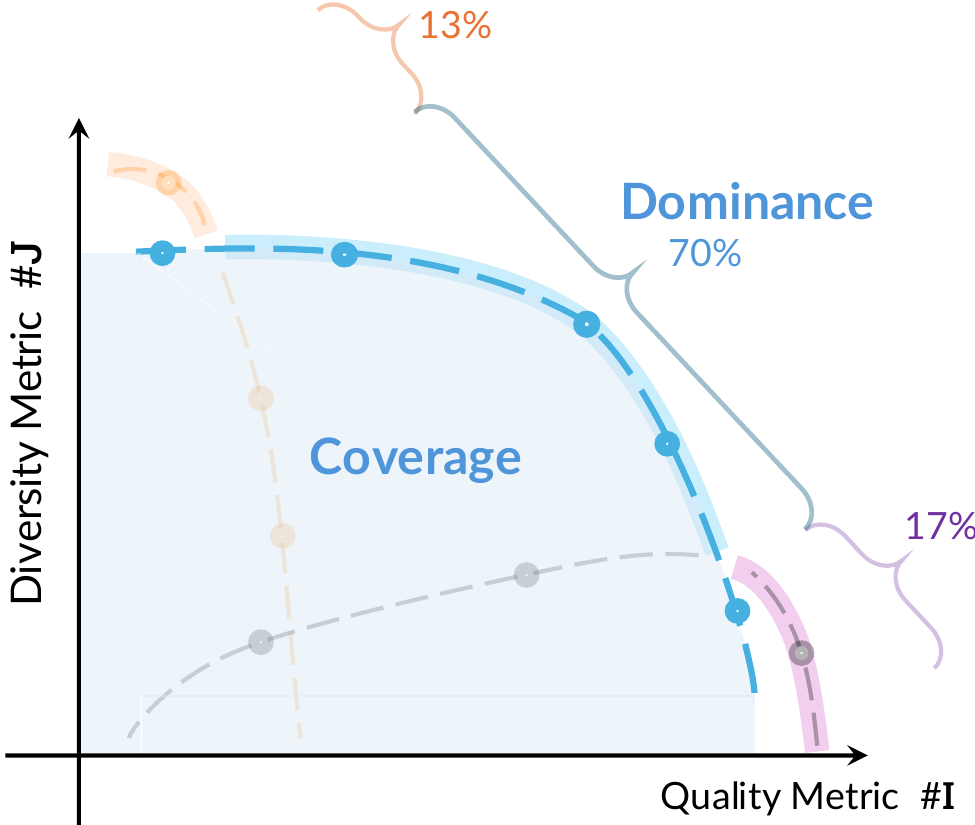}
    % \caption{Illustration of the indicators on a quality-diversity subspace. \textbf{Flex} is measured as the area under the curve (in blue); \textbf{Domn} is the length percentage of a method's contribution to the overall Pareto frontier (glowing curves).}
\caption{Illustration of the indicators on diversity-quality space: \textbf{Coverage}, the area under a method’s trade-off curve (blue shading for the blue method);
% \mina{terminology}
\textbf{Dominance}, the proportion of the global Pareto frontier (highlighted curves) contributed by the method.
% , indicating comparative strength.
% \mina{explain when one might care more about one over the other}
% In practice, methods with high Coverage are preferable when designing for general-purpose generation across diverse tasks, while high dominance is more desirable when selecting specialized methods to achieve the best trade-offs within specific regions of the space.
% In practice, high Coverage is preferable for general-purpose design, as it ensures a method offers a good trade-off across a wide range of tasks or user preferences. In contrast, high Dominance is desirable when selecting a specialized method to achieve the optimal trade-off within a specific, known target region of the space.
}
\vspace{-10pt}
    \label{fig:indicators}
\end{figure}

\begin{figure*}[t]
\centering
\vspace{6mm}
% Use a minipage for the table
\begin{minipage}[c]{0.5\textwidth}
    \centering
    \small
    \renewcommand{\arraystretch}{1.25}
    \setlength{\tabcolsep}{4pt}
    \resizebox{\linewidth}{!}{%
        \begin{tabular}{lcccccc}
            \toprule
            \textbf{Method} & \multicolumn{2}{c}{\textbf{Lexical}} & \multicolumn{2}{c}{\textbf{Semantic}} & \multicolumn{2}{c}{\textbf{Overall}} \\
             & \textit{Cov.} & \textit{Dom.} & \textit{Cov.} & \textit{Dom.} & \textit{Cov.} & \textit{Dom.} \\
            \midrule
            Base    & 0.098 & 12.7\% & 0.098 & 16.0\% & 0.098 & 14.3\% \\
            Aligned & 0.269 & \textbf{49.0\%} & 0.104 & 29.2\% & 0.186 & \textbf{39.0\%} \\
            Nudging & 0.276 & 9.3\%  & 0.247 & 9.9\%  & 0.261 & 9.6\%  \\
            Decoding        & - &  0.3\% & - &  0.3\% & - &  0.3\% \\
            Prompting (Best)  & - &  2.7\% & - &  2.2\% & - &  2.4\% \\
            Ensemble (Best)   & - &  1.1\% & - &  1.9\% & - &  1.5\% \\
            \name{} (Best) & \textbf{0.445} & 24.9\% & \textbf{0.360} & \textbf{40.5\%} & \textbf{0.403} & {32.7\%} \\
            \bottomrule
        \end{tabular}
    }
    % Use \captionof{table} to get a proper "Table X" caption
    \vspace{6pt}
    \captionof{table}{Averaged performance of all methods across all datasets and diversity–quality spaces.
    \name{} consistently outperforms baselines across all semantic and most lexical spaces, demonstrating stronger controllability and substantially improving the semantic diversity–quality trade-off.
    The overall gains, as driven primarily by improvements in semantic, suggest that \name{} produces more meaningful and content-level diversity, rather than superficial word-level changes, compared to other methods.
    See full results at \appref{app:detail_results}. 
    % For space-saving purposes, we present the best method's performance for each category as representative. ``Lexical'' denotes average results on lexical diversity-quality subspaces, ``Semantic'' denotes average results on semantic diversity-quality subspaces, and ``Overall'' averages all subspaces. Same denotation for tables hereafter.
    % \mina{this caption is hard to understand; suggestions:
    % - first sentence: describe the table in a self-contained manner; averaged results of what? what's the takeaway? say something like ``BACo outperforms baselines across all datasets and metrics.'' so that people can get the takeaway immediately
    % - the rest of the caption is too detailed; I think we can safely cut ``For space-saving purposes, we present the best method’s performance for each category as representative.''
    % - we can also cut ``“Lexical” denotes average results on lexical
% diversity-quality subspaces, “Semantic” denotes average
% results on semantic diversity-quality subspaces, and “Over-
% all” averages all subspaces.'' and instead explain it in Section 5.2
    % - put Best in parantheses, e.g., BACo (Best)
    % - interpret the results in a more intuitive manner (besides saying that it outperforms the baseline)
    % - and then simply add a pointer to appendix for the full results
    % }
    }
    \label{tab:avg_results}
\end{minipage}% <--- The '%' here is important to prevent extra horizontal space
\hspace{1em}
% Use a minipage for the figure
\begin{minipage}[c]{0.455\textwidth}
    \centering
    \vspace{-25pt}
    \includegraphics[width=\linewidth]{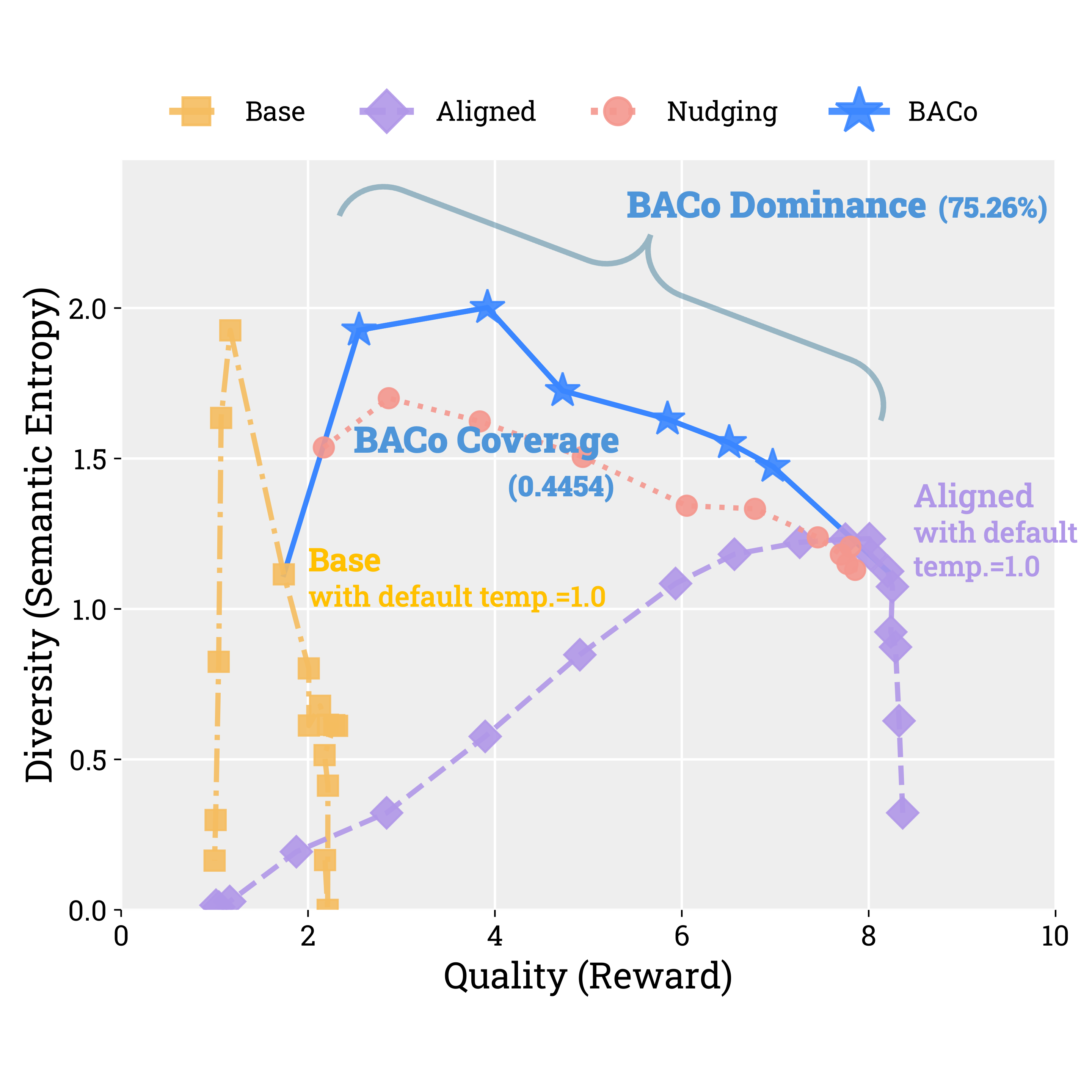}
    % A regular \caption works here, as it's inside a figure environment
    % \caption{A demonstration of \name{}'s outperformance on the diversity-quality space (one subspace x axis is reward, y axis is semantic entropy). \name{} (in blue) achieves more top-right region.}
    \vspace{-35pt}
    \caption{\name{}'s performance on one diversity-quality space (x: quality in terms of reward; y: diversity in terms of semantic entropy). Comparing with baselines, \name{} (blue curve) attains larger Coverage of the top-right region and contributes to most of the Dominance, indicating improvement and controllability on \problem{}.
    % \mina{
    % - some fonts are too small and hard to read
    % - use consistent font family and font size for labels across figures
    % }
    % \mina{this figure doesn't make it look like BACo is unarguably superior; (e.g., Aligned performs better on high reward region). since this figure is to demonstrate the point, create a hypothetical performance instead while reflecting the truth to some extent yet abstracting/smoothing out details so that it ihigh-reward understand.}
    % \yc{it is true that baco is performs better on high reward region, we admitted it as limitation. Fig 1B already serve as the figure show ideal superior, my intention for this figure is too show the real result. Two additional point is: 1) there might be self-evaluation bias, aligned model (reward model) prefer aligned model's output 2) the real quality gap between 8 and 9 is not that significant from human perspective.}
    }
    \label{fig:semantic-entropy}
\end{minipage}
% \vspace{-10pt}
\end{figure*}

% % prompt best: paraphrase
% % decoding
% % ensemble best

% % % \begin{wraptable}{r}{0.7\textwidth}
% % \begin{table}
% % \centering
% % \small
% % \renewcommand{\arraystretch}{1.2}
% % \begin{tabular}{lcccccc}
% % \toprule
% % \textbf{Method} & \multicolumn{2}{c}{\textbf{Lexical}} & \multicolumn{2}{c}{\textbf{Semantic}} & \multicolumn{2}{c}{\textbf{Overall}} \\
% %  & \textit{Flex} & \textit{Domn} & \textit{Flex} & \textit{Domn} & \textit{Flex} & \textit{Domn} \\
% % \midrule
% % Base & 0.098 & 12.7\% & 0.098 & 16.0\% & 0.098 & 14.3\% \\
% % Aligned & 0.269 & \textbf{47.7\%} & 0.104 & 28.7\% & 0.186 & \textbf{38.2\%} \\
% % Nudging & 0.276 & 9.3\% & 0.247 & 9.4\% & 0.261 & 9.3\% \\
% % Other baselines & - & 2.5\% & - & 3.7\% & - & 3.1\% \\
% % \midrule
% % \name{} best & \textbf{0.445} & 24.9\% & \textbf{0.360} & \textbf{39.2\%} & \textbf{0.403} & 32.1\% \\
% % \bottomrule
% % \end{tabular}
% % \caption{Averaged results across all datasets and all diversity-quality space. For space-saving purposes, we present the best router for each dataset as representative.}
% % \label{tab:avg_results}
% % \end{table}

\subsection{Automatic Evaluations}
\label{sec:auto_eval}
% Directly reporting raw scores on a large suite of lexical and semantic metrics would make comparison difficult and noisy. 

We next evaluate the \problem{} of each method using automatic metrics. 
Prior work has proposed a wide range of diversity evaluation that apply different lexical and semantic metrics,
% \footnote{The collection of diversity evaluation 
% \mina{is the repo about evaluation or diversity in general?}
% literature can be found at our \href{https://github.com/YichenZW/awesome-llm-diversity}{Awesome~LLM~Diversity} reading list.}
% \mina{in general, footnote should be placed after a punctuation; apply everywhere} 
reflecting different perspectives on language diversity.
Since our goal is to improve general diversity and quality rather than optimize for any specific metric, we adopt 11 established diversity metrics and 2 quality metrics which form 11 $\times$ 2 diversity-quality spaces, and then aggregate their results.
% To quantify \textit{controllability},

Moreover, we aim to quantify the \textit{controllability} of each method, \ie{} the ability to adjust along the diversity-quality
spectrum according to task or preference.
As shown earlier in \Cref{fig:intro}B, each method is not evaluated as an individual point in a fixed configuration, but a curve formed by a sequence of points, illustrating the diversity and quality performance in different configurations. 
Each curve illustrates the trade-off of a specific method.
To enable clear comparison, we apply two indicators from multi-objective optimization, {\texttt{Coverage (Cov.)}} and \texttt{Dominance (Dom.)}, to aggregate the curve-shaped performance across all spaces into numerical results (\Figref{fig:indicators}).\footnote{All curves in \Figref{fig:indicators} are illustrative only and do not correspond to the actual performance of any method.}
\textbf{Coverage (Cov.)} indicator quantifies the area under a method’s diversity-quality trade-off curve, following the hypervolume formulation used in multi-objective optimization. 
It measures how effectively a method traverses the diversity-quality spectrum as its control configuration varies.
A larger Coverage value indicates that the method maintains good performance across a wide range of regions on diversity-quality spaces, offering usability to more different tasks or preferences overall.
% For instance, coverage extending toward the high-diversity region supports ideation and creative applications, while coverage toward the high-quality region preserves reliability and factual accuracy.
% \ych{Confusing Terminology: Does ``coverage'' mean the same thing as the metric ``Coverage''?}
% For example, covering to high-diversity region enable ideation usages, while covering to high-quality region maintain the capbility for factuality. 
% offering more freedom to tune for different tasks or preferences.
% measures how effectively a method 
% \mina{a bit confusing to say a method but can't think of a better alternative} 
% traverses the diversity-quality trade-off curve as its control parameters vary. 
% We adapt hypervolume (HV), in terms of the area under a method’s trade-off curves.
% A larger area indicates general good performance across different trade-off balances \mina{undefined term ``balances''}, leading to better flexibility in control \mina{unclear trigram ``flexibility in control''}.
% % However, Flex does not capture whether a method is ever optimal \mina{unclear what this means} across different
% trade-off balances.
% implified to capture controllability across different balance points. 
% Higher Flex indicates broader coverage and more stable performance across parameter settings.  

\textbf{Dominance (Dom.)} 
indicator captures comparative optimality: whether and how often a method contributes to the global Pareto frontier among all methods.
 % captures whether a method is ever optimal relative to others. \mina{explain why we should care about this!}
We compute the global Pareto frontier across all methods and apply the C-metric \citep{zitzler1999evolutionary} to measure the portion of the frontier attributed to each method. 
% Higher Domn means that the method contributes more strongly to the global frontier. 
A higher Dominance value indicates that a method achieves uniquely strong trade-offs unattainable by others.

% The two indicators, Coverage and Dominance, capture different, and sometimes uncorrelated, properties. 
% From a practical standpoint, \texttt{Coverage} indicates how the method adapts to different tasks or user preferences for general-purpose systems; Conversely, \texttt{Dominance} is more relevant for systems designed for a specific task with a well-defined diversity-quality target, or where multiple methods are integrated and selected dynamically.

These indicators are instantiated under the 11 $\times$ 2 diversity-quality spaces. And we report the average across the spaces.

\textbf{Lexical diversity} spaces use diversity metrics such as Distinct-$n$ (Dist-$n$; \citealp{li2015diversity}), EAD-$n$ \citep{liu-etal-2022-rethinking}, and Self-BLEU \citep{montahaei2019jointly}.  
\textbf{Semantic diversity} spaces rely on diversity metrics such as embedding cosine dissimilarity \citep{kirk2023understanding}, Vendi Score (embedding) \citep{friedman2022vendi},  NLI diversity \citep{stasaski-hearst-2022-semantic}, and Semantic Entropy \citep{kuhn2023semantic}.\footnote{Each metric could include multiple variants. For example, Dist-1/2/3 for different n-gram, or cosine dissimilarity under different pretrained encoders. Each variance leads to a separate diversity-quality space.}
Since lexical and semantic metrics capture fundamentally different aspects of diversity, we analyze them separately in addition to reporting aggregated results. 
Increases in lexical diversity are relatively easy to achieve (for example, by raising the temperature), yet they mostly alter surface-level phrasing without changing meaning. 
In contrast, semantic diversity reflects deeper diversity in meaning, intent, and ideas, which is harder to elicit but more human-like and valuable in open-ended generation.
\textbf{Quality} metrics include (\textit{i}) perplexity under the aligned model, which reflects fluency and instruction following, and (\textit{ii}) reward modeling scores predicted by \texttt{Skywork-Reward-Gemma-2-27B} \citep{liu2024skywork}, the state-of-the-art model on RewardBench \citep{lambert2024rewardbench}.
These metrics are paired with diversity metrics to form subspaces.   
Finally, \textbf{overall} results average across all subspaces, yielding holistic method-level indicators. 
Hereafter, we use \emph{lexical} to denote average results on all lexical diversity-quality spaces, \emph{semantic} denotes semantic diversity-quality spaces, and \emph{overall} averages all spaces in every result table.
Full derivations and implementation details are in \appref{app:auto_eval:div}.

\section{Results}
\label{sec:results}

We first compare \name{} against strong inference-time baselines across datasets and
diversity-quality spaces (\Cref{ssec:main_result}). We then analyze the behavior of
different routing strategies in a controlled setting (\Cref{ssec:router_comparison}),
followed by ablations on model pairing (\Cref{ssec:ba_aa}). Finally, we evaluate
long-form diversity (\Cref{ssec:storyarc}) and validate automatic metrics with
human judgments (\Cref{ssec:human_eval}).

% In this section, we first compare \name{} against a wide range of baselines across multiple datasets and metrics (\Cref{ssec: main_result}). Next, we conduct router-level comparisons to analyze the effectiveness of different routing strategies (\Cref{ssec: router_comparison}, \Cref{ssec: ba_aa}). 
% % Short-form evaluations, such as on NoveltyBench, largely capture superficial lexical and semantic diversity, providing an incomplete picture. 
% We also demonstrate \name{}'s advantages on long-form diversity at the discourse level (\Cref{ssec: storyarc}).   
% % To demonstrate \name{}'s advantages in discourse-level diversity and creativity, 
% And we validate the results through human evaluation  (\Cref{ssec: human_eval}). 
% Lastly, we demonstrate \name{}'s performance on two additional verifiable tasks, instruction following and mathematical reasoning (\Cref{ssec: verifiable_tasks}).

% \subsection{\name{} Consistently Outperforms Baselines}
% \subsection{Main Results}
\subsection{Overall Performance}
\label{ssec:main_result}

\Tabref{tab:avg_results} summarizes the performance of \name{} and all baselines,
aggregated across datasets and metrics.
Overall, \name{} improves Coverage by \textbf{0.142} and achieves \textbf{32.7\%}
Dominance across all evaluation spaces.
Specifically, a Coverage improvement of 0.142 expands the achievable diversity-quality area by over 30\% relative to the strongest baseline.
And a Dominance of 32.7\% indicates that \name{} contributes nearly one-third of the global Pareto frontier over all baselines.
The advantage is particularly pronounced in semantic diversity, where Dominance
reaches \textbf{40.5\%} (see \Figref{fig:semantic-entropy}).\footnote{The aligned model's high lexical dominance arises from its high sampling temperature, which produces long, low-quality sequences that artificially inflate diversity scores while reducing controllability.}
On the NoveltyBench dataset, the gap further widens, with Coverage improving by
\textbf{0.274} and Dominance reaching \textbf{39.9\%}.
These trends are consistent across datasets and extend to the \texttt{Olmo2} model
family; full results are provided in Appendix~\Cref{app:detail_results}.
Qualitative output comparisons are shown in Appendix~\Cref{app:qual_examples}.

% \begin{wrapfigure}{r}{0.6\textwidth}
\begin{table}[t]
\small
% \vspace{-15pt}
\centering
    \renewcommand{\arraystretch}{1.2}
\resizebox{\linewidth}{!}{%
\begin{tabular}{lcccccc}
\toprule
\textbf{Routers} & \multicolumn{2}{c}{\textbf{Lexical}} & \multicolumn{2}{c}{\textbf{Semantic}} & \multicolumn{2}{c}{\textbf{Overall}} \\
 & \textit{Cov.} & \textit{Dom.} & \textit{Cov.} & \textit{Dom.} & \textit{Cov.} & \textit{Dom.} \\
\midrule
-\textsc{Rand} & 0.493 & 26.3\% & 0.409 & 17.0\% & 0.451 & 21.7\% \\
-\textsc{Judge} & 0.302 & 2.6\% & 0.254 & 0.6\% & 0.278 & 1.6\% \\
\midrule
-\textsc{P} & 0.433 & 4.8\% & 0.397 & 8.5\% & 0.415 & 6.7\% \\
-\textsc{FC} & 0.419 & 3.2\% & 0.382 & 4.7\% & 0.401 & 4.0\% \\
\midrule
-\textsc{P-Punc} &\textbf{0.495} & \textbf{30.7\%} &\textbf{0.452} & \textbf{31.3\%} & \textbf{0.474} & \textbf{31.0\%} \\
-\textsc{H-Punc} & 0.466 & 16.4\% & 0.427 & 18.6\% & 0.446 & 17.5\% \\
-\textsc{P-FC} & 0.435 & 16.0\% & 0.406 & 19.2\% & 0.421 & 17.6\% \\
\bottomrule
\end{tabular}
}
\vspace{7pt}
\captionof{table}{Averaged performance of routers within \name{} on NoveltyBench across all diversity–quality spaces.
The \textsc{-P-Punc} router achieves the best overall performance. 
While the random router (\textsc{-Rand}) attains moderately strong results, mainly from increased surface-level lexical diversity, its performance drops sharply on semantic metrics, confirming that unguided switching fails to produce meaningful diversity. 
In contrast, \textsc{-P-Punc} delivers the most balanced and consistent results across both lexical and semantic evaluations, showing combination of designed routing strategies leads to more meaningful diversity.
}
% \vspace{-5pt}
% showing that curated, information-driven 
% \mina{what do you mean by ``curated'' and ``information-driven''? this kind of vague use of random words strongly signals the use of chatgpt... stick with the terms that have been used before (e.g., logit-based/content-based routing, combination of routing strategies, etc.)} 
% routing leads to more meaningful diversity.}
% The \textsc{-Rand} as a strong baseline validates the \name{} frameworks' effectiveness without curated routing strategies. \mina{revise this part based on my comment on section 6.2} 
% \thh{What is the takeaway message from this table?} 
% To save space, we aggregate the results of the rest of the baselines into `Other baselines'.
% \vspace{-20pt}
\label{tab:noveltybench_all}
% \end{minipage}
% \vspace{-2pt}
% \end{wrapfigure}
\end{table}

\paragraph{Validation beyond open-ended evaluation metrics.}
To verify that these gains are not an artifact of open-ended evaluation metrics, we additionally evaluate \name{} on two verifiable benchmarks:
instruction following (IFEval) and mathematical reasoning (GSM8K).
Across both tasks, \name{} achieves higher diversity at matched quality or accuracy than the aligned baseline, confirming that the observed improvements are not only confined to open-ended generation.
% Detailed setups and results are reported in Appendix~\Cref{app:verifiable_tasks}.

\begin{figure}[h]
    \centering
    \includegraphics[width=0.95\linewidth]{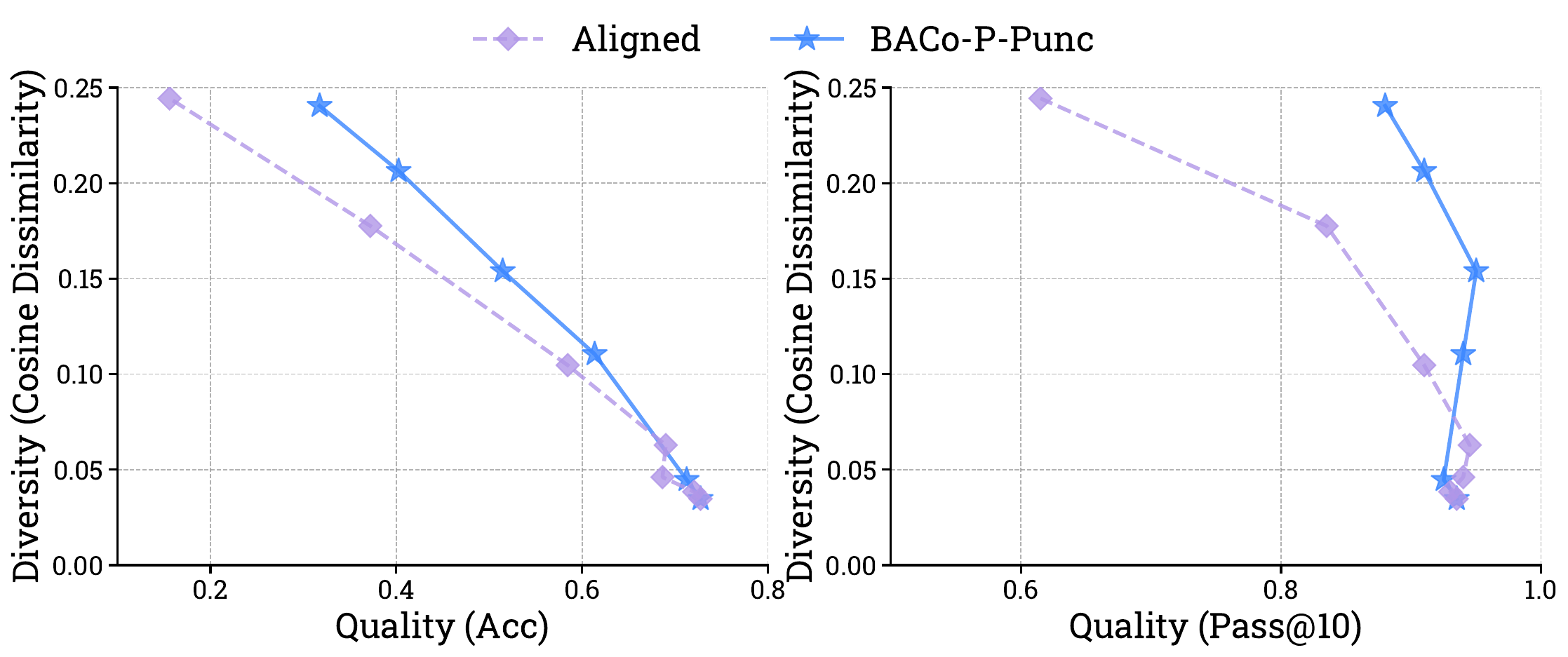}
\caption{Diversity-accuracy trade-off comparison on GSM8K. At comparable accuracy levels,
\name{} achieves higher diversity than the aligned baseline, indicating that the observed gains are not artifacts of open-ended evaluation metrics. 
% Refer full results in \Cref{fig:gsm8k_large_figure}.
}
% \vspace{-50pt}
\label{fig:gsm8k_small_figure}
\end{figure}

Notably, the qualitative behavior on GSM8K mirrors that on open-ended tasks:
naive temperature scaling degrades accuracy without delivering meaningful diversity, while base--aligned collaboration enables diversity improvements with accuracy maintained.
Complete results for both verifiable benchmarks are reported in
Appendix~\Cref{app:verifiable_tasks}.

\paragraph{Multi-turn tasks.}
To verify gains beyond single-turn settings, we evaluate on MT-Bench~\citep{zheng2023judgingllmasajudgemtbenchchatbot}.
\name{}-\textsc{P-Punc} substantially outperforms the aligned baseline
(Coverage: 0.681 vs. 0.320; Dominance: 72.8\% vs. 27.2\%) consistently
across diversity and quality metrics; full results in \Cref{app:verifiable_tasks}.

% \paragraph{Validation on multi-turn tasks.}
% To verify the gain beyond single-turn settings, we additionally evaluate on MTBench \citep{zheng2023judgingllmasajudgemtbenchchatbot}. \name{}-\textsc{P-Punc} substantially outperforms the aligned baseline (Coverage 0.681 vs.\ 0.320; Dominance 72.8\% vs.\ 27.2\%) consistently across diversity and quality metrics; full results in Appendix~\Cref{app:verifiable_tasks}.

\paragraph{Examination on the distribution of repetition.}
We additionally evaluate \name{} on Artificial Hivemind~\citep{jiang2025artificialhivemindopenendedhomogeneity}, which measures intra-model
repetition on open-ended chat queries. Following its setting, we compare \name{} against Top-$p$ and Min-$p$~\citep{nguyen2025turningheatminpsampling} on \texttt{Llama-3.1-70B-Instruct}. 

\begin{wrapfigure}{r}{0.56\linewidth}
    \vspace{-18pt}
    \centering
    \includegraphics[width=\linewidth]{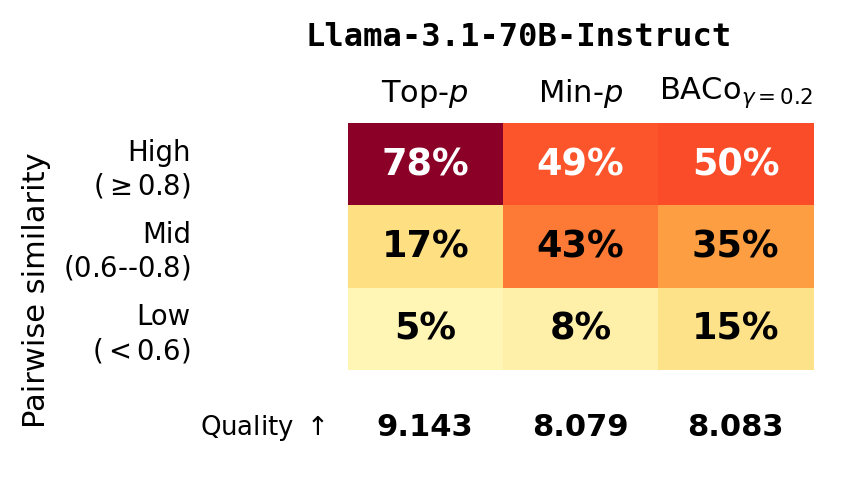}
    \caption{Pairwise intra-model repetition distribution under Artificial Hivemind. \name{}'s generation falls more in lower-similarity bins (bottom rows).}
    \label{fig:hivemind}
    % \vspace{-23pt}
\end{wrapfigure}

With \name{}'s threshold selected to match Min-$p$'s quality (avg.\ reward $8.08$; $\gamma{=}0.2$, \textsc{-P-Punc}), its repetition distribution is flatter (\Cref{fig:hivemind}), placing more (15\%) generations in the genuinely diverse regime (similarity $<$ 0.6), vs.\ 8\% for Min-$p$ and 5\% for Top-$p$.
This suggests \name{} pushes diversity beyond the limit reachable by sampling within a single aligned model.

% \Tabref{tab:avg_results} summarizes the performance of  \name{} and baselines, aggregated across datasets and metrics. 
% Overall, \name{} improves Coverage by \textbf{0.142} and achieves \textbf{32.7\%} Dominance. The advantage is particularly pronounced in semantic diversity, where its Dominance rises to \textbf{40.5\%} (\Figref{fig:semantic-entropy} as an example).\footnote{The aligned model's high lexical dominance arises from its high sampling temperature, which produces long, low-quality sequences that artificially inflate diversity scores while reducing controllability.} On the NoveltyBench dataset, the performance gap becomes even wider, with Coverage improving by \textbf{0.274} and Dominance reaching \textbf{39.9\%}. These gains are consistent across all datasets and extend to the \texttt{Olmo2} model family; see \appref{app:detail_results} 
% % and \appref{app:detail_results:nb} 
% for full results. See the output comparison examples at \appref{app:qual_examples}. 
% \mina{love these examples! the examples really help readers get a sense of what model outputs are like. 
% on the other hand, I think it's still hard to understand what these numbers really mean; e.g., coverage improvement by 0.142 vs. 0.274 and stuff. it says it improved, which is great, but it's still hard to know what that improvement actually looks like.}

\subsection{Router Performance Comparison}
\label{ssec:router_comparison}
We analyze routers on NoveltyBench, which provides a representative yet computationally efficient setting for controlled comparison.

\paragraph{Sanity-check routers (-\textsc{Rand}, -\textsc{Judge}).}
At first glance, \textsc{-Rand} appears competitive on aggregate, but its gains concentrate almost entirely on lexical metrics (\eg{} Dist-$n$) and collapse on semantic metrics such as Semantic Entropy (\(0\%\) Dominance; Appendix~\Cref{app:detail_results}).
This is consistent with the observation that random or nonsensical text can inflate lexical diversity without producing semantically meaningful variation; unguided switching injects surface-level randomness rather than meaningful semantic diversity.
By contrast, principled routers such as \textsc{-P-Punc} achieve clear Dominance on both lexical and semantic spaces (Table~\ref{tab:noveltybench_all}).
We further consider \textsc{-Judge}, a prompt-based router inspired by multi-agent systems \citep{talebirad2023multi}.
Despite extensive prompt engineering\footnote{Including a step-by-step decision pipeline, curated heuristic rules, and few-shot examples with rationales (prompts are in \Tabref{tab:judge-model-prompt}).}, it consistently underperforms simpler heuristic routers while incurring substantially higher computational cost.
\begin{table}[t]
\centering
\small
% \vspace{-14pt}
\resizebox{\linewidth}{!}{%
\renewcommand{\arraystretch}{1.1}
\begin{tabular}{l cc cc cc }
\toprule
\textbf{Method} & \multicolumn{2}{c}{\textbf{Lexical}}& \multicolumn{2}{c}{\textbf{Semantic}} & \multicolumn{2}{c}{\textbf{Overall}} \\
 & \textit{Cov.} & \textit{Dom.} & \textit{Cov.} & \textit{Dom.} & \textit{Cov.} & \textit{Dom.} \\
\midrule
Base & 0.142 & 9.8\% & 0.142 & 13.2\% & 0.142 & 11.5\%  \\
Aligned & 0.273 & \textbf{43.8\%} & 0.128 & 19.9\% & 0.200 & 31.8\%\\
\midrule
\textsc{AACo} & 0.022 & 4.5\% & 0.006 & 7.0\% &  0.014 & 5.7\% \\
\name{} & \textbf{0.495} & 42.0\% & \textbf{0.452} & \textbf{59.8\%} & \textbf{0.474 }& \textbf{50.9\%} \\
\toprule
\end{tabular}
}
\vspace{4pt}
\caption{Comparison of base-aligned and aligned-aligned collaboration (denoted as ``\textsc{AACo}'') on NoveltyBench across all diversity–quality spaces.
% While both settings leverage collaboration, 
\textsc{AACo} yields little improvement on the diversity–quality trade-off compared with \name{}, particularly in semantic diversity. The results demonstrate the necessity of involving a base and aligned model.
% \mina{say ``Aligned-Aligned'' instead? ``AACo'' makes it look like it's one of our methods}
% 
% In contrast,  achieves substantially higher performance, particularly in semantic diversity. 
% These results highlight the limited diversity that arises from collaborating between two aligned models.
% underscoring that complementarity between less-aligned and more-aligned models is crucial for achieving both high diversity and high quality.
% In the experiment, \textsc{AACo} uses the same collaborative framework and router as \name{} but replaces the base model with another aligned model. 
% denotes aligned-aligned collaboration, which uses the same framework and router as \name{} but with two aligned models instead.}
}
% \vspace{-5mm}
\label{tab:novelbench-ab-vs-aa}
% \vspace{-12mm}
% \end{wraptable}
\end{table}

% \begin{wraptable}{r}{0.6\textwidth}
% \centering
% \small
% \renewcommand{\arraystretch}{1.1}
% \begin{tabular}{l cc cc cc }
% \toprule
% \textbf{Method} & \multicolumn{2}{c}{\textbf{Lexical}}& \multicolumn{2}{c}{\textbf{Semantic}} & \multicolumn{2}{c}{\textbf{Overall}} \\
%  & \textit{Flex} & \textit{Domn} & \textit{Flex} & \textit{Domn} & \textit{Flex} & \textit{Domn} \\
% \midrule

% \textsc{AACo} & 0.022 & 12.8\% & 0.006 & 12.4\% & 0.000 | 0.0%
% % 0.002 & 1.6\% & 0.002 & 2.7\% & 0.002 & 2.2\% \\
%  \name{} & \textbf{0.360} & \textbf{80.5\%} & \textbf{0.305} & \textbf{55.1\%} & \textbf{0.333} & \textbf{67.8\%} \\

% % Base & 0.142 & 9.8\% & 0.142 & 13.2\% & 0.142 & 11.5\%  \\
% % Aligned & 0.273 & \textbf{43.8\%} & 0.128 & 19.9\% & 0.200 & 31.8\%\\
% % \midrule
% % \textsc{AACo} & 0.022 & 4.5\% & 0.006 & 7.0\% &  0.014 & 5.7\% \\
% % \name{} & \textbf{0.495} & 42.0\% & \textbf{0.452} & \textbf{59.8\%} & \textbf{0.474 }& \textbf{50.9\%} \\
% \toprule
% \end{tabular}
% \caption{Comparison of methods across diversity metrics on NoveltyBench. \textsc{AACo} denotes Aligned-Aligned Collaboration, which uses the same framework and router as \name{} but with two aligned models instead.}
% \label{tab:novelbench-ab-vs-aa}
% \end{wraptable}
\begin{figure*}[h] 
\centering
\begin{minipage}[h]{0.48\textwidth}
  \centering
  \includegraphics[width=\linewidth]{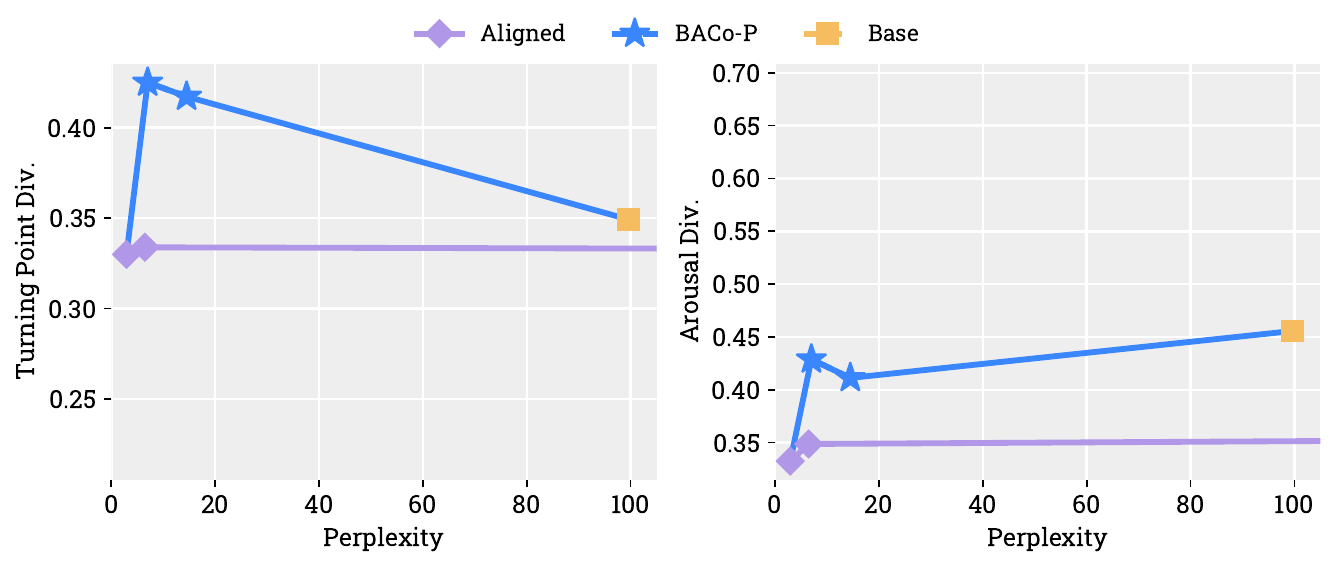}
    % \caption{Results of narrative discourse evaluation. The x-axis measures quality (perplexity; lower is better), and the y-axis measures discourse-level diversity (turning-point or arousal; higher is better). Better methods lie toward the top-left region.}
    \caption{
    Comparison of \name{}'s and baselines' discourse-level diversity-quality trade-off curve on Narrative Discourse. 
    \name{} obtains a larger Coverage, achievable in the high-diversity, high-quality region (top-left). The results demonstrate it has richer discourse-level diversity without sacrificing quality largely, compared with baselines.
    The x-axis is quality (perplexity; lower is better), and the y-axis is discourse-level diversity, either turning-point diversity (left figure) or arousal diversity (right figure) (higher is better). 
    % Methods closer to the top-left corner achieve better diversity–quality trade-offs. 
    }
  % \caption{Result of narrative discourse evaluation. X-axis is perplexity measure quality (smaller better); Y-axis is turning point or arousal diversity (larger better).}
  \label{fig:narrative-dis}
\end{minipage}\hfill
\begin{minipage}[t]{0.48\textwidth}
\vspace{-8\baselineskip}
  \centering
  \small
  \renewcommand{\arraystretch}{1.25} 

  \setlength{\tabcolsep}{4pt}          
  
  \resizebox{0.88\linewidth}{!}{%
  \begin{tabular}{@{}llcc|cc@{}}       
    \toprule
             & \textbf{} & \multicolumn{2}{c|}{\textbf{NoveltyBench}} & \multicolumn{2}{c}{\textbf{WildChat}} \\
    \midrule
    \textbf{\textit{}} & \textbf{} & Aligned & \name{} & Aligned & \name{} \\
    \midrule
    \multicolumn{2}{c}{\textit{Quality}}  & 2.83 & \textbf{4.04} & 3.44 & \textbf{3.83} \\
    \midrule
     & \textit{Overall} & 21.0\% & \textbf{79.0\%} & 36.1\% & \textbf{63.9\%} \\
    \textit{Diversity} & \textit{Format} & 25.4\% & \textbf{74.6\%} & 26.8\% & \textbf{73.2\%} \\
     & \textit{Content} & 22.9\% & \textbf{77.1\%} & 41.6\% & \textbf{58.4\%} \\
    \midrule
    \multicolumn{2}{c}{\textit{Creativity}}  & 20.4\% & \textbf{79.6\%} & 38.2\% & \textbf{61.8\%} \\
    \bottomrule
  \end{tabular}%
  } % end \resizebox
\vspace{7pt}
\captionof{table}{Human evaluation comparing \name{} (best variant, \textsc{-P-Punc}) with the aligned model baseline on NoveltyBench and WildChat.
\name{} is consistently preferred by human judges across all aspects, demonstrating that it produces outputs that are not only more human-perceived diverse but also higher in quality and creativity.
Scores of quality are on a 1–5 Likert scale, and others indicate the pairwise win rate. }
% \vspace{-100pt}
    % \captionof{table}{Human evaluation comparing \name{} (best variant, \textsc{-P-Punc}) with the aligned model baseline. \name{} consistently outperforms across assessed dimensions.}
  % \captionof{table}{Human evaluation between aligned model baseline and \name{} best (\ie{} -\textsc{P-Punc}). \name{} consistently outperforms. 
  % For fair comparison, we select the parameters of both methods to match their quality score in automatic evaluation.
  \label{tab:human_eval}
\end{minipage}
% \vspace{-5mm}
\end{figure*}

\paragraph{Simple routers (-\textsc{P}, -\textsc{FC}).}
Single-strategy routers based on uncertainty (\textsc{-P}) or linguistic cues (\textsc{-FC}) trade weaker aggregate performance for clear gains in semantic diversity.
Both outperform \textsc{-Rand} on metrics such as Cosine Dissimilarity and Semantic Entropy, demonstrating that informed routing is critical for meaningful diversity.

\paragraph{Combining strategies yields the strongest performance.}
Combining complementary strategies proves most effective.
In particular, \textsc{-P-Punc} achieves the best overall controllability, with high
Coverage and Dominance across both lexical and semantic spaces.
Other combinations (\textsc{-H-Punc}, \textsc{-P-FC}) capture additional regions of
the Pareto frontier (17.5\% and 17.6\% Dom.), confirming that compositional routing provides complementary
benefits.

Overall, \name{} with most of the routers outperforms the baselines. We therefore emphasize that the contribution of \name{} is the \emph{framework} of token-level base-aligned collaboration, not any single router.
% Token-level base-aligned collaboration consistently optimizes the trade-off, 
And the framework is robust to router choice without heavy tuning.

% \paragraph{Combining strategies achieves the best overall performance (-\textsc{P-Punc}, -\textsc{H-Punc}, and -\textsc{P-FC}).}
% Combining strategies proves to be most effective in our experiments. \textbf{The \textsc{-P-Punc} router achieves the best overall performance}, demonstrating high controllability (both Coverage and Dominance) across both lexical and semantic spaces. Other combinations like \textsc{-H-Punc} and \textsc{-P-FC}, while not as strong overall, are still valuable as they capture unique regions of the global Pareto frontier (17.5\% and 17.6\% Dom.). 
% This confirms that combining strategies yields complementary strengths, further advancing the trade-off frontier.

% \subsection{Base-Aligned versus Aligned-Aligned Collaboration}  

% We validate that base-aligned collaboration (\name{}) is more effective and efficient than aligned-aligned collaboration in mitigating \problem{}. 
% We validate it empirically on NoveltyBench.
% We use \textsc{-P-Punc} strategy on model pairs within the same family: \circone aligned-aligned collaboration between \texttt{Llama-3-8B-Instruct} and \texttt{Llama-3.2-11B-Vision}; \circtwo base-aligned collaboration (\name{}) between \texttt{Llama-3-8B} and \texttt{Llama-3-8B- Instruct}.
% Results are shown in \Tabref{tab:novelbench-ab-vs-aa}, where \textbf{base-aligned collaboration significantly outperforms aligned-aligned collaboration in both controllability and dominance}, achieving a better \problem{}. 

\subsection{Base-Aligned vs.\ Aligned-Aligned Collaboration}
% \subsection{Base-Aligned Collaboration Outperforms Aligned-Aligned}
\label{ssec:ba_aa}

We compare base-aligned collaboration (\name{}, \texttt{Llama-3 -8B} family) with aligned-aligned collaboration (\texttt{Llama-3-8B} and \texttt{Llama-3.2-11B})
under identical router (\textsc{-P-Punc}).
On NoveltyBench, base-aligned collaboration significantly outperforms the aligned-aligned setup in both Coverage and Dominance.
These results highlight the limited diversity achievable when collaborating between two aligned models and support our hypothesis that complementarity between less-aligned
and more-aligned models are essential for jointly optimizing diversity and quality.

\subsection{Long-Form Diversity Evaluation}
\label{ssec:storyarc}

To assess diversity beyond short-form outputs, we evaluate long-form creative writing
using a discourse-level framework following \citet{tian2024large}.
Rather than relying on surface lexical or single-vector semantic metrics, this
evaluation captures narrative variation through plot structure and affective dynamics (detailed setups are provided in \appref{app:storyarc}).

As shown in \Figref{fig:narrative-dis}, \name{} achieves substantially higher
turning-point and arousal diversity than the aligned baseline at comparable quality,
demonstrating that base--aligned collaboration extends effectively to long-form
generation.
Example outputs are provided in \Tabref{tab:case_storyarc}.

\subsection{Human Evaluation}
\label{ssec:human_eval}

To complement automatic metrics, we conduct a three-phase human evaluation assessing quality, group-level diversity, and creativity.
Four expert annotators evaluate outputs on NoveltyBench and WildChat.

As shown in \Cref{tab:human_eval}, despite near-identical automatic quality
scores (aligned: 5.93; \name{}: 5.85), human raters assign \name{} substantially
higher quality ratings (4.04 vs.\ 2.83 on NoveltyBench; 3.83 vs.\ 3.44 on
WildChat), with strong inter-rater agreement (Pearson $r{=}0.816$,
ICC$(2,k){=}0.907$).
Meanwhile, \name{} achieves significant large diversity win rates across both format and content
sub-dimensions (Fleiss' $\kappa{=}0.268$ overall), confirming that \name{} produces human-perceivable diversity at both levels. Full protocols and annotation examples are provided in Appendix~\Cref{app:human_eval}. 

% Despite comparable automatic quality, human evaluators consistently prefer \name{}
% over the aligned baseline, with significant gains in both format and content diversity,
% as well as overall perceived creativity.
\paragraph{From diversity-quality trade-off to creativity.}
Building on \citet{jaarsveld2012creative}, creativity requires both
\textit{divergent thinking}, supported by group-level output diversity, and
\textit{convergent thinking}, underpinned by per-output quality.
We preliminarily test whether an improved diversity-quality trade-off translates
into higher human-perceived creativity using \name{}.
As shown in \Cref{tab:human_eval}, \name{}'s outputs are judged most creative in
{79.6\%} of NoveltyBench prompts and {61.8\%} of WildChat prompts
(Fleiss' $\kappa{=}0.485$), suggesting that easing the diversity-quality trade-off yields outputs humans find genuinely more creative.
We leave a larger-scale and deeper investigation of this connection to future work.
% Overall, these results demonstrate that base-aligned collaboration provides a
% simple yet robust mechanism for expanding the diversity--quality frontier across
% settings; further analysis is in \appref{sec:analysis}.

\section{Analysis and Discussion}
\label{sec:analysis}

\subsection{Contribution Distribution and Switching Frequency}
\label{sec:anal:distribution}

For \name{} with the best router (\textsc{-P-Punc}), we see a consist pattern that {base-model
contribution and switching frequency are high at the start of generation and
decrease over time} across all three tasks (Figures~A9-A11 in \appref{app:analysis}).
We hypothesize this reflects increasing predictive confidence as context grows,
which naturally aligns with, e.g., in creative writing, early tokens allow more divergence
(\eg{} introducing characters) while later tokens demand coherence. However, the pattern may be
less suitable for list-structured tasks that benefit from uniform exploration level over time.
Dynamic, position-aware thresholds are a promising mitigation. %, which we leave for future.

\subsection{Failure Mode: Inherent Early Stopping}
\label{sec:earlystop}

When the router is tuned more aggressively toward diversity, the system exhibits
a higher tendency to terminate early. 
One model emits an end-of-sequence token
prematurely (examples in \appref{app:analysis}).
This \emph{inherent early stopping} is emergent rather than deliberate, unlike
early-exit mechanisms in reasoning
systems~\citep{ding2025dynamicparalleltreesearch}.
We take a two-sided view of this trend: it can risk truncating valid continuations, but also acts as a safeguard against incoherent repetition or off-topic continuation.
Truncated outputs are easily detectable by length, allowing a simple restart strategy as mitigation with marginal cost overhead.

\subsection{Future Work}
\label{sec:future_work}

% We envision \name{} as a preliminary step toward a ``breadth thinking'' mode for
% LLMs, complementing deep
% thinking~\citep{OpenAI2025IntroducingDeepResearch,DeepSeekAI2025DeepSeekR1IR}, where
% a model explores a wide, validated space of perspectives rather than converging
% on a single output.
% For open-ended tasks, breadth thinking offers a way to help humans break out of
% their ``information cocoons''~\citep{piao2023human}, expand ideation space, and
% think beyond conventional boundaries.
% \name{} is controllable, inference-time only, and deployable without additional
% post-training; it can also serve as a plug-in for broader systems, such as an
% ideation agent in multi-agent
% collaboration~\citep{siddiqui2025script,2025arXiv250614758C,song2025outcomebasedexplorationllmreasoning}.
% A natural direction is to design task-aware routers that adaptively learn routing
% strategies from data or rewards.

% The controllability of \name{} also opens opportunities for user-facing
% interfaces, such as a slider that let users shift along the diversity-quality
% spectrum, making breadth thinking practically useful beyond research benchmarks, which can serve as a more effective tune on how diverse or indiverse people what model to be additional to tuning temperture which harm quality easily.
% % More broadly, \name{} has potential as a general paradigm for inference-time collaboration across objectives beyond diversity, including
% % privacy~\citep{chua2024mind} and controllable
% % safety~\citep{zhang2025controllablesafetyalignmentinferencetime}.

We envision \name{} as a preliminary step toward a ``breadth thinking'' mode for
LLMs, complementing deep
thinking~\citep{OpenAI2025IntroducingDeepResearch,DeepSeekAI2025DeepSeekR1IR}, where
a model explores a wide, validated space of perspectives rather than converging
on a single output.
For open-ended tasks, breadth thinking offers a way to help humans break out of
their ``information cocoons''~\citep{piao2023human}, expand ideation space, and
think beyond conventional boundaries.
\name{} is controllable, inference-time only, and deployable without additional
post-training; it can also serve as a plug-in for broader systems, such as an
ideation agent in multi-agent
collaboration~\citep{siddiqui2025script,2025arXiv250614758C,song2025outcomebasedexplorationllmreasoning}.
% A natural direction is to design task-aware routers that adaptively learn routing
% strategies from data or rewards.
The controllability of \name{} also opens opportunities for user-facing
interfaces, such as a slider that lets users shift along the diversity-quality
spectrum, offering a more principled way to control output diversity than temperature scaling, which degrades might quality.

\section*{Acknowledgment}
We thank the anonymous reviewers and area chair for their constructive comments, which helped improve this work. We also thank the members of the Communication and Intelligence (C\&I) group at the University of Chicago for valuable discussions and proofreading. We thank Peter West (Stanford; University of British Columbia) for helpful discussions on the idea. We are grateful to the UChicago PhD Writing Program for their feedback on polishing the writing.
This material is based upon work supported by the ASA(ALT) SBIR CCoE under Contract No. W51701-25-C-0061.

\section*{Impact Statement}
This work does not involve any sensitive personal data or personally identifiable information in the human study. 
All datasets used are publicly available research benchmarks released under appropriate licenses. 
Our experiments focus on inference time with existing LLMs without additional fine-tuning, and thus pose minimal risks of memorization or privacy leakage beyond those already associated with the underlying models.  
While our method aims to improve diversity in model outputs, we acknowledge the possibility that enhanced diversity might amplify harmful generations if safeguards are not in place. 
This is a shared risk for most diversity-promoting technologies.
We therefore emphasize that \name{} is complementary to safety alignment and should be deployed with standard content-filtering or moderation practices. 
Importantly, \name{} is designed for open-ended generation, rather than high-stakes or safety-critical domains.  
All authors have read and adhered to the ICML Code of Ethics.

\bibliography{custom}

@article{ouyang2022training,
  title={Training language models to follow instructions with human feedback},
  author={Ouyang, Long and Wu, Jeffrey and Jiang, Xu and Almeida, Diogo and Wainwright, Carroll and Mishkin, Pamela and Zhang, Chong and Agarwal, Sandhini and Slama, Katarina and Ray, Alex and others},
  journal={Advances in neural information processing systems},
  volume={35},
  pages={27730--27744},
  year={2022}
}

@article{zhang2025noveltybench,
  title={NoveltyBench: Evaluating Language Models for Humanlike Diversity},
  author={Zhang, Yiming and Diddee, Harshita and Holm, Susan and Liu, Hanchen and Liu, Xinyue and Samuel, Vinay and Wang, Barry and Ippolito, Daphne},
  journal={arXiv preprint arXiv:2504.05228},
  year={2025}
}

@article{kirk2023understanding,
  title={Understanding the effects of rlhf on llm generalisation and diversity},
  author={Kirk, Robert and Mediratta, Ishita and Nalmpantis, Christoforos and Luketina, Jelena and Hambro, Eric and Grefenstette, Edward and Raileanu, Roberta},
  journal={arXiv preprint arXiv:2310.06452},
  year={2023}
}

@article{west2025base,
  title={Base models beat aligned models at randomness and creativity},
  author={West, Peter and Potts, Christopher},
  journal={arXiv preprint arXiv:2505.00047},
  year={2025}
}

@article{Shumailov2024,
  author    = {Ilia Shumailov and Zakhar Shumaylov and Yiren Zhao and Nicolas Papernot and Ross Anderson and Yarin Gal},
  title     = {AI models collapse when trained on recursively generated data},
  journal   = {Nature},
  year      = {2024},
  volume    = {631},
  number    = {8022},
  pages     = {755--759},
  doi       = {10.1038/s41586-024-07566-y},
  url       = {https://doi.org/10.1038/s41586-024-07566-y}
}

@article{hamilton2024detecting,
  title={Detecting mode collapse in language models via narration},
  author={Hamilton, Sil},
  journal={arXiv preprint arXiv:2402.04477},
  year={2024}
}

@article{cui2025entropy,
  title={The entropy mechanism of reinforcement learning for reasoning language models},
  author={Cui, Ganqu and Zhang, Yuchen and Chen, Jiacheng and Yuan, Lifan and Wang, Zhi and Zuo, Yuxin and Li, Haozhan and Fan, Yuchen and Chen, Huayu and Chen, Weize and others},
  journal={arXiv preprint arXiv:2505.22617},
  year={2025}
}

@inproceedings{chakrabarty2025can,
  title={Can ai writing be salvaged? mitigating idiosyncrasies and improving human-ai alignment in the writing process through edits},
  author={Chakrabarty, Tuhin and Laban, Philippe and Wu, Chien-Sheng},
  booktitle={Proceedings of the 2025 CHI Conference on Human Factors in Computing Systems},
  pages={1--33},
  year={2025}
}

@article{zhang2024lists,
  title={From lists to emojis: How format bias affects model alignment},
  author={Zhang, Xuanchang and Xiong, Wei and Chen, Lichang and Zhou, Tianyi and Huang, Heng and Zhang, Tong},
  journal={arXiv preprint arXiv:2409.11704},
  year={2024}
}

@article{chen2024not,
  title={Do not think that much for 2+ 3=? on the overthinking of o1-like llms},
  author={Chen, Xingyu and Xu, Jiahao and Liang, Tian and He, Zhiwei and Pang, Jianhui and Yu, Dian and Song, Linfeng and Liu, Qiuzhi and Zhou, Mengfei and Zhang, Zhuosheng and others},
  journal={arXiv preprint arXiv:2412.21187},
  year={2024}
}

@article{ding2025dynamic,
  title={Dynamic parallel tree search for efficient llm reasoning},
  author={Ding, Yifu and Jiang, Wentao and Liu, Shunyu and Jing, Yongcheng and Guo, Jinyang and Wang, Yingjie and Zhang, Jing and Wang, Zengmao and Liu, Ziwei and Du, Bo and others},
  journal={arXiv preprint arXiv:2502.16235},
  year={2025}
}

@article{kim2024evaluating,
  title={Evaluating language models as synthetic data generators},
  author={Kim, Seungone and Suk, Juyoung and Yue, Xiang and Viswanathan, Vijay and Lee, Seongyun and Wang, Yizhong and Gashteovski, Kiril and Lawrence, Carolin and Welleck, Sean and Neubig, Graham},
  journal={arXiv preprint arXiv:2412.03679},
  year={2024}
}

@article{yang2025measuring,
  title={Measuring data diversity for instruction tuning: A systematic analysis and a reliable metric},
  author={Yang, Yuming and Nan, Yang and Ye, Junjie and Dou, Shihan and Wang, Xiao and Li, Shuo and Lv, Huijie and Wu, Mingqi and Gui, Tao and Zhang, Qi and others},
  journal={arXiv preprint arXiv:2502.17184},
  year={2025}
}

@misc{Xiao2025SubmodularOptimization,
  author       = {Han Xiao},
  title        = {Submodular Optimization for Diverse Query Generation in DeepResearch},
  year         = {2025},
  month        = jul,
  day          = {4},
  howpublished = {Jina AI Tech Blog},
  url          = {https://jina.ai/news/submodular-optimization-for-diverse-query-generation-in-deepresearch/},
  note         = {13 minutes read}
}

@article{padmakumar2023does,
  title={Does writing with language models reduce content diversity?},
  author={Padmakumar, Vishakh and He, He},
  journal={arXiv preprint arXiv:2309.05196},
  year={2023}
}

@inproceedings{ashkinaze2025ai,
  title={How AI ideas affect the creativity, diversity, and evolution of human ideas: evidence from a large, dynamic experiment},
  author={Ashkinaze, Joshua and Mendelsohn, Julia and Qiwei, Li and Budak, Ceren and Gilbert, Eric},
  booktitle={Proceedings of the ACM Collective Intelligence Conference},
  pages={198--213},
  year={2025}
}

@article{Meincke2025,
  author    = {Lennart Meincke and Gideon Nave and Christian Terwiesch},
  title     = {ChatGPT decreases idea diversity in brainstorming},
  journal   = {Nature Human Behaviour},
  year      = {2025},
  volume    = {9},
  number    = {6},
  pages     = {1107--1109},
  month     = {June},
  doi       = {10.1038/s41562-025-02173-x},
  pmid      = {40369212},
  issn      = {2397-3374},
  language  = {English},
  note      = {Epub 2025 May 14},
  type      = {Letter},
  publisher = {Nature Publishing Group},
  address   = {England},
}

@article{lanchantin2025diverse,
  title={Diverse preference optimization},
  author={Lanchantin, Jack and Chen, Angelica and Dhuliawala, Shehzaad and Yu, Ping and Weston, Jason and Sukhbaatar, Sainbayar and Kulikov, Ilia},
  journal={arXiv preprint arXiv:2501.18101},
  year={2025}
}

@article{chung2025modifying,
  title={Modifying Large Language Model Post-Training for Diverse Creative Writing},
  author={Chung, John Joon Young and Padmakumar, Vishakh and Roemmele, Melissa and Sun, Yuqian and Kreminski, Max},
  journal={arXiv preprint arXiv:2503.17126},
  year={2025}
}

@article{vijayakumar2016diverse,
  title={Diverse beam search: Decoding diverse solutions from neural sequence models},
  author={Vijayakumar, Ashwin K and Cogswell, Michael and Selvaraju, Ramprasath R and Sun, Qing and Lee, Stefan and Crandall, David and Batra, Dhruv},
  journal={arXiv preprint arXiv:1610.02424},
  year={2016}
}

@article{meyerson2024language,
  title={Language model crossover: Variation through few-shot prompting},
  author={Meyerson, Elliot and Nelson, Mark J and Bradley, Herbie and Gaier, Adam and Moradi, Arash and Hoover, Amy K and Lehman, Joel},
  journal={ACM Transactions on Evolutionary Learning},
  volume={4},
  number={4},
  pages={1--40},
  year={2024},
  publisher={ACM New York, NY}
}

@article{wang2025multilingual,
  title={Multilingual Prompting for Improving LLM Generation Diversity},
  author={Wang, Qihan and Pan, Shidong and Linzen, Tal and Black, Emily},
  journal={arXiv preprint arXiv:2505.15229},
  year={2025}
}

@article{yang2025alignment,
  title={How Alignment Shrinks the Generative Horizon},
  author={Yang, Chenghao and Holtzman, Ari},
  journal={arXiv preprint arXiv:2506.17871},
  year={2025}
}

@misc{fei2025nudging,
      title={Nudging: Inference-time Alignment of LLMs via Guided Decoding}, 
      author={Yu Fei and Yasaman Razeghi and Sameer Singh},
      year={2025},
      eprint={2410.09300},
      archivePrefix={arXiv},
      primaryClass={cs.CL},
      url={https://arxiv.org/abs/2410.09300}, 
}

@inproceedings{
lu2025ai,
title={{AI} as Humanity{\textquoteright}s Salieri: Quantifying Linguistic Creativity of Language Models via Systematic Attribution of Machine Text against Web Text},
author={Ximing Lu and Melanie Sclar and Skyler Hallinan and Niloofar Mireshghallah and Jiacheng Liu and Seungju Han and Allyson Ettinger and Liwei Jiang and Khyathi Chandu and Nouha Dziri and Yejin Choi},
booktitle={The Thirteenth International Conference on Learning Representations},
year={2025},
url={https://openreview.net/forum?id=ilOEOIqolQ}
}

@inproceedings{chakrabarty2024art,
  title={Art or artifice? large language models and the false promise of creativity},
  author={Chakrabarty, Tuhin and Laban, Philippe and Agarwal, Divyansh and Muresan, Smaranda and Wu, Chien-Sheng},
  booktitle={Proceedings of the 2024 CHI Conference on Human Factors in Computing Systems},
  pages={1--34},
  year={2024}
}

@article{wang2025beyond,
  title={Beyond the 80/20 rule: High-entropy minority tokens drive effective reinforcement learning for llm reasoning},
  author={Wang, Shenzhi and Yu, Le and Gao, Chang and Zheng, Chujie and Liu, Shixuan and Lu, Rui and Dang, Kai and Chen, Xionghui and Yang, Jianxin and Zhang, Zhenru and others},
  journal={arXiv preprint arXiv:2506.01939},
  year={2025}
}

@article{zhao2024wildchat,
  title={Wildchat: 1m chatgpt interaction logs in the wild},
  author={Zhao, Wenting and Ren, Xiang and Hessel, Jack and Cardie, Claire and Choi, Yejin and Deng, Yuntian},
  journal={arXiv preprint arXiv:2405.01470},
  year={2024}
}

@article{tian2024large,
  title={Are large language models capable of generating human-level narratives?},
  author={Tian, Yufei and Huang, Tenghao and Liu, Miri and Jiang, Derek and Spangher, Alexander and Chen, Muhao and May, Jonathan and Peng, Nanyun},
  journal={arXiv preprint arXiv:2407.13248},
  year={2024}
}

@article{jiang2020can,
  title={How can we know what language models know?},
  author={Jiang, Zhengbao and Xu, Frank F and Araki, Jun and Neubig, Graham},
  journal={Transactions of the Association for Computational Linguistics},
  volume={8},
  pages={423--438},
  year={2020},
  publisher={MIT Press One Rogers Street, Cambridge, MA 02142-1209, USA journals-info~…}
}

@article{grattafiori2024llama,
  title={The llama 3 herd of models},
  author={Grattafiori, Aaron and Dubey, Abhimanyu and Jauhri, Abhinav and Pandey, Abhinav and Kadian, Abhishek and Al-Dahle, Ahmad and Letman, Aiesha and Mathur, Akhil and Schelten, Alan and Vaughan, Alex and others},
  journal={arXiv preprint arXiv:2407.21783},
  year={2024}
}

@article{olmo20242,
  title={2 OLMo 2 Furious},
  author={OLMo, Team and Walsh, Pete and Soldaini, Luca and Groeneveld, Dirk and Lo, Kyle and Arora, Shane and Bhagia, Akshita and Gu, Yuling and Huang, Shengyi and Jordan, Matt and others},
  journal={arXiv preprint arXiv:2501.00656},
  year={2024}
}

@inproceedings{liu-etal-2022-rethinking,
    title = "Rethinking and Refining the Distinct Metric",
    author = "Liu, Siyang  and
      Sabour, Sahand  and
      Zheng, Yinhe  and
      Ke, Pei  and
      Zhu, Xiaoyan  and
      Huang, Minlie",
    editor = "Muresan, Smaranda  and
      Nakov, Preslav  and
      Villavicencio, Aline",
    booktitle = "Proceedings of the 60th Annual Meeting of the Association for Computational Linguistics (Volume 2: Short Papers)",
    month = may,
    year = "2022",
    address = "Dublin, Ireland",
    publisher = "Association for Computational Linguistics",
    url = "https://aclanthology.org/2022.acl-short.86/",
    doi = "10.18653/v1/2022.acl-short.86",
    pages = "762--770"
}

@article{li2015diversity,
  title={A diversity-promoting objective function for neural conversation models},
  author={Li, Jiwei and Galley, Michel and Brockett, Chris and Gao, Jianfeng and Dolan, Bill},
  journal={arXiv preprint arXiv:1510.03055},
  year={2015}
}

@article{montahaei2019jointly,
  title={Jointly measuring diversity and quality in text generation models},
  author={Montahaei, Ehsan and Alihosseini, Danial and Baghshah, Mahdieh Soleymani},
  journal={arXiv preprint arXiv:1904.03971},
  year={2019}
}

@article{friedman2022vendi,
  title={The vendi score: A diversity evaluation metric for machine learning},
  author={Friedman, Dan and Dieng, Adji Bousso},
  journal={arXiv preprint arXiv:2210.02410},
  year={2022}
}

@article{gao2021simcse,
  title={Simcse: Simple contrastive learning of sentence embeddings},
  author={Gao, Tianyu and Yao, Xingcheng and Chen, Danqi},
  journal={arXiv preprint arXiv:2104.08821},
  year={2021}
}

@inproceedings{spangher2025newsinterview,
  title={NewsInterview: a dataset and a playground to evaluate LLMs’ grounding gap via informational interviews},
  author={Spangher, Alexander and Lu, Michael and Kalyan, Sriya and Cho, Hyundong Justin and Huang, Tenghao and Shi, Weiyan and May, Jonathan},
  booktitle={Proceedings of the 63rd Annual Meeting of the Association for Computational Linguistics (Volume 1: Long Papers)},
  pages={32895--32925},
  year={2025}
}

@inproceedings{spangher2025novel,
  title={A Novel Multi-Document Retrieval Benchmark: Journalist Source-Selection in Newswriting},
  author={Spangher, Alexander and Huang, Tenghao and Huang, Yiqin and Spangher, Lucas and Min, Sewon and Dredze, Mark},
  booktitle={Proceedings of the 4th International Workshop on Knowledge-Augmented Methods for Natural Language Processing},
  pages={180--204},
  year={2025}
}

@inproceedings{stasaski-hearst-2022-semantic,
    title = "Semantic Diversity in Dialogue with Natural Language Inference",
    author = "Stasaski, Katherine  and
      Hearst, Marti",
    editor = "Carpuat, Marine  and
      de Marneffe, Marie-Catherine  and
      Meza Ruiz, Ivan Vladimir",
    booktitle = "Proceedings of the 2022 Conference of the North American Chapter of the Association for Computational Linguistics: Human Language Technologies",
    month = jul,
    year = "2022",
    address = "Seattle, United States",
    publisher = "Association for Computational Linguistics",
    url = "https://aclanthology.org/2022.naacl-main.6/",
    doi = "10.18653/v1/2022.naacl-main.6",
    pages = "85--98"
}

@article{kuhn2023semantic,
  title={Semantic uncertainty: Linguistic invariances for uncertainty estimation in natural language generation},
  author={Kuhn, Lorenz and Gal, Yarin and Farquhar, Sebastian},
  journal={arXiv preprint arXiv:2302.09664},
  year={2023}
}

@article{liu2024skywork,
  title={Skywork-reward: Bag of tricks for reward modeling in llms},
  author={Liu, Chris Yuhao and Zeng, Liang and Liu, Jiacai and Yan, Rui and He, Jujie and Wang, Chaojie and Yan, Shuicheng and Liu, Yang and Zhou, Yahui},
  journal={arXiv preprint arXiv:2410.18451},
  year={2024}
}

@article{lambert2024rewardbench,
  title={Rewardbench: Evaluating reward models for language modeling},
  author={Lambert, Nathan and Pyatkin, Valentina and Morrison, Jacob and Miranda, LJ and Lin, Bill Yuchen and Chandu, Khyathi and Dziri, Nouha and Kumar, Sachin and Zick, Tom and Choi, Yejin and others},
  journal={arXiv preprint arXiv:2403.13787},
  year={2024}
}

@article{zhou2023lima,
  title={Lima: Less is more for alignment},
  author={Zhou, Chunting and Liu, Pengfei and Xu, Puxin and Iyer, Srinivasan and Sun, Jiao and Mao, Yuning and Ma, Xuezhe and Efrat, Avia and Yu, Ping and Yu, Lili and others},
  journal={Advances in Neural Information Processing Systems},
  volume={36},
  pages={55006--55021},
  year={2023}
}

@inproceedings{lin2023unlocking,
  title={The unlocking spell on base llms: Rethinking alignment via in-context learning},
  author={Lin, Bill Yuchen and Ravichander, Abhilasha and Lu, Ximing and Dziri, Nouha and Sclar, Melanie and Chandu, Khyathi and Bhagavatula, Chandra and Choi, Yejin},
  booktitle={The Twelfth International Conference on Learning Representations},
  year={2023}
}

@misc{ismayilzada2025creativepreferenceoptimization,
      title={Creative Preference Optimization}, 
      author={Mete Ismayilzada and Antonio Laverghetta Jr. and Simone A. Luchini and Reet Patel and Antoine Bosselut and Lonneke van der Plas and Roger Beaty},
      year={2025},
      eprint={2505.14442},
      archivePrefix={arXiv},
      primaryClass={cs.CL},
      url={https://arxiv.org/abs/2505.14442}, 
}

@inproceedings{wang2024beyond,
  title={Beyond Reverse KL: Generalizing Direct Preference Optimization with Diverse Divergence Constraints},
  author={Wang, Chaoqi and Jiang, Yibo and Yang, Chenghao and Liu, Han and Chen, Yuxin},
  year={2024},
  booktitle={The Twelfth International Conference on Learning Representations}
}

@article{kwon2024gdpo,
  title={GDPO: Learning to Directly Align Language Models with Diversity Using GFlowNets},
  author={Kwon, Oh Joon and Matsunaga, Daiki E and Kim, Kee-Eung},
  journal={arXiv preprint arXiv:2410.15096},
  year={2024}
}

@inproceedings{peeperkorn2024temperature,
  title={Is Temperature the Creativity Parameter of Large Language Models?},
  author={Peeperkorn, Max and Kouwenhoven, Tom and Brown, Dan and Jordanous, Anna},
  booktitle={ICCC},
  year={2024}
}

@article{wang2025large,
  title={Large language models that replace human participants can harmfully misportray and flatten identity groups},
  author={Wang, Angelina and Morgenstern, Jamie and Dickerson, John P},
  journal={Nature Machine Intelligence},
  pages={1--12},
  year={2025},
  publisher={Nature Publishing Group UK London}
}

@article{wong2024simplestrat,
  title={Simplestrat: Diversifying language model generation with stratification},
  author={Wong, Justin and Orlovskiy, Yury and Luo, Michael and Seshia, Sanjit A and Gonzalez, Joseph E},
  journal={arXiv preprint arXiv:2410.09038},
  year={2024}
}

@article{rafailov2023direct,
  title={Direct preference optimization: Your language model is secretly a reward model},
  author={Rafailov, Rafael and Sharma, Archit and Mitchell, Eric and Manning, Christopher D and Ermon, Stefano and Finn, Chelsea},
  journal={Advances in neural information processing systems},
  volume={36},
  pages={53728--53741},
  year={2023}
}

@article{rame2023rewarded,
  title={Rewarded soups: towards pareto-optimal alignment by interpolating weights fine-tuned on diverse rewards},
  author={Rame, Alexandre and Couairon, Guillaume and Dancette, Corentin and Gaya, Jean-Baptiste and Shukor, Mustafa and Soulier, Laure and Cord, Matthieu},
  journal={Advances in Neural Information Processing Systems},
  volume={36},
  pages={71095--71134},
  year={2023}
}

@inproceedings{zheng-etal-2025-model,
    title = "Model Extrapolation Expedites Alignment",
    author = "Zheng, Chujie  and
      Wang, Ziqi  and
      Ji, Heng  and
      Huang, Minlie  and
      Peng, Nanyun",
    editor = "Che, Wanxiang  and
      Nabende, Joyce  and
      Shutova, Ekaterina  and
      Pilehvar, Mohammad Taher",
    booktitle = "Proceedings of the 63rd Annual Meeting of the Association for Computational Linguistics (Volume 1: Long Papers)",
    month = jul,
    year = "2025",
    address = "Vienna, Austria",
    publisher = "Association for Computational Linguistics",
    url = "https://aclanthology.org/2025.acl-long.51/",
    doi = "10.18653/v1/2025.acl-long.51",
    pages = "1025--1041",
    ISBN = "979-8-89176-251-0",
}

@article{shi2025flexolmo,
  title={FlexOlmo: Open Language Models for Flexible Data Use},
  author={Shi, Weijia and Bhagia, Akshita and Farhat, Kevin and Muennighoff, Niklas and Walsh, Pete and Morrison, Jacob and Schwenk, Dustin and Longpre, Shayne and Poznanski, Jake and Ettinger, Allyson and others},
  journal={arXiv preprint arXiv:2507.07024},
  year={2025}
}

@inproceedings{shazeer2017outrageously,
  title={Outrageously Large Neural Networks: The Sparsely-Gated Mixture-of-Experts Layer},
  author={Shazeer, Noam and Mirhoseini, Azalia and Maziarz, Krzysztof and Davis, Andy and Le, Quoc and Hinton, Geoffrey and Dean, Jeff},
  booktitle={International Conference on Learning Representations},
  year={2017}
}

@inproceedings{li2023contrastive,
  title={Contrastive Decoding: Open-ended Text Generation as Optimization},
  author={Li, Xiang Lisa and Holtzman, Ari and Fried, Daniel and Liang, Percy and Eisner, Jason and Hashimoto, Tatsunori B and Zettlemoyer, Luke and Lewis, Mike},
  booktitle={Proceedings of the 61st Annual Meeting of the Association for Computational Linguistics (Volume 1: Long Papers)},
  pages={12286--12312},
  year={2023}
}

@inproceedings{
zheng2024citer,
title={{CITER}: Collaborative Inference for Efficient Large Language Model Decoding with Token-Level Routing},
author={Wenhao Zheng and Yixiao Chen and Weitong Zhang and Souvik Kundu and Yun Li and Zhengzhong Liu and Eric P. Xing and Hongyi Wang and Huaxiu Yao},
booktitle={Adaptive Foundation Models: Evolving AI for Personalized and Efficient Learning},
year={2024},
url={https://openreview.net/forum?id=0tSUpTUgfP}
}

@inproceedings{leviathan2023fast,
  title={Fast inference from transformers via speculative decoding},
  author={Leviathan, Yaniv and Kalman, Matan and Matias, Yossi},
  booktitle={International Conference on Machine Learning},
  pages={19274--19286},
  year={2023},
  organization={PMLR}
}

@inproceedings{lu2024llm,
  title={LLM Discussion: Enhancing the Creativity of Large Language Models via Discussion Framework and Role-Play},
  author={Lu, Li-Chun and Chen, Shou-Jen and Pai, Tsung-Min and Yu, Chan-Hung and Lee, Hung-yi and Sun, Shao-Hua},
  year={2024},
  booktitle={First Conference on Language Modeling}
}

@inproceedings{venkatraman2025collabstory,
  title={CollabStory: Multi-LLM Collaborative Story Generation and Authorship Analysis},
  author={Venkatraman, Saranya and Tripto, Nafis Irtiza and Lee, Dongwon},
  booktitle={Findings of the Association for Computational Linguistics: NAACL 2025},
  pages={3665--3679},
  year={2025}
}

@inproceedings{
huot2025agents,
title={Agents' Room:  Narrative Generation through Multi-step Collaboration},
author={Fantine Huot and Reinald Kim Amplayo and Jennimaria Palomaki and Alice Shoshana Jakobovits and Elizabeth Clark and Mirella Lapata},
booktitle={The Thirteenth International Conference on Learning Representations},
year={2025},
url={https://openreview.net/forum?id=HfWcFs7XLR}
}

@article{dubey2024llama,
  title={The llama 3 herd of models},
  author={Dubey, Abhimanyu and Jauhri, Abhinav and Pandey, Abhinav and Kadian, Abhishek and Al-Dahle, Ahmad and Letman, Aiesha and Mathur, Akhil and Schelten, Alan and Yang, Amy and Fan, Angela and others},
  journal={arXiv preprint arXiv:2407.21783},
  year={2024}
}

@article{bai2022training,
  title={Training a helpful and harmless assistant with reinforcement learning from human feedback},
  author={Bai, Yuntao and Jones, Andy and Ndousse, Kamal and Askell, Amanda and Chen, Anna and DasSarma, Nova and Drain, Dawn and Fort, Stanislav and Ganguli, Deep and Henighan, Tom and others},
  journal={arXiv preprint arXiv:2204.05862},
  year={2022}
}

@article{10.1007/s00158-016-1469-3,
author = {Zheng, Kai and Yang, Ren-Jye and Xu, Hongyi and Hu, Jie},
title = {A new distribution metric for comparing Pareto optimal solutions},
year = {2017},
issue_date = {January   2017},
publisher = {Springer-Verlag},
address = {Berlin, Heidelberg},
volume = {55},
number = {1},
issn = {1615-147X},
url = {https://doi.org/10.1007/s00158-016-1469-3},
doi = {10.1007/s00158-016-1469-3},
journal = {Struct. Multidiscip. Optim.},
month = jan,
pages = {53–62},
numpages = {10}
}

@book{zitzler1999evolutionary,
  title={Evolutionary algorithms for multiobjective optimization: Methods and applications},
  author={Zitzler, Eckart},
  volume={63},
  year={1999},
  publisher={Shaker Ithaca}
}

@misc{li2025jointlyreinforcingdiversityquality,
      title={Jointly Reinforcing Diversity and Quality in Language Model Generations}, 
      author={Tianjian Li and Yiming Zhang and Ping Yu and Swarnadeep Saha and Daniel Khashabi and Jason Weston and Jack Lanchantin and Tianlu Wang},
      year={2025},
      eprint={2509.02534},
      archivePrefix={arXiv},
      primaryClass={cs.CL},
      url={https://arxiv.org/abs/2509.02534}, 
}

@article{DeepSeekAI2025DeepSeekR1IR,
  title={DeepSeek-R1: Incentivizing Reasoning Capability in LLMs via Reinforcement Learning},
  author={DeepSeek-AI and Daya Guo and Dejian Yang and Haowei Zhang and Jun-Mei Song and Ruoyu Zhang and Runxin Xu and Qihao Zhu and Shirong Ma and Peiyi Wang and Xiaoling Bi and Xiaokang Zhang and Xingkai Yu and Yu Wu and Z. F. Wu and Zhibin Gou and Zhihong Shao and Zhuoshu Li and Ziyi Gao and Aixin Liu and Bing Xue and Bing-Li Wang and Bochao Wu and Bei Feng and Chengda Lu and Chenggang Zhao and Chengqi Deng and Chenyu Zhang and Chong Ruan and Damai Dai and Deli Chen and Dong-Li Ji and Erhang Li and Fangyun Lin and Fucong Dai and Fuli Luo and Guangbo Hao and Guanting Chen and Guowei Li and H. Zhang and Han Bao and Hanwei Xu and Haocheng Wang and Honghui Ding and Huajian Xin and Huazuo Gao and Hui Qu and Hui Li and Jianzhong Guo and Jiashi Li and Jiawei Wang and Jingchang Chen and Jingyang Yuan and Junjie Qiu and Junlong Li and Jiong Cai and Jiaqi Ni and Jian Liang and Jin Chen and Kai Dong and Kai Hu and Kaige Gao and Kang Guan and Kexin Huang and Kuai Yu and Lean Wang and Lecong Zhang and Liang Zhao and Litong Wang and Liyue Zhang and Lei Xu and Leyi Xia and Mingchuan Zhang and Minghua Zhang and M. Tang and Meng Li and Miaojun Wang and Mingming Li and Ning Tian and Panpan Huang and Peng Zhang and Qiancheng Wang and Qinyu Chen and Qiushi Du and Ruiqi Ge and Ruisong Zhang and Ruizhe Pan and Runji Wang and R. J. Chen and Ruiqi Jin and Ruyi Chen and Shanghao Lu and Shangyan Zhou and Shanhuang Chen and Shengfeng Ye and Shiyu Wang and Shuiping Yu and Shunfeng Zhou and Shuting Pan and S. S. Li and Shuang Zhou and Shao-Kang Wu and Tao Yun and Tian Pei and Tianyu Sun and T. Wang and Wangding Zeng and Wanjia Zhao and Wen Liu and Wenfeng Liang and Wenjun Gao and Wen-Xia Yu and Wentao Zhang and Wangding Xiao and Wei An and Xiaodong Liu and Xiaohan Wang and Xi-aokang Chen and Xiaotao Nie and Xin Cheng and Xin Liu and Xin Xie and Xingchao Liu and Xinyu Yang and Xinyuan Li and Xuecheng Su and Xuheng Lin and X. Q. Li and Xiangyu Jin and Xi-Cheng Shen and Xiaosha Chen and Xiaowen Sun and Xiaoxiang Wang and Xinnan Song and Xinyi Zhou and Xianzu Wang and Xinxia Shan and Y. K. Li and Y. Q. Wang and Y. X. Wei and Yang Zhang and Yanhong Xu and Yao Li and Yao Zhao and Yaofeng Sun and Yaohui Wang and Yi Yu and Yichao Zhang and Yifan Shi and Yi Xiong and Ying He and Yishi Piao and Yisong Wang and Yixuan Tan and Yiyang Ma and Yiyuan Liu and Yongqiang Guo and Yuan Ou and Yuduan Wang and Yue Gong and Yu-Jing Zou and Yujia He and Yunfan Xiong and Yu-Wei Luo and Yu-mei You and Yuxuan Liu and Yuyang Zhou and Y. X. Zhu and Yanping Huang and Yao Li and Yi Zheng and Yuchen Zhu and Yunxiang Ma and Ying Tang and Yukun Zha and Yuting Yan and Zehui Ren and Zehui Ren and Zhangli Sha and Zhe Fu and Zhean Xu and Zhenda Xie and Zhen-guo Zhang and Zhewen Hao and Zhicheng Ma and Zhigang Yan and Zhiyu Wu and Zihui Gu and Zijia Zhu and Zijun Liu and Zi-An Li and Ziwei Xie and Ziyang Song and Zizheng Pan and Zhen Huang and Zhipeng Xu and Zhongyu Zhang and Zhen Zhang},
  journal={ArXiv},
  year={2025},
  volume={abs/2501.12948},
  url={https://api.semanticscholar.org/CorpusID:275789950}
}

@article{huang2025teaching,
  title={Teaching Language Models To Gather Information Proactively},
  author={Huang, Tenghao and Chen, Sihao and Chen, Muhao and May, Jonathan and Yang, Longqi and Wan, Mengting and Zhou, Pei},
  journal={Findings of the Association for Computational Linguistics: EMNLP 2025},
  pages={15588--15599},
  year={2025}
}

@inproceedings{spangher-etal-2025-creative,
    title = "Creative Planning with Language Models: Practice, Evaluation and Applications",
    author = "Spangher, Alexander  and
      Huang, Tenghao  and
      Laban, Philippe  and
      Peng, Nanyun",
    editor = "Lomeli, Maria  and
      Swayamdipta, Swabha  and
      Zhang, Rui",
    booktitle = "Proceedings of the 2025 Annual Conference of the Nations of the Americas Chapter of the Association for Computational Linguistics: Human Language Technologies (Volume 5: Tutorial Abstracts)",
    month = may,
    year = "2025",
    address = "Albuquerque, New Mexico",
    publisher = "Association for Computational Linguistics",
    url = "https://aclanthology.org/2025.naacl-tutorial.1/",
    doi = "10.18653/v1/2025.naacl-tutorial.1",
    pages = "1--9",
    ISBN = "979-8-89176-193-3"
}

@inproceedings{yao2019plan,
  title={Plan-and-write: Towards better automatic storytelling},
  author={Yao, Lili and Peng, Nanyun and Weischedel, Ralph and Knight, Kevin and Zhao, Dongyan and Yan, Rui},
  booktitle={Proceedings of the AAAI Conference on Artificial Intelligence},
  volume={33},
  number={01},
  pages={7378--7385},
  year={2019}
}

@inproceedings{sims2019literary,
  title={Literary event detection},
  author={Sims, Matthew and Park, Jong Ho and Bamman, David},
  booktitle={Proceedings of the 57th annual meeting of the association for computational linguistics},
  pages={3623--3634},
  year={2019}
}

@misc{chiang2024chatbotarenaopenplatform,
      title={Chatbot Arena: An Open Platform for Evaluating LLMs by Human Preference}, 
      author={Wei-Lin Chiang and Lianmin Zheng and Ying Sheng and Anastasios Nikolas Angelopoulos and Tianle Li and Dacheng Li and Hao Zhang and Banghua Zhu and Michael Jordan and Joseph E. Gonzalez and Ion Stoica},
      year={2024},
      eprint={2403.04132},
      archivePrefix={arXiv},
      primaryClass={cs.AI},
      url={https://arxiv.org/abs/2403.04132}, 
}

@misc{OpenAI2025IntroducingDeepResearch,
  author       = {OpenAI},
  title        = {Introducing deep research},
  howpublished = {\url{https://openai.com/index/introducing-deep-research/}},
  year         = {2025},
  month        = feb,
  day          = {2},
  note         = {Accessed: \today}
}

@article{piao2023human,
  title={Human--AI adaptive dynamics drives the emergence of information cocoons},
  author={Piao, Jinghua and Liu, Jiazhen and Zhang, Fang and Su, Jun and Li, Yong},
  journal={Nature Machine Intelligence},
  volume={5},
  number={11},
  pages={1214--1224},
  year={2023},
  publisher={Nature Publishing Group UK London}
}

@inproceedings{siddiqui2025script,
  title={Script\&Shift: A layered interface paradigm for integrating content development and rhetorical strategy with llm writing assistants},
  author={Siddiqui, Momin N and Pea, Roy D and Subramonyam, Hari},
  booktitle={Proceedings of the 2025 CHI Conference on Human Factors in Computing Systems},
  pages={1--19},
  year={2025}
}

@ARTICLE{2025arXiv250614758C,
       author = {{Cheng}, Daixuan and {Huang}, Shaohan and {Zhu}, Xuekai and {Dai}, Bo and {Zhao}, Wayne Xin and {Zhang}, Zhenliang and {Wei}, Furu},
        title = "{Reasoning with Exploration: An Entropy Perspective on Reinforcement Learning for LLMs}",
      journal = {arXiv e-prints},
         year = 2025,
        month = jun,
          eid = {arXiv:2506.14758},
        pages = {arXiv:2506.14758},
          doi = {10.48550/arXiv.2506.14758},
archivePrefix = {arXiv},
       eprint = {2506.14758},
 primaryClass = {cs.CL},
       adsurl = {https://ui.adsabs.harvard.edu/abs/2025arXiv250614758C}
}

@misc{song2025outcomebasedexplorationllmreasoning,
      title={Outcome-based Exploration for LLM Reasoning}, 
      author={Yuda Song and Julia Kempe and Remi Munos},
      year={2025},
      eprint={2509.06941},
      archivePrefix={arXiv},
      primaryClass={cs.LG},
      url={https://arxiv.org/abs/2509.06941}, 
}

@misc{leviathan2023fastinferencetransformersspeculative,
      title={Fast Inference from Transformers via Speculative Decoding}, 
      author={Yaniv Leviathan and Matan Kalman and Yossi Matias},
      year={2023},
      eprint={2211.17192},
      archivePrefix={arXiv},
      primaryClass={cs.LG},
      url={https://arxiv.org/abs/2211.17192}, 
}

@misc{wu2024mixtureloraexperts,
      title={Mixture of LoRA Experts}, 
      author={Xun Wu and Shaohan Huang and Furu Wei},
      year={2024},
      eprint={2404.13628},
      archivePrefix={arXiv},
      primaryClass={cs.CL},
      url={https://arxiv.org/abs/2404.13628}, 
}

@article{im2024understanding,
  title={Understanding the learning dynamics of alignment with human feedback},
  author={Im, Shawn and Li, Yixuan},
  journal={arXiv preprint arXiv:2403.18742},
  year={2024}
}

@misc{ren2025learningdynamicsllmfinetuning,
      title={Learning Dynamics of LLM Finetuning}, 
      author={Yi Ren and Danica J. Sutherland},
      year={2025},
      eprint={2407.10490},
      archivePrefix={arXiv},
      primaryClass={cs.LG},
      url={https://arxiv.org/abs/2407.10490}, 
}

@misc{ding2025dynamicparalleltreesearch,
      title={Dynamic Parallel Tree Search for Efficient LLM Reasoning}, 
      author={Yifu Ding and Wentao Jiang and Shunyu Liu and Yongcheng Jing and Jinyang Guo and Yingjie Wang and Jing Zhang and Zengmao Wang and Ziwei Liu and Bo Du and Xianglong Liu and Dacheng Tao},
      year={2025},
      eprint={2502.16235},
      archivePrefix={arXiv},
      primaryClass={cs.AI},
      url={https://arxiv.org/abs/2502.16235}, 
}

@inproceedings{papineni2002bleu,
  title={Bleu: a method for automatic evaluation of machine translation},
  author={Papineni, Kishore and Roukos, Salim and Ward, Todd and Zhu, Wei-Jing},
  booktitle={Proceedings of the 40th annual meeting of the Association for Computational Linguistics},
  pages={311--318},
  year={2002}
}

@inproceedings{lin2004rouge,
  title={Rouge: A package for automatic evaluation of summaries},
  author={Lin, Chin-Yew},
  booktitle={Text summarization branches out},
  pages={74--81},
  year={2004}
}

@article{talebirad2023multi,
  title={Multi-agent collaboration: Harnessing the power of intelligent llm agents},
  author={Talebirad, Yashar and Nadiri, Amirhossein},
  journal={arXiv preprint arXiv:2306.03314},
  year={2023}
}

@article{jaarsveld2012creative,
  title={Creative reasoning across developmental levels: Convergence and divergence in problem creation},
  author={Jaarsveld, Saskia and Lachmann, Thomas and Van Leeuwen, Cees},
  journal={Intelligence},
  volume={40},
  number={2},
  pages={172--188},
  year={2012},
  publisher={Elsevier}
}

@article{qi2023fine,
  title={Fine-tuning aligned language models compromises safety, even when users do not intend to!},
  author={Qi, Xiangyu and Zeng, Yi and Xie, Tinghao and Chen, Pin-Yu and Jia, Ruoxi and Mittal, Prateek and Henderson, Peter},
  journal={arXiv preprint arXiv:2310.03693},
  year={2023}
}

@article{cobbe2021training,
  title={Training verifiers to solve math word problems},
  author={Cobbe, Karl and Kosaraju, Vineet and Bavarian, Mohammad and Chen, Mark and Jun, Heewoo and Kaiser, Lukasz and Plappert, Matthias and Tworek, Jerry and Hilton, Jacob and Nakano, Reiichiro and others},
  journal={arXiv preprint arXiv:2110.14168},
  year={2021}
}

@misc{zhou2023instructionfollowingevaluationlargelanguage,
      title={Instruction-Following Evaluation for Large Language Models}, 
      author={Jeffrey Zhou and Tianjian Lu and Swaroop Mishra and Siddhartha Brahma and Sujoy Basu and Yi Luan and Denny Zhou and Le Hou},
      year={2023},
      eprint={2311.07911},
      archivePrefix={arXiv},
      primaryClass={cs.CL},
      url={https://arxiv.org/abs/2311.07911}, 
}

@misc{zheng2023judgingllmasajudgemtbenchchatbot,
      title={Judging LLM-as-a-Judge with MT-Bench and Chatbot Arena}, 
      author={Lianmin Zheng and Wei-Lin Chiang and Ying Sheng and Siyuan Zhuang and Zhanghao Wu and Yonghao Zhuang and Zi Lin and Zhuohan Li and Dacheng Li and Eric P. Xing and Hao Zhang and Joseph E. Gonzalez and Ion Stoica},
      year={2023},
      eprint={2306.05685},
      archivePrefix={arXiv},
      primaryClass={cs.CL},
      url={https://arxiv.org/abs/2306.05685}, 
}

@misc{hu2021loralowrankadaptationlarge,
      title={LoRA: Low-Rank Adaptation of Large Language Models}, 
      author={Edward J. Hu and Yelong Shen and Phillip Wallis and Zeyuan Allen-Zhu and Yuanzhi Li and Shean Wang and Lu Wang and Weizhu Chen},
      year={2021},
      eprint={2106.09685},
      archivePrefix={arXiv},
      primaryClass={cs.CL},
      url={https://arxiv.org/abs/2106.09685}, 
}

@misc{liu2025droidspeakkvcachesharing,
      title={DroidSpeak: KV Cache Sharing for Cross-LLM Communication and Multi-LLM Serving}, 
      author={Yuhan Liu and Yuyang Huang and Jiayi Yao and Shaoting Feng and Zhuohan Gu and Kuntai Du and Hanchen Li and Yihua Cheng and Junchen Jiang and Shan Lu and Madan Musuvathi and Esha Choukse},
      year={2025},
      eprint={2411.02820},
      archivePrefix={arXiv},
      primaryClass={cs.MA},
      url={https://arxiv.org/abs/2411.02820}, 
}

@misc{jiang2025artificialhivemindopenendedhomogeneity,
      title={Artificial Hivemind: The Open-Ended Homogeneity of Language Models (and Beyond)}, 
      author={Liwei Jiang and Yuanjun Chai and Margaret Li and Mickel Liu and Raymond Fok and Nouha Dziri and Yulia Tsvetkov and Maarten Sap and Alon Albalak and Yejin Choi},
      year={2025},
      eprint={2510.22954},
      archivePrefix={arXiv},
      primaryClass={cs.CL},
      url={https://arxiv.org/abs/2510.22954}, 
}

@misc{nguyen2025turningheatminpsampling,
      title={Turning Up the Heat: Min-p Sampling for Creative and Coherent LLM Outputs}, 
      author={Minh Nhat Nguyen and Andrew Baker and Clement Neo and Allen Roush and Andreas Kirsch and Ravid Shwartz-Ziv},
      year={2025},
      eprint={2407.01082},
      archivePrefix={arXiv},
      primaryClass={cs.CL},
      url={https://arxiv.org/abs/2407.01082}, 
}
\bibliographystyle{icml2025}

%%%%%%%%%%%%%%%%%%%%%%%%%%%%%%%%%%%%%%%%%%%%%%%%%%%%%%%%%%%%%%%%%%%%%%%%%%%%%%%
%%%%%%%%%%%%%%%%%%%%%%%%%%%%%%%%%%%%%%%%%%%%%%%%%%%%%%%%%%%%%%%%%%%%%%%%%%%%%%%
% APPENDIX
%%%%%%%%%%%%%%%%%%%%%%%%%%%%%%%%%%%%%%%%%%%%%%%%%%%%%%%%%%%%%%%%%%%%%%%%%%%%%%%
%%%%%%%%%%%%%%%%%%%%%%%%%%%%%%%%%%%%%%%%%%%%%%%%%%%%%%%%%%%%%%%%%%%%%%%%%%%%%%%
\newpage
\appendix
\onecolumn

\renewcommand{\thefigure}{A\arabic{figure}}
\renewcommand{\thetable}{A\arabic{table}}
\appendix

\section*{Limitations}
We see several limitations in our work.

(\textit{i}) Collaboration between two models, by nature, incurs additional FLOPs overhead per token relative to single-model inference; the prototype is not yet perfectly engineered for speed, but we see speculative decoding, LoRA-based aligned models, and KV-cache sharing are mitigations (\appref{app:analysis}).

(\textit{ii}) Beyond the open-ended generation tasks studied here, other tasks such as agent and code generation can also benefit from diversity, and can extend to much longer horizons; we leave these extensions for future work.

(\textit{iii}) Our focus is inference-time improvement, so directions such as learned routers and leveraging diversity within RL rollouts, while relevant, are out of scope for this paper.

(\textit{iv}) We mainly study the canonical base and aligned checkpoints; intermediate alignment checkpoints, where available, likely offer additional Pareto gains and are a promising direction (\appref{app:analysis}).

\section{Related Work}
\label{sec:related_work}
% \yc{@chenghao, planned to only keep at most half-page, and the full version go the appendix.}
\paragraph{Diversity Degradation in Alignment.}
% \subsection{The Alignment-Diversity Trade-off}
While alignment techniques like RLHF enhance model performance in instruction following and reasoning, they systematically curtail output diversity. This trade-off is widely recognized, with a growing body of evidence demonstrating that aligned models are less diverse than their base counterparts. For example, studies have questioned their artistic authenticity \citep{chakrabarty2024art}, and benchmarks like NoveltyBench reveal their diminished capacity for humanlike diversity and creativity \citep{zhang2025noveltybench, tian2024large, lu2025ai, west2025base}. The underlying mechanism for this degradation is ``probability concentration," as the alignment process ``sharpens'' the model's output distribution, thereby steering it into low-entropy generation paths \citep{yang2025alignment}.
The diversity degradation impacts the downstream applications. 
It manifests as a loss of linguistic idiosyncrasies \citep{chakrabarty2025can}, increased format homogeneity \citep{zhang2024lists}, and diminished creativity \citep{west2025base} in generated text.
Beyond linguistic characteristics, alignment-induced constraints lead models to converge on a restricted repertoire of strategies, thereby diminishing diversity in reasoning \citep{chen2024not, ding2025dynamic}, data synthesis \citep{kim2024evaluating, yang2025measuring}, deep research \citep{Xiao2025SubmodularOptimization}, social simulation \citep{wang2025large}, and gaming \citep{west2025base}. 
More critically, the lack of output diversity has been shown to further reduce outcome diversity \citep{padmakumar2023does} and creativity \citep{Meincke2025, ashkinaze2025ai} in human interactions with these models. 
% They demonstrate that diminished variation in responses diminished response variation adversely affects how humans ideate, create, and engage \mina{ungrammatical}.
These studies demonstrate that diminished diversity in model outputs adversely affects how humans ideate, create, and engage.
% The evidences underscore the importance of developing diversity-promoting methodology for aligned LLMs.
Our work addresses this challenge directly, proposing an inference-time collaborative decoding framework that optimizes the \problem{} by combining the strengths of both base and aligned models.

\paragraph{Diversity-Promoting Methods.}
Approaches to enhance the diversity of aligned LLMs fall into two main categories: training-time and inference-time methods.
\textbf{Training-time methods} typically modify the learning objective to encourage varied outputs. A prominent line of work adapts Direct Preference Optimization (DPO; \citet{rafailov2023direct}) by incorporating diversity-aware mechanisms, such as f-divergence penalties \citep{wang2024beyond}, set-level diversity rewards \citep{lanchantin2025diverse}, or re-weighted loss objectives \citep{chung2025modifying, ismayilzada2025creativepreferenceoptimization}. Other approaches leverage different architectures, like generative flow networks, to the same end \citep{kwon2024gdpo}. While these methods can instill diversity directly into the model, they require substantial computational resources for retraining and offer little flexibility for user-specific diversity needs at inference.
\textbf{Inference-time methods} offer a more lightweight and adaptable alternative. These include modifications to decoding algorithms like diverse beam search \citep{vijayakumar2016diverse} and various prompt engineering strategies, such as paraphrasing \citep{meyerson2024language, wang2025multilingual, zhang2025noveltybench, wang2025large, wong2024simplestrat}. Existing inference-time methods for improving diversity typically incur high computational costs via multiple decoding passes or long-horizon planning. Or they significantly degrade generation quality \citep{peeperkorn2024temperature}. Achieving a stable diversity-quality trade-off with these techniques remains a challenge.
Our proposed method \name, is an inference-time framework designed to offer more explicit and reliable control over this trade-off. We therefore focus our comparison on baselines from this category.

\paragraph{Multi-Model Collaborative Generation.}
Prior work has explored collaborative frameworks where multiple language models work in concert to improve generation quality and efficiency (\eg{} computation cost and latency). These approaches can be grouped by their collaboration mechanism.
One line of work focuses on \textbf{weight-level collaobration}. This includes merging reward or policy models, or using Mixture-of-Experts (MoE)~\citep{shazeer2017outrageously} architectures to create a single, more capable system better aligned with diverse human preferences \citep{rame2023rewarded, zheng-etal-2025-model, shi2025flexolmo}. Another mechanism is \textbf{token-level collaboration}, where multiple models collaborate during decoding by exchanging next-token probability distributions or candidate token choices at each step, to improve attributes like coherence and factuality, or reduce latency \citep{leviathan2023fast, li2023contrastive, zheng2024citer, fei2025nudging}. More recently, \textbf{multi-agent systems} have emerged, in which models debate or discuss to leverage their complementary strengths for complex, creative tasks \citep{lu2024llm, venkatraman2025collabstory, huot2025agents}.
Our \name, advances token-level collaboration. While existing methods in this area primarily target quality or efficiency, we focus on navigating the diversity-quality trade-off. 

\section{\name{} Framework Implementation and Router Details}
\label{app: implementation_details}

\paragraph{Models.}

We apply each model's default chat template during collaboration (for base models, we apply a plain shifting template), served on vLLM local host.
We disable every model's tool call and thinking for fair comparison.

\paragraph{Tokenization Alignment.}
Tokenization sometimes mismatches between the two models (even base and aligned pairs), particularly around punctuation, special tokens, or rare words, can lead to incoherent sub-word boundaries. To address this, we enforce that tokens representing a single semantic unit (e.g., a word or format element) must all be produced by the same model. This avoids artifacts such as broken punctuation or malformed words.

\paragraph{Algorithm Overview.}

Algorithm~\ref{alg:baco} formalizes the full \name{} decoding procedure, including the additional rules described above (aligned first token, word-boundary switching, aligned-controlled EOS) and reconciles Eq.~\eqref{eq:moe_main} with the actual sampling step. We express the router as an ordered tuple of routing strategies $\mathcal{R} = (\mathcal{R}_1, \dots, \mathcal{R}_K)$, where each $\mathcal{R}_k$ may fire in either direction (\textsf{base} or \textsf{aligned}) on its trigger condition, or abstain otherwise (move to next strategy). 
% For example, \textsc{-Punc} fires \textsf{aligned} on punctuation or formatting tokens and abstains on the rest; \textsc{-FC} fires \textsf{base} when both models predict a content word and abstains otherwise; \textsc{-P} always fires, returning \textsf{base} when $\max_{y_t} P_{\text{base}}(y_t \mid \cdot) < \gamma$ and \textsf{aligned} otherwise. 
At a routing step, the strategies are evaluated in order and the first $\mathcal{R}_k$ that fires determines $m$. By construction, the last strategy in the tuple always fires, so the cascade always terminates with a definite decision. Single-strategy routers correspond to $K = 1$; combination routers correspond to $K \geq 2$ with complementary signals in practice.

\begin{algorithm}[h]
\caption{\name{} Decoding}
\label{alg:baco}
\begin{algorithmic}[1]
\REQUIRE prompt $\inputval$; base model $P_{\text{base}}$;
         aligned model $P_{\text{aligned}}$;
         router $\mathcal{R} = (\mathcal{R}_1, \dots, \mathcal{R}_K)$
         with threshold $\gamma$; max length $T$
\STATE $\outputval_1 \sim P_{\text{aligned}}(\cdot \mid \inputval)$
       \hfill $\triangleright$ sample first token from aligned
\STATE $m \gets \text{aligned}$
\FOR{$t = 2, \dots, T$}
  \STATE $c_t \gets [\inputval, \outputval_{<t}]$
  \IF{$\outputval_{t-1}$ ends a word}
     \STATE $m \gets \mathcal{R}(c_t, P_{\text{base}}, P_{\text{aligned}}, \gamma)$
            \hfill $\triangleright$ go through each strategies within router $\mathcal{R}$ and returns \textsf{base} or \textsf{aligned}
  \ENDIF
  \STATE $\outputval_t \sim P_m(\cdot \mid c_t)$
  \IF{$\outputval_t = \langle \text{eos} \rangle$ \AND $\arg\max P_{\text{aligned}}(\cdot \mid c_t) \neq \langle \text{eos} \rangle$}
     \STATE resample $\outputval_t \sim P_m(\cdot \mid c_t)$ excluding $\langle \text{eos} \rangle$
            \hfill $\triangleright$ aligned-controlled termination
  \ENDIF
  \IF{$\outputval_t = \langle \text{eos} \rangle$} \STATE \textbf{break} \ENDIF
\ENDFOR
\STATE \textbf{return} $\outputval_1, \dots, \outputval_t$
\end{algorithmic}
\end{algorithm}

\paragraph{Framework Cost.}
\name{} requires two forward passes per decoding step at the worst case, incurring $\sim$2$\times$ FLOPs overhead per token regardless of sample count. Crucially, this overhead is independent of the desired group size $n$, in contrast to baselines such as in-context resampling and paraphrase prompting (which require $n$ sequential passes) or diverse beam search (which expands $\geq n$ beams).
Moreover, stems from \textit{superficial alignment}~\citep{zhou2023lima, lin2023unlocking}, interventions between the largely-agreeing base and aligned models can be sparse~\citep{fei2025nudging}. This sparsity enables practical optimizations such as caching multi-token chunks from one model to minimize switching costs and overlapping communication with computation to hide context-switching latency. Further engineering-level optimizations are complementary and applicable when deploying at scale, including speculative decoding~\citep{leviathan2023fast}, LoRA-based aligned models \citep{hu2021loralowrankadaptationlarge}, and KV cache sharing \citep{liu2025droidspeakkvcachesharing}.

\paragraph{Wall-Clock Runtime Comparison.}
We empirically measure runtime under identical hardware (single A100 80\,GB) for inference-time methods, using the HuggingFace library across the board for fair comparison. Results are in Table~\ref{tab:wallclock}. \name{}'s current prototype is slower than single-model baselines but is comparable to or faster than several diversity-promoting baselines (\eg{} Diverse Beam Search, Logits Ensemble) while substantially outperforming them on the diversity-quality trade-off. The implementation is not yet engineered for speed; the optimizations listed above can further reduce wall-clock cost.

\begin{table}[h]
\centering
\small
\renewcommand{\arraystretch}{1.2}
\begin{tabular}{lccc}
\toprule
\textbf{Method} & \textbf{Time/Sample (s)} & \textbf{Tok/s} & \textbf{Peak Memory} \\
\midrule
Aligned                & $9.8 \pm 3.9$    & 40.0 & 15.1 GiB \\
Response Ensemble      & $10.9 \pm 4.0$   & 34.8 & 15.4 GiB \\
Diverse Beam Search    & $17.0 \pm 5.9$   & 17.0 & 15.8 GiB \\
Logits Ensemble        & $25.5 \pm 0.5$   & 20.1 & 30.1 GiB \\
In-Context Resampling  & $90.5 \pm 35.3$  & 4.1  & 15.5 GiB \\
Paraphrase Prompting   & $117.9 \pm 41.1$ & 3.3  & 15.1 GiB \\
Back Translation       & $120.5 \pm 65.6$ & 3.8  & 15.1 GiB \\
Mix-of-Agents          & $224.9 \pm 123.2$& 3.8  & 15.6 GiB \\
Multi-Agent Debate     & $239.0 \pm 81.5$ & 3.5  & 15.6 GiB \\
\midrule
\name{}                & $22.9 \pm 4.5$   & 19.2 & 29.2 GiB \\
\bottomrule
\end{tabular}
\caption{Wall-clock runtime comparison on a single A100 80\,GB. Tok/s is computed over the final output. \name{} is comparable to or faster than several inference-time diversity baselines while substantially outperforming them on diversity--quality (\Cref{tab:avg_results}).}
\label{tab:wallclock}
\end{table}

% \section{Router Designs}
\label{app:router}
\subsection{Additional Rules in Router\label{app:router:addrule}}  
We follow \cite{fei2025nudging} in always using the aligned model to generate the first token. Early decoding steps have an outsized influence on generation and typically show greater disagreement between models. Starting from the aligned model improves trajectory quality and reduces the chance of degenerate completions.

Incorporating low-probability tokens from the base model introduces new challenges in sequential generation. 
When switch between models, the receiving model may struggle to continue from an unfamiliar context. 
In particular, the aligned model may terminate the output prematurely, while the base model may fall into degenerate behaviors such as repetition or verbose listing. 
To mitigate this, we constrain output termination by only accepting the end-of-sentence token when it is the top-1 prediction of the aligned model. 
% This heuristic ensures more reliable stopping behavior and avoids early exits triggered by uncertain base model completions. 
% \yc{not sure if we need this here} We view that some premature terminations can be beneficial in best-of-$n$ generation settings, as they may yield shorter low-quality candidates, which save cost and are easy to filter. Further analysis in \secref{}. \thh{dont forget to fix}

\subsection{All Routing Strategies and Notations}
\label{app:router}
\label{app:router:notation}
The following are all strategies and their corresponding notations that are mentioned in this paper:
\begin{itemize}
    \item -\textsc{Rand}: Route to the base model by random chance $\gamma$.
    \item -\textsc{P}: Route to the base model when base model's top-1 token probability $\max_{y_t} P_{\text{base}}(y_t \mid x, y_{<t}) < \gamma ~_{\gamma \in [0, 1]}$, otherwise to the aligned model.
    \item -\textsc{P-a}: Route to the base model when aligned model's top-1 token probability $\max_{y_t} P_{\text{aligned}}(y_t \mid x, y_{<t}) < \gamma ~_{\gamma \in [0, 1]}$, otherwise to the aligned model.
    % {Comparatively, we also have the \name{}-\textsc{P-Aligned} variant when use the aligned model's top 1 token probability. The results are at \appref{}. As the mode collapses, the aligned model's top 1 token probabilities are always huge and lack discriminativeness. It also frequently makes highly confident predictions at token positions that should allow open-endedness. }
    \item -\textsc{H}: Route to the base model when the entropy of the base model's next token prediction distribution $H_{\text{base}}(y_t \mid x, y_{<t}) > \gamma ~_{\gamma \in [0, +\infty)}$, otherwise to the aligned model.
    \item -\textsc{H-a}: Route to the base model when the entropy of the aligned model's next token prediction distribution $H_{\text{aligned}}(y_t \mid x, y_{<t}) > \gamma ~_{\gamma \in [0, +\infty)}$, otherwise to the aligned model.
    
    \item -\textsc{PR}: Route to the base model when the ratio between the base model's top 1 token probability and the aligned model's, \ie{} $\frac{\max_{y_t} P_{\text{base}}(y_t \mid x, y_{<t})}{\max_{y_t} P_{\text{aligned}}(y_t \mid x, y_{<t})} < \gamma ~_{\gamma \in (0, 1]}$, otherwise to the aligned model.
    % instead of the absolute value, we use the ratio between the base model's top 1 token probability and the aligned model's. Route to the base model when $\frac{\max_{c_i} P_{\text{base}}(c_i \mid q, c_{<i})}{\max_{c_i} P_{\text{aligned}}(c_i \mid q, c_{<i})} < \gamma ~_{\gamma \in (0, 1]}$, otherwise to the aligned model.
    \item -\textsc{HR}: Route to the base model when the ratio between the base model's entropy and the aligned model's, \ie{} $\frac{H_{\text{base}}(c_i \mid q, c_{<i})}{ H_{\text{aligned}}(c_i \mid q, c_{<i})} > \gamma ~_{\gamma \in [1, +\infty)}$, otherwise to the aligned model.
    \item -\textsc{FC}: Route to the base model when both the aligned model and base model sample the next token is a content word\footnote{Long words could be composed by multiple tokens. If so, we will route to the same model multiple steps until the word is finished.}, otherwise to the aligned model.
    % Different from the above logits-based router, the function-content rule uses the content of the next token for routing. If the next word sampled by either the aligned model or base model is a function word, we route to the aligned model; if , we route to the base model.
    
    % \item -\textsc{Punc}: Simpler than -\textsc{FC}, we only consider if the aligned model's next token is punctuation or formative tokens (\eg{}  `$\backslash n$') or not. If so, we route to the aligned model; otherwise, to the base model.
    \item -\textsc{Punc}: Route to the base model when the base model's and the aligned model's next token is not punctuation or formative tokens (\eg{}  `$\backslash n$'), otherwise, to the aligned model.
    \item -\textsc{Judge}: Route to the base model when an external judge LLM (another aligned model) determines that both of the following conditions are satisfied: \textit{1}) the next token continuation has space to diverge; \textit{2}) the sampled base model continuation is acceptable (\ie{} reasonable and meaningful). Otherwise, to the aligned model. 
    
    % We serve an external LLM (another aligned model) as a judge for routing. We prompt it given the instruction $x$ and current content $c_{<i}$,  to judge \textit{1}) if the next token continuation has space to diverge, if so, route to the base model; then additional given the sampled candidate next token $c_{i}$ from the base model to judge \textit{2}) if the continuation is valid and acceptable, if not, discard it and route to the aligned model.  We prompt the judge model with heuristic rules and few-shot examples with rationals (\appref{}). This router, as a much costly strategy, serves as an extended comparison.

\end{itemize}

Comparatively, we observe that the aligned model's logits-based metrics are less distinctive, which aligns with the literature on entropy decrease, hence making routing strategies such as \textsc{-P-A} and \textsc{-H-A} less effective compared with the same metric under the base model's logits.  

\textsc{-FC} and \textsc{-Punc} fail under the same motivation of using content-based linguistic features as a routing strategy, where \textsc{-Punc} is more lightweight than \textsc{-FC}. 
From our empirical observation, the two have on-tier performance. 
However, the introduction of part-of-speech parsing for \textsc{-FC} takes additional computational cost. 
Given the cost limitation, some experiments and analyses take \textsc{-Punc} as representative.

We prompt -\textsc{Judge} with curated heuristic rules and few-shot examples with rationals. Detailed prompt designs are at \Cref{tab:judge-model-prompt}. 
As a more costly strategy, it serves as an extended comparison.

Following on, we have multi-condition routers, which are some possible combinations of the above single-condition routers:

\begin{itemize}
    \item -\textsc{P-FC}: First apply the \textsc{-FC} rule and then the -\textsc{P}. Route to the base model when any one of the following conditions is met: \textit{1}) base model sampled next token is a function word;  \textit{2}) both models' sampled next token are a content word; \textit{3}) base model's top-1 token probability $\max_{y_t} P_{\text{base}}(y_t \mid x, y_{<t}) < \gamma ~_{\gamma \in [0, 1]}$. Otherwise, to the aligned model. 
    \item -\textsc{P-Punc}: First apply the \textsc{-Punc} rule and then the -\textsc{P}. Route to the base model when any one of the following conditions is met: \textit{1}) base model sampled next token is not a punctuation or formatting tokens;  \textit{2}) base model's top-1 token probability $\max_{y_t} P_{\text{base}}(y_t \mid x, y_{<t}) < \gamma ~_{\gamma \in [0, 1]}$. Otherwise, to the aligned model. 
    \item -\textsc{H-FC}: First apply the \textsc{-FC} rule and then the -\textsc{H}. Route to the base model when any one of the following conditions is met: \textit{1}) base model sampled next token is a function word;  \textit{2}) both models' sampled next token are a content word; \textit{3}) base model's entropy of next token prediction distribution $H_{\text{base}}(y_t \mid x, y_{<t}) > \gamma ~_{\gamma \in [0, +\infty)}$. Otherwise, to the aligned model. 
    \item -\textsc{H-Punc}: First apply the \textsc{-Punc} rule and then the -\textsc{P}. Route to the base model when any one of the following conditions is met: \textit{1}) base model sampled next token is not a punctuation or formatting tokens;  \textit{2})  base model's entropy of next token prediction distribution $H_{\text{base}}(y_t \mid x, y_{<t}) > \gamma ~_{\gamma \in [0, +\infty)}$. Otherwise, to the aligned model. 
\end{itemize}

% =====================================================
% \section{Framework and Router Details}
% \label{app:framework_router}

% =====================================================
\section{Dataset Details}
\label{app:datasets}

\paragraph{NoveltyBench}
is a human-curated benchmark designed to evaluate the ability of LLMs to produce
multiple distinct yet high-quality outputs. The instructions are constructed
such that multiple valid answers exist, spanning four categories: randomness
(\eg{} ``the result of a die roll''), underspecified factual knowledge
(\eg{} ``tell me a capital city in Africa''), creative writing
(\eg{} ``short poem or story''), and subjective queries
(\eg{} ``recommendation or opinion'').
While effective for fine-grained diversity evaluation, NoveltyBench prompts are
intentionally simple and often yield short outputs with limited opportunities
for variation. We therefore complement it with more complex datasets.

\paragraph{WildChat}
is a large-scale dataset of real human--LLM conversations.
Following \citet{zhang2025noveltybench}, we select a subset of prompts without
fixed ground-truth answers to emphasize open-endedness, enabling evaluation
under more realistic and challenging settings.

\paragraph{Narrative-Discourse}
is a dataset for long-form creative writing, where models extend fictional film
synopses in English. The dataset provides structured annotations of discourse-level
elements such as turning points, story arcs, and arousal \citep{tian2024large},
enabling evaluation of long-form structural diversity.

% =====================================================
% \section{Evaluation Protocol}
% \label{app:evaluation}

\section{Automation Evaluation Details}
\label{app:auto_eval}
\subsection{Diversity Metrics}
\label{app:auto_eval:div}
We are mainly interested in diversity across a group of outputs.
For each prompt \(x\), we sample \(n=10\) outputs: \(\{y_0, \dots, y_{n-1}\}\). We evaluate the diversity of \(\{y_i\}\) using a broad set of automated metrics. Below are full derivations and definitions, grouped by category.

% \mina{they are not yet grouped by category}

\subsubsection{Lexical Metrics}

\paragraph{Distinct-\(n\).} Ratio of unique $n$-grams to total $n$-grams.   
Let \(\mathcal{G}_n = \bigcup_{i=0}^{n-1} \mathrm{ngrams}(y_i,n)\), tokenized by NLTK word-tokenize.
\[
\mathrm{Distinct}\text{-}n = \frac{|\mathrm{set}(\mathcal{G}_n)|}{|\mathcal{G}_n| + \varepsilon} \in [0,1]
\] 
Higher values indicate higher lexical diversity.

\paragraph{Expectation-Adjusted Distinct (EAD-\(n\)).}
A length- and vocabulary-normalized variant of Distinct-$n$, mitigating bias from long outputs. 
Define \(V\) as the $n$ power of the vocabulary size \(V\) of the aligned model's tokenizer and the union of all \(n\)-grams similarly by the aligned model's tokenizer. 
\[
V = \text{vocabulary size}^n
\]
\[
\mathrm{EAD}\text{-}n = \frac{|\mathrm{set}(\mathcal{G}_n)|}{V \cdot \left(1 - \left(\frac{V-1}{V}\right)^{|\mathcal{G}_n|}\right) + \varepsilon} \in [0,1]
\]
Higher values indicate higher lexical diversity. 
% Corrects length and vocabulary bias based on Distinct-\(n\).

\paragraph{Self-BLEU.} Average pairwise BLEU ~\citep{papineni2002bleu}. 
For each output \(y_i\), use the other outputs \(\{y_j\}_{j \neq i}\) as references:
\[
\mathrm{Self}\text{-}\mathrm{BLEU} = \frac{1}{n} \sum_{i=0}^{n-1} \mathrm{BLEU}(y_i, \{y_j\}_{j \neq i})
\in [0,1]
\]
Lower values indicate higher lexical diversity.

\paragraph{Self-ROUGE-L.}  Average pairwise ROUGE-L scores~\citep{lin2004rouge}.  
\[
\mathrm{Self}\text{-}\mathrm{ROUGE\text{-}L} = \frac{1}{n} \sum_{i=0}^{n-1}  \mathrm{ROUGE\text{-}L}(y_i, \{y_j\}_{j \neq i}) \in [0,1]
\]
Lower values indicate higher lexical diversity.

\subsubsection{Semantic Metrics}

\paragraph{Embedding Cosine Dissimilarity.}  
Embed each \(y_i\) using a sentence embedding model (e.g., SBERT or Qwen3), obtaining \(\mathbf{e}_i\). Compute pairwise cosine distances:
\[
d_{ij} = 1 - \cos(\mathbf{e}_i, \mathbf{e}_j)
\]
\[
\mathrm{Embedding\ Diversity} = \frac{2}{n(n-1)} \sum_{i<j} d_{ij} \in [0, 1]
\]
Higher values indicate higher semantic diversity.

\paragraph{Vendi Score.} 
The exponential entropy of eigenvalues of the similarity matrix based on n-gram Jaccard overlap, capturing the effective number of independent modes.
First, construct a similarity matrix \(K \in \mathbb{R}^{n \times n}\) via either n-gram Jaccard overlap or pairwise embedding similarity, which is positive semi-definite. Let \(\lambda_1, \dots, \lambda_n\) be the eigenvalues of \(\tfrac{K}{n}\). Then:
\[
\mathrm{Vendi\ Score} = \exp\left( -\sum_{i=1}^n \lambda_i \log \lambda_i \right) \in [1, n]
\]
This is the exponential of the Shannon entropy of the normalized similarity matrix, interpretable as the effective number of distinct modes. We construct similarity matrix  based on SimCSE embeddings \citep{gao2021simcse}. 
Higher values indicate higher semantic diversity.

% \subsection{Neural Metrics}

\paragraph{NLI Diversity.} Average contradiction probability across output pairs, computed using a RoBERTa NLI model.\footnote{\url{https://huggingface.co/sentence-transformers/nli-roberta-base-v2}}  
For each pair \((y_i, y_j)\), apply an NLI model (RoBERTa-based) to compute the entailment probability:
\[
\mathrm{NLI\ Diversity} = \frac{2}{n(n-1)} \sum_{i<j} P_{\mathrm{entailment}}(y_i, y_j) \in [0,1]
\]
Lower values (less entailment) indicate higher diversity.

% \subsection{Complex Metrics}

\paragraph{Distinct Score (NoveltyBench).} The number of unique functional equivalence classes predicted by a DeBERTa classifier trained on human annotation;
The DeBERTa classifier is trained to predict whether two outputs are functionally equivalent. Cluster the outputs $\{y_i\}$ equivalence classes. The metric is:
\[
\mathrm{Distinctivity\ Score} = \# \{\text{unique equivalence classes among } \{y_i\}\} \in [0, n-1]
\]
Larger values indicate higher diversity.

\paragraph{Semantic Entropy.} Rao’s quadratic entropy over clusters of semantically equivalent outputs grouped via entailment and aggregated via log-likelihood. It works by first 
clustering outputs \(\{y_i\}\) into semantic groups \(\{C_1, \dots, C_k\}\) using entailment-based NLI. Then compute cluster-level probabilities using likelihoods:
\[
\log p(C_k) = \log \left( \sum_{y_i \in C_k} \exp(\log p(y_i)) \right)
\]
Finally:
\[
\mathrm{Semantic\ Entropy} = -\sum_{k} p(C_k) \log p(C_k) \in [0, \log n]
\]
% Range \([0, \log n]\); l
Larger values indicate higher diversity.

\subsection{Aggregate Metrics over Quality--Diversity Spaces}
\label{app:aggregate-metrics}
% \mina{consistency}

Let a \emph{space} be defined by a pair of metrics \((m_x, m_y)\), where \(m_x\) measures quality (higher is better) and \(m_y\) measures diversity (either higher or lower is better, depending on the metric). Varying a method’s control parameter (e.g., decoding temperature, routing threshold) traces a set of points \(\{(x_t, y_t)\}\) in this space.

\paragraph{Feasible Region and Normalization.}
To make values comparable across metrics, we normalize each space to the unit square \([0,1]^2\).
Let \(\mathcal{F}=[x_{\min},x_{\max}]\times[y_{\min},y_{\max}]\) denote the feasible region, anchored using two reference operating points at temperature \(1.0\): the \emph{base} model and the \emph{aligned} model.\footnote{Concretely, \(x_{\min}\) is set to the base model’s quality at \(T=1.0\); \(x_{\max}\) to the aligned model’s quality at \(T=1.0\). For the diversity axis, if higher is better we set \(y_{\min}\) to the aligned model’s diversity at \(T=1.0\) and \(y_{\max}\) to the theoretical maximum (e.g., \(\log N\) for Semantic Entropy with \(N\) samples). If lower is better, we set \(y_{\min}\) to the theoretical lower bound and \(y_{\max}\) to the aligned model’s diversity at \(T=1.0\).}
Observed points are normalized via:
\[
\hat{x}=\frac{x-x_{\min}}{x_{\max}-x_{\min}},\qquad
\hat{y}=
\begin{cases}
\dfrac{y-y_{\min}}{y_{\max}-y_{\min}}, & \text{if higher is better},\\[6pt]
1-\dfrac{y-y_{\min}}{y_{\max}-y_{\min}}, & \text{if lower is better}.
\end{cases}
\]
Points outside \(\mathcal{F}\) are discarded for aggregation, as the outputs of the represented setting might have limited usage. It has no strength, in terms of the metrics, compared with the two single-model baselines.

\textbf{Coverage (Cov.)} measures how effectively a method traverses the \problem{} as its control parameters vary (\eg{} decoding temperature of single-model baselines and threshold for \name{} routers). 
The indicator is simplified from Hypervolume (HV) \citep{10.1007/s00158-016-1469-3} in multiobjective optimization problems. 
Concretely, we normalize each space into a unit square (anchored by the default baseline: base and aligned models at temperature 1.0), and compute the area under the curve (AUC) traced by the method’s normalized points. 
Higher Coverage values indicate greater controllability, general good performance across different trade-off balances, and robustness across parameters.
However, Coverage does not capture whether a method is \emph{ever} optimal across different trade-off balances.
% \paragraph{Coverageibility (Coverage).}
% Coverage quantifies a method’s controllability across the \problem{}. 

For method \(k\), we consider the piecewise-linear curve obtained from its normalized points \(\{(\hat{x}_t,\hat{y}_t)\}\) (ordered by \(\hat{x}\)), augmented with boundary points to close the curve inside \([0,1]^2\). We define:
\[
\mathrm{Coverage}_k(m_x,m_y)=\int_{0}^{1}\hat{y}_k(\hat{x})\,d\hat{x},
\]
computed using the trapezoidal rule. Because the domain is fixed to \([0,1]\), \(\mathrm{Coverage}\in[0,1]\). Higher values indicate that the method maintains strong quality and diversity as its control parameters vary.

\textbf{Dominance (Dom.)} complements Coverage by capturing whether a method ever achieves optimality relative to others. We utilize the C-metric \citep{zitzler1999evolutionary} to evaluate Dominance of pairwise comparison, which captures the portion of the frontier that one method dominates over the other one. 
In our problem, the portion is in terms of intervals along the diversity (denoted as -D) or quality (denoted as -Q) axes. Dom takes the harmonic mean of Dom-D and Dom-Q.
For global comparison across all methods, we compute the global Pareto frontier across all methods. We apply the C-metric between each method and the global frontier, equivalent to the portion of the frontier attributed to the method.  

% \paragraph{Dominance (Dom).}
% Coverage does not capture whether a method is \emph{ever} optimal. 
We compute the global Pareto frontier \(\mathcal{P}\) over the union of all methods’ normalized points in a space. For each Pareto point, we assign an \emph{interval of responsibility} along the diversity or quality axis by splitting at midpoints between adjacent frontier points. Summing these interval lengths for Pareto points contributed by method \(k\) yields its coverage along that axis, normalized by the total frontier span:
\[
\mathrm{Dom\text{-}D}_k(m_x,m_y),\quad
\mathrm{Dom\text{-}Q}_k(m_x,m_y)\;\in[0,1].
\]
We report a single Dominance score as their {harmonic mean}:
\[
\mathrm{Dom}_k(m_x,m_y)=
% \frac{2}{\tfrac{1}{\mathrm{Dom\text{-}Q}_k}+\tfrac{1}{\mathrm{Dom\text{-}D}_k}}.
\frac{2 \cdot \mathrm{Dom\text{-}D}_k \cdot \mathrm{Dom\text{-}Q}_k}{\mathrm{Dom\text{-}D}_k + \mathrm{Dom\text{-}Q}_k}
\]

\paragraph{Holistic aggregation.}
Since quality and diversity admit multiple measurements, we average over all spaces \(\mathcal{S}\) to obtain metric-agnostic summaries:
\[
\overline{\mathrm{Cov}}_k = \frac{1}{|\mathcal{S}|}\sum_{(m_x,m_y)\in\mathcal{S}}\mathrm{Cov}_k(m_x,m_y),\qquad
\overline{\mathrm{Dom}}_k = \frac{1}{|\mathcal{S}|}\sum_{(m_x,m_y)\in\mathcal{S}}\mathrm{Dom}_k(m_x,m_y).
\]
% Figure~\ref{fig:main} reports \(\overline{\mathrm{Coverage}}\) and \(\overline{\mathrm{Dom}}\) for all methods. \thh{dont forget to fix}
% \subsection{Human Evaluation Details}

% =====================================================
\section{Experiment Setup}
\label{app:exp_setup}

% \paragraph{Inference Setup.}
% Our study focuses on group-level diversity.
% For each prompt, we generate a group of $n=10$ outputs.
% Unless otherwise specified, sampling is performed with a temperature of $1.0$
% and nucleus sampling (top-$p$) with $p=0.9$, applied consistently across all methods.
% Exceptions apply only to baselines that inherently require alternative decoding
% strategies.
\paragraph{Inference Setup.}
Our study focuses on group-level diversity.
For each prompt, we generate a group of $n=10$ outputs.
Unless otherwise specified, sampling uses temperature $1.0$ and nucleus sampling (top-$p$) with $p=0.9$ and no top-$k$ truncation, following \citet{zhang2025noveltybench}.
For Diverse Beam Search \citep{vijayakumar2016diverse}, we use $\text{beams}{=}2n{=}20$, $\text{beam\_groups}{=}n{=}10$, and $\text{diversity\_penalty}{=}1.0$, following the original paper.
Other exceptions apply only to baselines that inherently require alternative decoding strategies.

% =====================================================
\section{Validation on Verifiable Benchmarks}
\label{app:verifiable_tasks}

This section reports detailed results on two verifiable benchmarks used to
validate that the improvements of \name{} are not artifacts of open-ended
evaluation metrics: verifiable instruction following (IFEval) and mathematical
reasoning (GSM8K).

\paragraph{Verifiable instruction following (IFEval).}
IFEval provides instruction-following tasks with automatically verifiable
constraints \citep{zhou2023instructionfollowingevaluationlargelanguage}.
We evaluate instruction-following accuracy as a quality metric alongside diversity metrics.
As shown in \Cref{fig:ifeval_large_figure}, at matched quality levels,
\name{} achieves consistently higher diversity than the aligned baseline.

\paragraph{Mathematical reasoning (GSM8K).}
For mathematical reasoning, we evaluate on GSM8K \citep{cobbe2021training} and
measure Acc and Pass@10 as the accuracy metrics.
Diversity is computed over sampled solution outputs.
As shown in \Cref{fig:gsm8k_large_figure}, \name{} maintains high accuracy (e.g., ~90\% Pass@10) while
achieving substantially higher diversity across a wide range of operating points.

Overall, the qualitative trends on both benchmarks mirror those observed on
open-ended generation tasks, confirming that the gains of \name{} generalize
beyond open-ended evaluation metrics.

% =====================================================
\section{Detailed Results on Open-ended Tasks}
\label{app:additional_results}

% \section{Detailed Results}
\label{app:detail_results}

\subsection{Instruction Following on NoveltyBench}
\label{app:detail_results:nb}

\Tabref{tab:noveltybench} shows the result on NoveltyBench of comparing \name{} on the best router compared with baselines. 

% \begin{wraptable}{r}{0.7\textwidth}
\begin{table}[H]
\centering
\small
\renewcommand{\arraystretch}{1.2}

\begin{tabular}{lcccccc}
\toprule
\textbf{Method} & \multicolumn{2}{c}{\textbf{Lexical}} & \multicolumn{2}{c}{\textbf{Semantic}} & \multicolumn{2}{c}{\textbf{Overall}} \\
 & \textit{Cov.} & \textit{Dom.} & \textit{Cov.} & \textit{Dom.} & \textit{Cov.} & \textit{Dom.} \\
 \midrule
Base & 0.142 & 9.8\% & 0.142 & 13.1\% & 0.142 & 11.4\% \\
Aligned & 0.273 & \textbf{40.1\%} & 0.128 & 17.2\% & 0.200 & 28.6\% \\
Nudging & 0.192 & 6.8\% & 0.161 & 7.6\% & 0.176 & 7.2\% \\
Decoding & - & 0.8\% & - & 1.0\% & - & 0.9\% \\
Prompting best & - & 8.0\% & - & 6.5\% & - & 7.3\% \\
Ensemble best & - & 3.4\% & - & 5.8\% & - & 4.6\% \\
\midrule
\name{} best & \textbf{0.495} & 31.0\% & \textbf{0.452} & 
\textbf{48.8\% }& \textbf{0.474} & \textbf{39.9\%} \\

% Base & 0.142 & 9.8\% & 0.142  & 13.1\% & 0.142 & 11.4\% \\
% Aligned & 0.273 & \textbf{36.2\%} & 0.128 & 15.8\% & 0.200 & 26.0\% \\
% Nudging & 0.192 & 6.7\% & 0.161 & 6.0\% & 0.176 & 6.3\% \\
% Other baselines & - & 7.4\% & - & 11.1\% & - & 9.3\% \\
% \midrule
% \name{} best & \textbf{0.495} & 31.0\% & \textbf{0.452} & \textbf{44.8\% }& \textbf{0.474} & \textbf{37.9\%} \\
\toprule
\end{tabular}
\caption{Comparison results on NoveltyBench. For space-saving, we present the best method in each category.}
\label{tab:noveltybench}
% \end{wraptable}
\end{table}

\textbf{Results.} \name{} outperforms all baselines on all metrics except lexical Dominance. 
Compared with all baselines, \name{} improves Coverage by \textbf{0.274} overall (0.222 lexical, 0.291 semantic).
It dominates \textbf{39.9\%} (the most) of the diversity-quality frontier overall (31.0\% lexical, 48.8\% semantic). 
% Compared with the aligned model baseline, \name{}-\textsc{P-Punc} improves controllability (Flex) by 29.8\% and dominates 30.8\% of the frontier. 

Beyond the LLaMA-3 base–aligned pair, we also validate that \name{} consistently outperforms baselines on another model family, \texttt{Olmo2}. The results are reported in \Cref{tab:nb:olmo2}.

\subsection{Dialogue on WildChat}

% As WildChat naturally involves more complex and nuanced prompts, its average response length is much longer than Novelty-Bench's. The results are in \Tabref{tab:wildchat}.

WildChat involves naturally complex and nuanced prompts, leading to much longer outputs on average compared with NoveltyBench. \Cref{tab:wildchat} summarizes the results.  

% \begin{wraptable}{r}{0.7\textwidth}
\begin{table}[H]
\setlength{\intextsep}{0pt}
\captionsetup[table]{skip=2pt,aboveskip=2pt,belowskip=0pt}
\centering
\small
\renewcommand{\arraystretch}{1.2}
\begin{tabular}{l cc cc cc cc}
\toprule
\textbf{Method} & \multicolumn{2}{c}{\textbf{Lexical}}& \multicolumn{2}{c}{\textbf{Semantic}} & \multicolumn{2}{c}{\textbf{Overall}} \\
 & \textit{Cov.} & \textit{Dom.} & \textit{Cov.} & \textit{Dom.} & \textit{Cov.} & \textit{Dom.} \\
\midrule
Base & 0.000	& 1.9\%	& 0.000	& 6.8\%	& 0.000	& 4.38\% \\
Aligned & 0.253	& \textbf{59.2\%}	& 0.077	& 29.1\%	& 0.165	& \textbf{44.1\%}\\
Nudging & 0.430	& 11.4\%	& 0.387	& 15.6\%	& 0.408	& 13.5\%  \\
% Other baselines & - & 0.0\% & - & 0.0\% & - & 0.0\% \\

\midrule
\name{} best & \textbf{0.473}	& 27.4\%	& \textbf{0.454} & \textbf{48.5\%}	& \textbf{0.463}	& 38.0\% \\
\toprule
\end{tabular}
\caption{Comparison results on WildChat. For space saving, we present the best router, -\textsc{P-Punc}, as \name{}'s representative.}
\label{tab:wildchat}
\end{table}

\textbf{Results.} The superiority of \name{} persists on WildChat. Compared with the aligned model baseline, \name{}-\textsc{P-Punc} improves Coverage by 29.8\% and dominates 30.8\% of the frontier. 
Moreover, \name{} demonstrates a particularly strong advantage in semantic diversity, where it dominates 48.5\% of the frontier. 
These findings confirm that base–aligned collaboration scales effectively from short-form prompts (NoveltyBench) to longer, more conversational dialogue.

\subsection{Creative Writing on Narrative-Discourse}

% We extend our study to creative writing on Narrative-Discourse, further evaluating whether \name{} can generate structure-diverse and long-term coherent narratives.

We further evaluate \name{} on Narrative-Discourse to test its ability to generate structure-diverse and long-term coherent narratives. 
This dataset emphasizes sustained creativity and narrative arc, placing distinct demands beyond instruction following and dialogue. 

% \begin{wraptable}{r}{0.7\textwidth}
\begin{table}
\centering
\small
\renewcommand{\arraystretch}{1.2}

\begin{tabular}{lcccccc}
\toprule
\textbf{Method} & \multicolumn{2}{c}{\textbf{Lexical}} & \multicolumn{2}{c}{\textbf{Semantic}} & \multicolumn{2}{c}{\textbf{Overall}} \\
 & \textit{Cov.} & \textit{Dom.} & \textit{Cov.} & \textit{Dom.} & \textit{Cov.} & \textit{Dom.} \\
\midrule
Base & 0.151 & 26.3\% & 0.153 & 28.1\% & 0.152 & 27.2\% \\
Aligned & 0.282 & \textbf{47.7\%} & 0.106 & \textbf{41.2\% }& 0.194 & \textbf{44.4\%} \\
Nudging & 0.205 & 9.7\% & 0.194 & 6.5\% & 0.199 & 8.1\% \\
% Other baselines & 0.000 & 0.0\% & 0.000 & 0.0\% & 0.000 & 0.0\% \\
\midrule
\name{} best & \textbf{0.367} & 16.3\% & \textbf{0.174} & 24.2\% & \textbf{0.271} & 20.3\% \\
\bottomrule
\end{tabular}

\caption{Comparison results on Narrative-Discourse. For space saving, we present the best router, -\textsc{P-Punc}, as \name{}'s representative.}
% \caption{Results on Narrative-Discourse. For space saving, we present the best router as BACO’s representative. BACO-P-PUNC serves as the best.}
\label{tab:storyarc}
\end{table}

\textbf{Results.} 
As shown in \Cref{tab:storyarc}, \name{} again outperforms all baselines. It achieves 13.5\% higher Coverage and dominates 20.3\% of the overall diversity–quality frontier.

\begin{table}[]
    \small
    \centering
    \renewcommand{\arraystretch}{1.2}
\begin{tabular}{lcccccc}

\toprule
\textbf{Method} & \multicolumn{2}{c}{\textbf{Lexical}} & \multicolumn{2}{c}{\textbf{Semantic}} & \multicolumn{2}{c}{\textbf{Overall}} \\
 & \textit{Cov.} & \textit{Dom.} & \textit{Cov.} & \textit{Dom.} & \textit{Cov.} & \textit{Dom.} \\
\toprule
Base & 0.142 & 7.6\% & 0.142 & 10.9\% & 0.142 & 9.2\% \\
Aligned & 0.273 & 36.5\% & 0.128 & 15.8\% & 0.200 & 26.1\% \\
\midrule
In-context Prompt & - & 0.0\% & - & 2.2\% & - & 1.1\% \\
Paraphrase Prompt & - & 8.0\% & - & 5.6\% & - & 6.8\% \\
Diverse BS Decoding & - & 0.8\% & - & 1.0\% & - & 0.9\% \\
\midrule
Response Ensemble & - & 3.4\% & - & 3.7\% & - & 3.6\% \\
Logits Ensemble & - & 0.0\% & - & 0.0\% & - & 0.0\% \\
Nudging & 0.192 & 4.9\% & 0.161 & 4.0\% & 0.176 & 4.5\% \\
\midrule
\name{} All & 0.495 & 39.0\% & 0.452 & 56.8\% & 0.474 & 47.9\% \\
\midrule
\midrule
\name{}-\textsc{Judge} & 0.302 & 0.5\% & 0.254 & 0.1\% & 0.278 & 0.3\% \\
\name{}-\textsc{Rand} & 0.493 & 13.1\% & 0.409 & 5.9\% & 0.451 & 9.5\% \\
\midrule
\name{}-\textsc{FC} & 0.419 & 2.4\% & 0.382 & 4.0\% & 0.401 & 3.2\% \\
\name{}-\textsc{P} & 0.433 & 2.6\% & 0.397 & 7.7\% & 0.415 & 5.2\% \\
\midrule
\name{}-\textsc{P-Punc} & 0.495 & 11.4\% & 0.452 & 17.2\% & 0.474 & 14.3\% \\
\name{}-\textsc{H-Punc} & 0.466 & 5.9\% & 0.427 & 11.0\% & 0.446 & 8.4\% \\
\name{}-\textsc{P-FC} & 0.435 & 3.1\% & 0.406 & 10.9\% & 0.421 & 7.0\% \\

\toprule
\end{tabular}
    \caption{Comparison of all methods (baselines and \name{} routers) on NoveltyBench. \name{} All reports the best \textit{Cov.} across all routers and the \textit{Dom.} sum over all routers. The lower half of the table provides the performance of individual routers. Note that routers distribute \textit{Dom.} values given to the metric definition; therefore, \textit{Dom.} values in the top and bottom halves of the table are not directly comparable.}
    % \caption{Comparison results of all methods (baselines and \name{} routers) on Novelty-Bench. \name{} All presents the best \textit{Flex} across all routers and the \textit{Domn} sum of all routers. The bottom half of the table details each router's performance. Please note that similar routers would distribute the \textit{Domn} given the metric definition. Hence, it is not reliable to compare the \textit{Domn} absolute value between lines in the top half with lines in the bottom half.}
    \label{tab:novelty-bench-full}
\end{table}

\subsection{Extensional Tasks Results}

\begin{figure}
    \centering
    \includegraphics[width=1.0\linewidth]{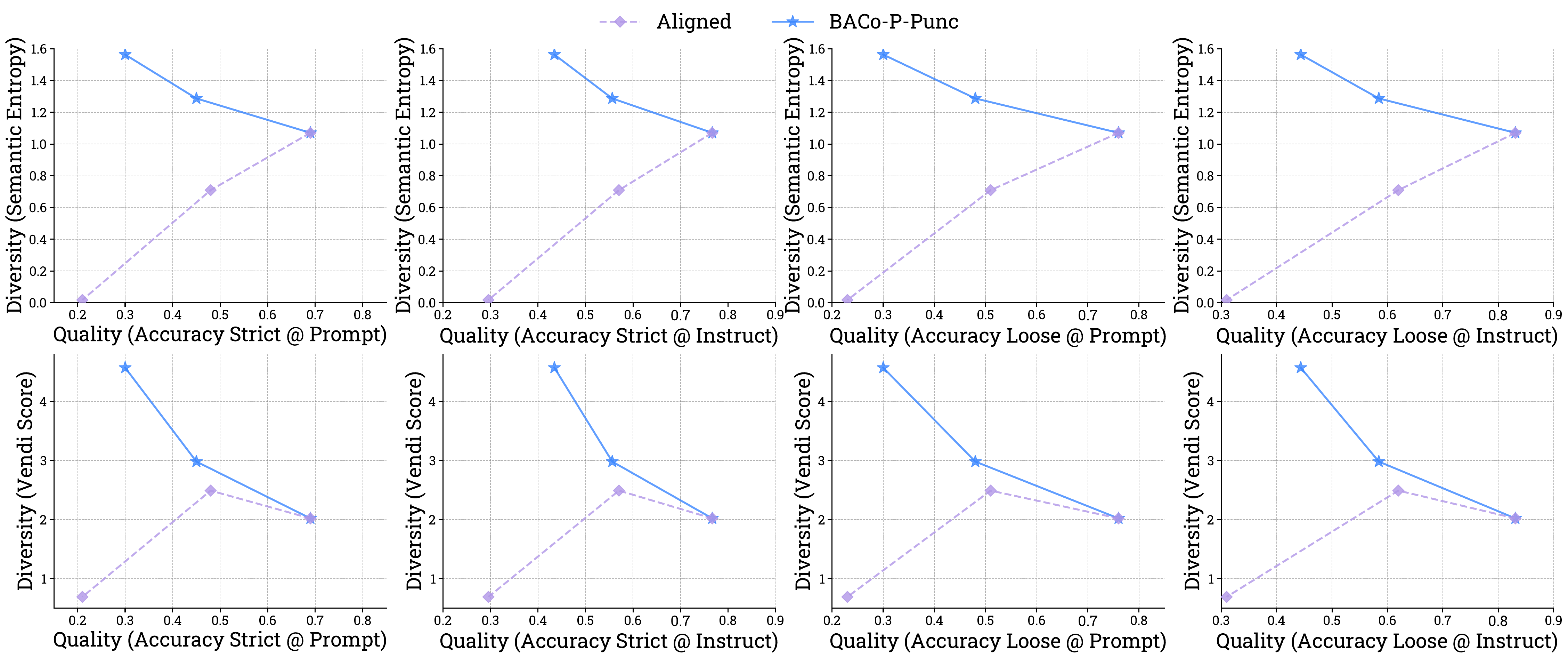}
    \caption{Diversity-quality trade-off comparison between \name{} and the aligned model baseline on IFEval. \name{} demonstrates a more effective optimization of the trade-off. The table presents all the trade-off spaces tested.}
    \label{fig:ifeval_large_figure}
\end{figure}

\begin{figure}
    \centering
    \includegraphics[width=0.5\linewidth]{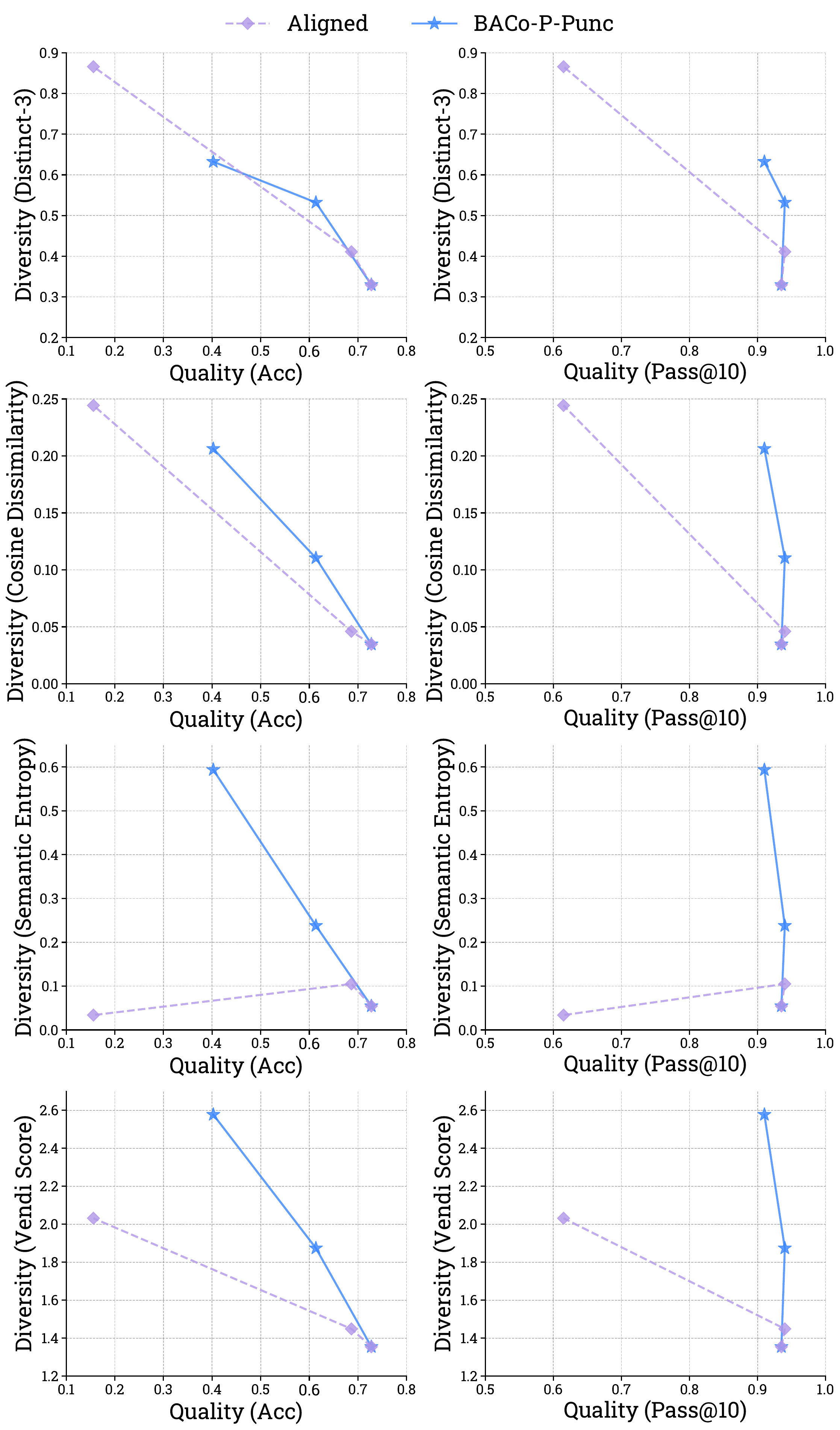}
    \caption{Diversity-accuracy trade-off comparison between \name{} and the aligned model baseline on GSM8K. \name{} demonstrates a more effective optimization of the trade-off. The table presents all the trade-off spaces tested.}
    \label{fig:gsm8k_large_figure}
\end{figure}

We compare \name{}\textsc{-P-Punc} (i.e., our best router) with the aligned model baseline by adjusting temperature.
\Cref{fig:ifeval_large_figure} presents the full results of the verifiable instruction following task on the IFEval dataset, and \Cref{fig:gsm8k_large_figure} presents the full results of the mathematical reasoning task on the GSM8K dataset.

% \begin{wrapfigure}{h}{0.48\linewidth}
% % \begin{figure}
% \vspace{-25pt}
%     \centering
%     \includegraphics[width=\linewidth]{figures/rebutal/ifeval_small_figure.pdf}
% \caption{Diversity-quality trade-off comparison on IFEval. \name{} demonstrates a more effective optimization of the trade-off. Refer full results to \Cref{xx}.}
% % \vspace{-30pt}
% \label{fig:gsm8k_small_figure}
% % \end{figure}

% % \vspace{-25pt}
%     \centering
%     \includegraphics[width=\linewidth]{figures/rebutal/gsm8k_small_figure.pdf}
% \caption{Diversity-accuracy trade-off comparison on GSM8K. \name{} demonstrates a more effective optimization of the trade-off. Refer full results in \Cref{xx}.}
% \vspace{-30pt}
% \label{fig:gsm8k_small_figure}
% % \end{figure}

% \begin{table}[]
%     \small
%     \centering
%     \renewcommand{\arraystretch}{1.2}
% \begin{tabular}{lcccccc}

% \toprule
% \textbf{Method} & \multicolumn{2}{c}{\textbf{Lexical}} & \multicolumn{2}{c}{\textbf{Semantic}} & \multicolumn{2}{c}{\textbf{Overall}} \\
%  & \textit{Cov.} & \textit{Dom.} & \textit{Cov.} & \textit{Dom.} & \textit{Cov.} & \textit{Dom.} \\
% \toprule

% Base & 0.094 & 0.0\% & 0.101 & 0.0\% & 0.098 & 0.0\% \\
% Aligned & 0.295 & 0.0\% & 0.137 & 0.0\% & 0.216 & 0.0\% \\
% \midrule
% Nudging & 0.299 & 19.1\% & 0.266 & 10.6\% & 0.283 & 14.9\% \\
% \name{} & 0.242 & 80.9\% & 0.231 & 89.4\% & 0.237 & 85.1\% \\

\begin{table}[ht]
\centering
\small
\renewcommand{\arraystretch}{1.2}
\begin{tabular}{lcccc}
\toprule
\textbf{Method} & \textbf{Overall} \textit{Cov.} & \textbf{Overall} \textit{Dom.} \\
\midrule
Base & 0.098 &  11.7\% \\
Aligned  & 0.209 &  \underline{30.5\%} \\
Nudging & \textbf{0.281} &  5.5\% \\
Others & - & 8.3\% \\
\midrule
\name-\textsc{P-Punc} & \underline{0.236} & \textbf{44.0\%} \\
\bottomrule
% \toprule
% \textbf{Method} & \textbf{Flex} & \textbf{Domn-Q} & \textbf{Domn-D} & \textbf{Domn} \\
% \midrule
% Base & 0.0978 & 16.75\% & 10.38\% & 11.74\% \\
% Aligned  & 0.2090 & \underline{29.87\%} & \underline{39.15\%} & \underline{30.53\%} \\
% Nudging & \textbf{0.2810} & 4.76\% & 7.42\% & 5.52\% \\
% \midrule
% \name-\textsc{P-Punc} & \underline{0.2363} & \textbf{48.62\%} & \textbf{43.05\%} & \textbf{43.95\%} \\
% \bottomrule
\end{tabular}
\caption{Results of \name{} on \texttt{Olmo2-7B} and \texttt{Olmo2-7B-Instruct} on NoveltyBench.}
\label{tab:nb:olmo2}
\end{table}

% =====================================================
\section{Narrative Discourse Evaluation Details}
\label{app:storyarc}

\subsection{Creative Writing Task Setup}

\label{app:storyarc}

We frame creative writing as a continuation task, where the model is given the beginning of a story and asked to complete it.
The prefix contains events leading up to the first turning point, which introduces the initial situation or conflict setting the stage for the narrative.
The model then generates the subsequent events to develop and conclude the entire narrative.
% This setup allows us to evaluate how well different methods produce coherent and structurally diverse continuations beyond the shared beginning.
% \mina{I can't understand what this means based on reading section 5.4; why continuation task? what's prefix? what's turning point 1? what's remaining story?
% the paper needs to be self-contained even with citations. in other words, you shouldn't expect readers to go off and read many other papers in order to understand this work; try to provide enough necessary information so that people can understand what you did here}
To capture discourse-level variation, we measure two structural dimensions:  

\circone~ \emph{Turning-point diversity} quantifies differences in the relative positions of annotated plot inflections across outputs.  

\circtwo~ \emph{Arousal diversity} tracks divergence in emotional trajectories, obtained by sampling sentence-level arousal scores and comparing smoothed curves via KL divergence.

Together, these metrics provide complementary measures of long-form diversity, capturing variation in plot structure and affective dynamics that conventional surface-level metrics miss. 
% Detailed setups are provided in \appref{app:storyarc}.
Prompt is shown in \Cref{tab:storyarc-prompt}. Details of the dataset and annotation schema follow \citet{tian2024large}.

% % \paragraph{.}
% We cast the task as narrative continuation: given a prefix consisting of the initial \emph{Opportunity} section (up to Turning Point 1), the model must continue the story until completion (covering the next four turning points). 
% and are summarized in \appref{app:storyarc}.  

\begin{table}[H]
\small
\begin{tabular}{M{0.95\linewidth}}
\toprule
{\raggedright\ttfamily
``role'': ``user'', ``content'': Continue the story and bring it to an ending based on the title and the story sketch provided below. The sketch introduces the event that sets the initial stage for the narrative leads up to the first major turning point—but does not present a full plot. Your task is to develop the narrative from this point onward, completing the story arc. \\
Title: \{title\} \\
Story Sketch: \{sketch\}
} \\
\bottomrule \\
\end{tabular}
\caption{Generation prompt for the creative writing task.}
\label{tab:storyarc-prompt}
\end{table}

% \begin{table}[H]
% \small
% \begin{tabular}{M{0.95\linewidth}}
% \toprule
% \texttt{``role'': ``user'', ``content'': Continue the story and bring it to an ending based on the title and the story sketch provided below. The sketch introduces the event that sets the initial stage for the narrative leads up to the first major turning point—but does not present a full plot. Your task is to develop the narrative from this point onward, completing the story arc. \\
% Title: \{title\} \\
% Story Sketch: \{sketch\} \\
% } 
% \bottomrule
% \end{tabular}
% \caption{Generation prompt for the creative writing task.}
% \label{tab:storyarc-prompt}
% \end{table}

\subsection{Narrative-Discourse Evaluation Metrics}
For \textit{turning points}, each generated narrative \(y\) is segmented into sentences, with total length \(L\). The relative position of the turning point \(k\) annotated is
$r_k\!\left(y\right) = \frac{\text{Index}_{tp_k}\!\left(y\right)}{L}, \quad r_k\!\left(y\right) \in [0,1].
$
For a group of $n$ outputs \(\{y^{(1)}, \dots, y^{(n)}\}\), we compute pairwise distances:
\[
D_{\text{TP}}\!\left(y^{(i)}, y^{(j)}\right) = \frac{1}{K} \sum_{k=1}^K \left| r_k\!\left(y^{(i)}\right) - r_k\!\left(y^{(j)}\right) \right|,
\]
where \(K=5\) is the number of turning points. The turning-point diversity score is then
\[
\text{TP-Div} = \frac{2}{n(n-1)} \sum_{i<j} D_{\text{TP}}\!\left(y^{(i)}, y^{(j)}\right).
\]

For \textit{arousal}, we sample sentences at fixed intervals from each \(y^{(i)}\) and obtain arousal scores via LLM-as-a-judge. Let \(a_t\!\left(y^{(i)}\right)\) denote the arousal score at sampled position \(t\). We fit a smooth trajectory \(\hat{a}\!\left(y^{(i)}\right)\) via polynomial interpolation. For two narratives \(y^{(i)}\) and \(y^{(j)}\), their affective divergence is
\[
D_{\text{Arousal}}\!\left(y^{(i)}, y^{(j)}\right) = \mathrm{KL}\!\left(\hat{a}\!\left(y^{(i)}\right) \;\|\; \hat{a}\!\left(y^{(j)}\right)\right).
\]
The overall arousal diversity is
\[
\text{Arousal-Div} = \frac{2}{n(n-1)} \sum_{i<j} D_{\text{Arousal}}\!\left(y^{(i)}, y^{(j)}\right).
\]

\section{Additional Analysis Material}
\label{app:analysis}

\begin{figure}[t]
    \begin{subfigure}[t]{0.45\textwidth}
        \includegraphics[width=\linewidth]{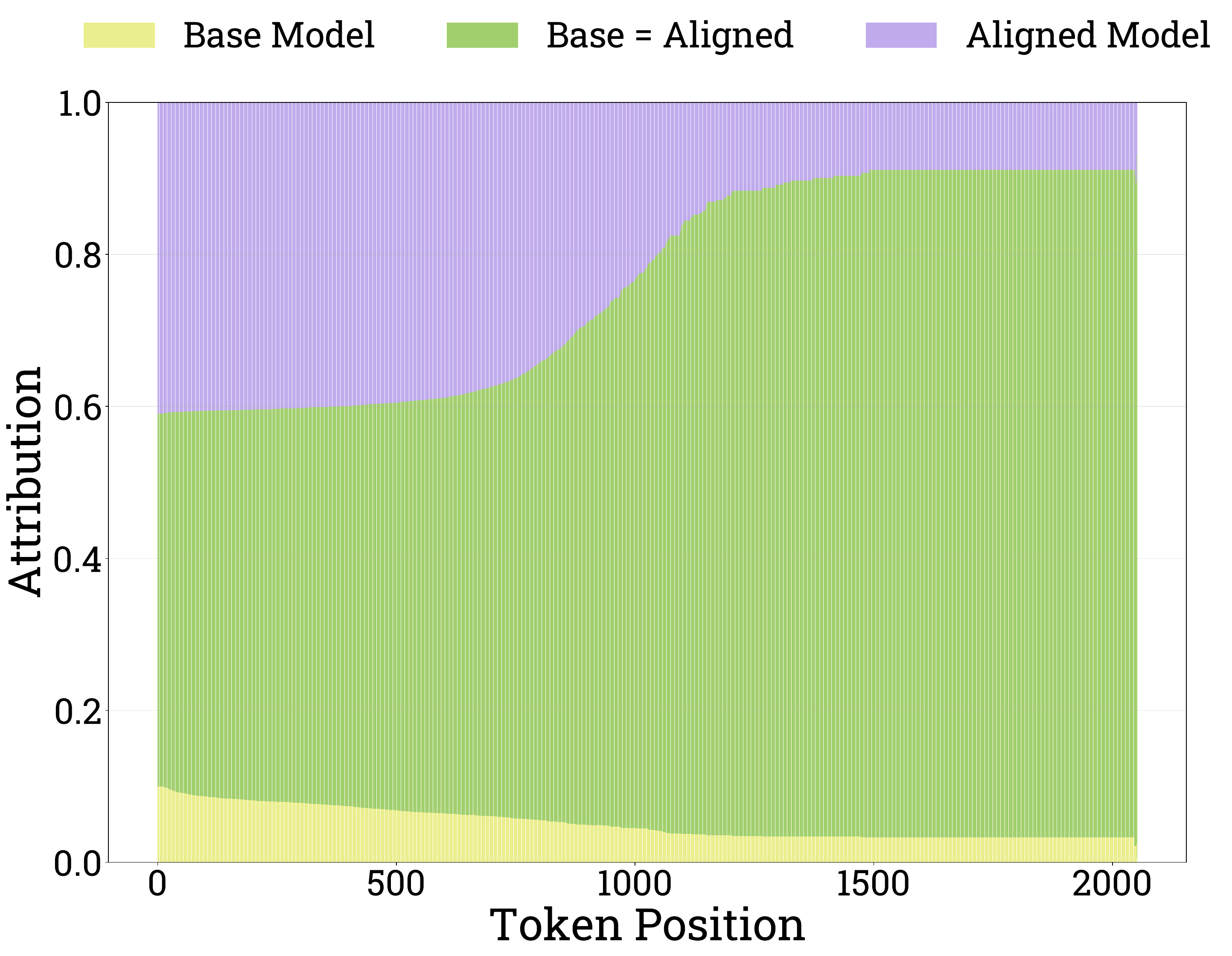}
        \caption{Model Contribution Distribution}
    \end{subfigure}
    \begin{subfigure}[t]{0.45\textwidth}
        \includegraphics[width=\linewidth]{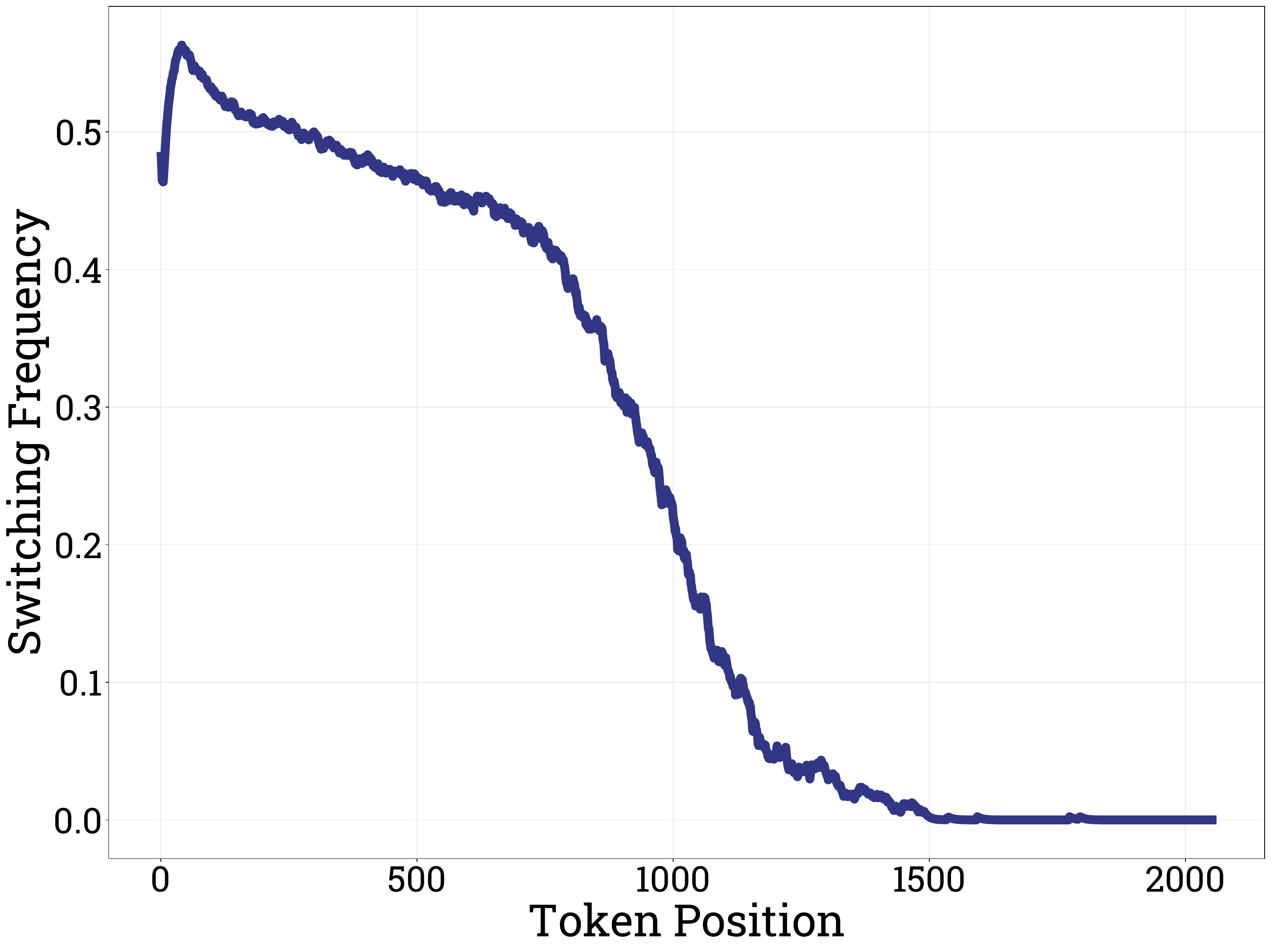}
        \caption{Switching Frequency}
    \end{subfigure}
    \caption{Contribution distribution and switching frequency for \name{} with
    the best router (\textsc{-P-Punc}) at Narrative-Discourse dataset.}
    \label{fig: switching_storyarc}
\end{figure}

\begin{figure}[t]
    \begin{subfigure}[t]{0.45\textwidth}
        \includegraphics[width=\linewidth]{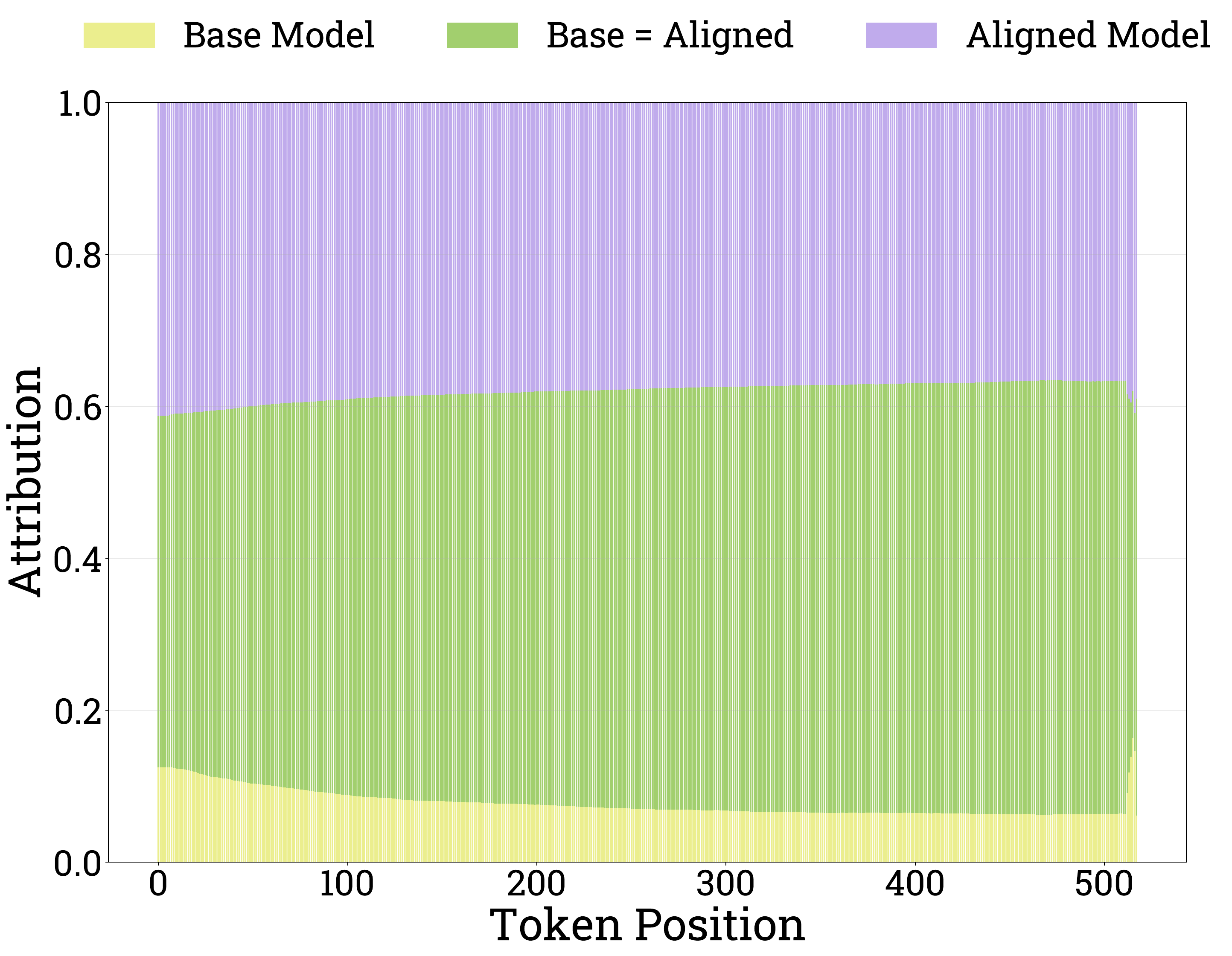}
        \caption{Model Contribution Distribution}
    \end{subfigure}
    \begin{subfigure}[t]{0.45\textwidth}
        \includegraphics[width=\linewidth]{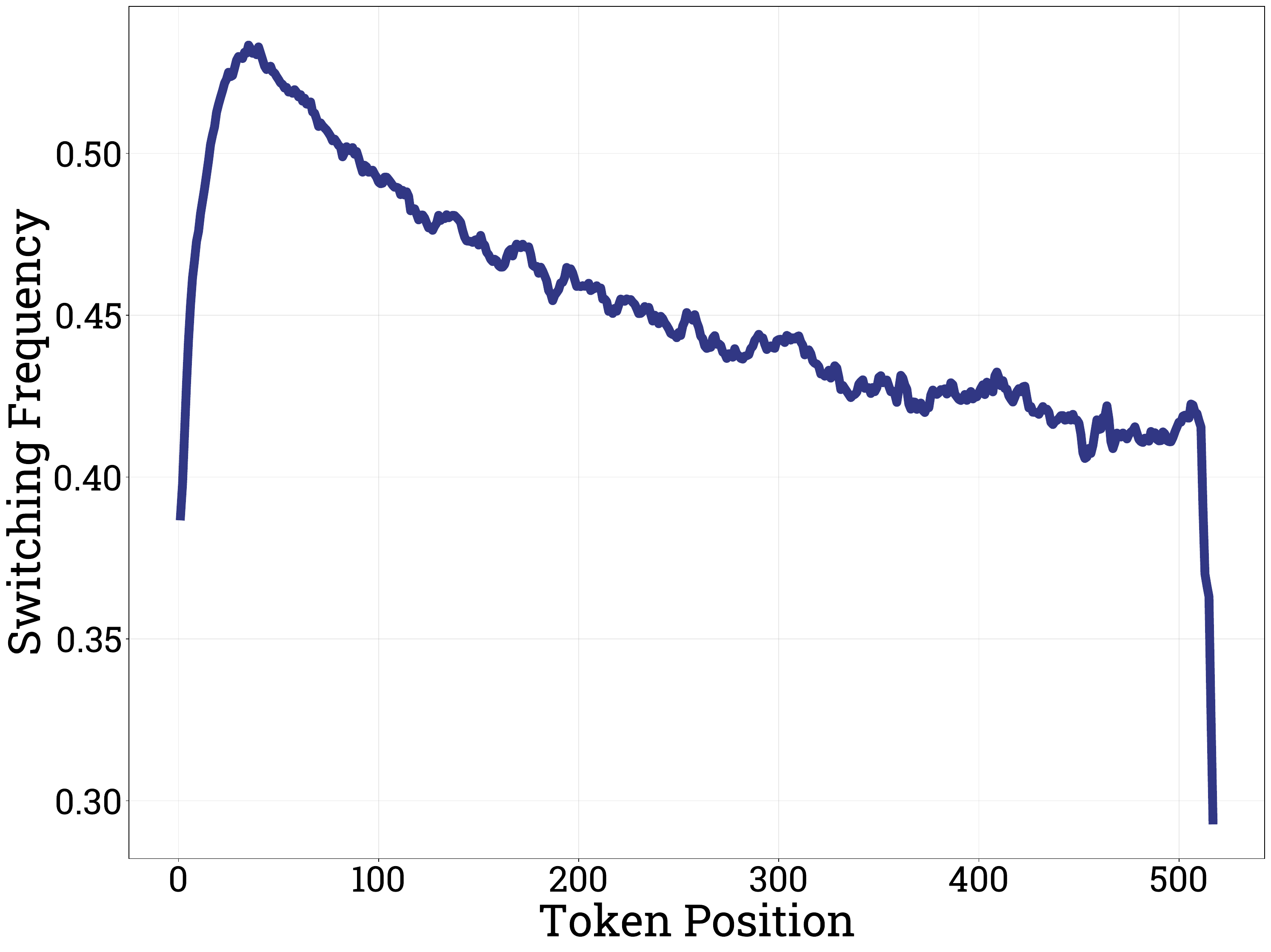}
        \caption{Switching Frequency}
    \end{subfigure}
    \caption{Contribution distribution and switching frequency for \name{} with
    the best router (\textsc{-P-Punc}) at WildChat dataset.}
    \label{fig: switching_wildchat}
\end{figure}

\begin{figure}[t]
    \begin{subfigure}[t]{0.45\textwidth}
        \includegraphics[width=\linewidth]{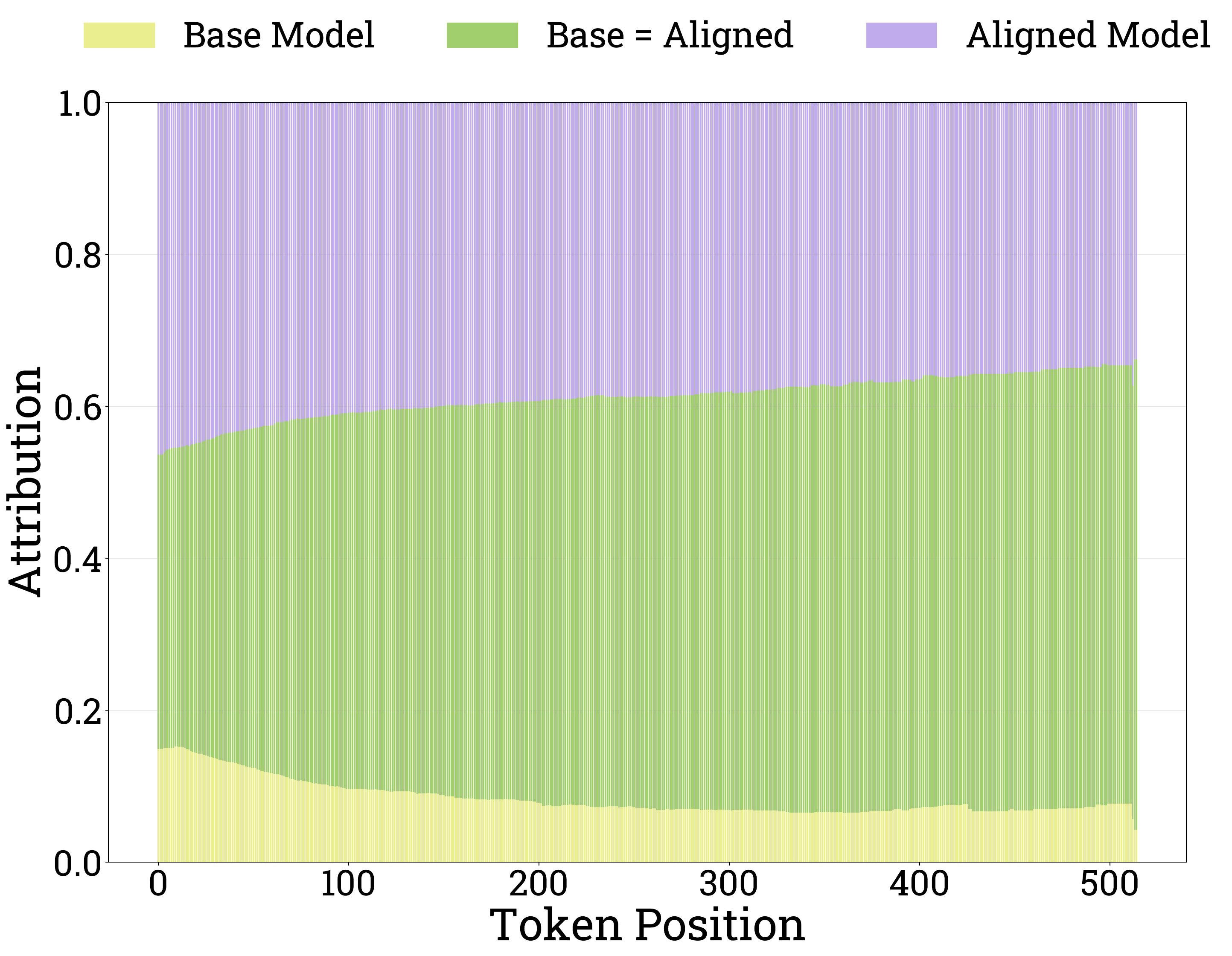}
        \caption{Model Contribution Distribution}
    \end{subfigure}
    \begin{subfigure}[t]{0.45\textwidth}
        \includegraphics[width=\linewidth]{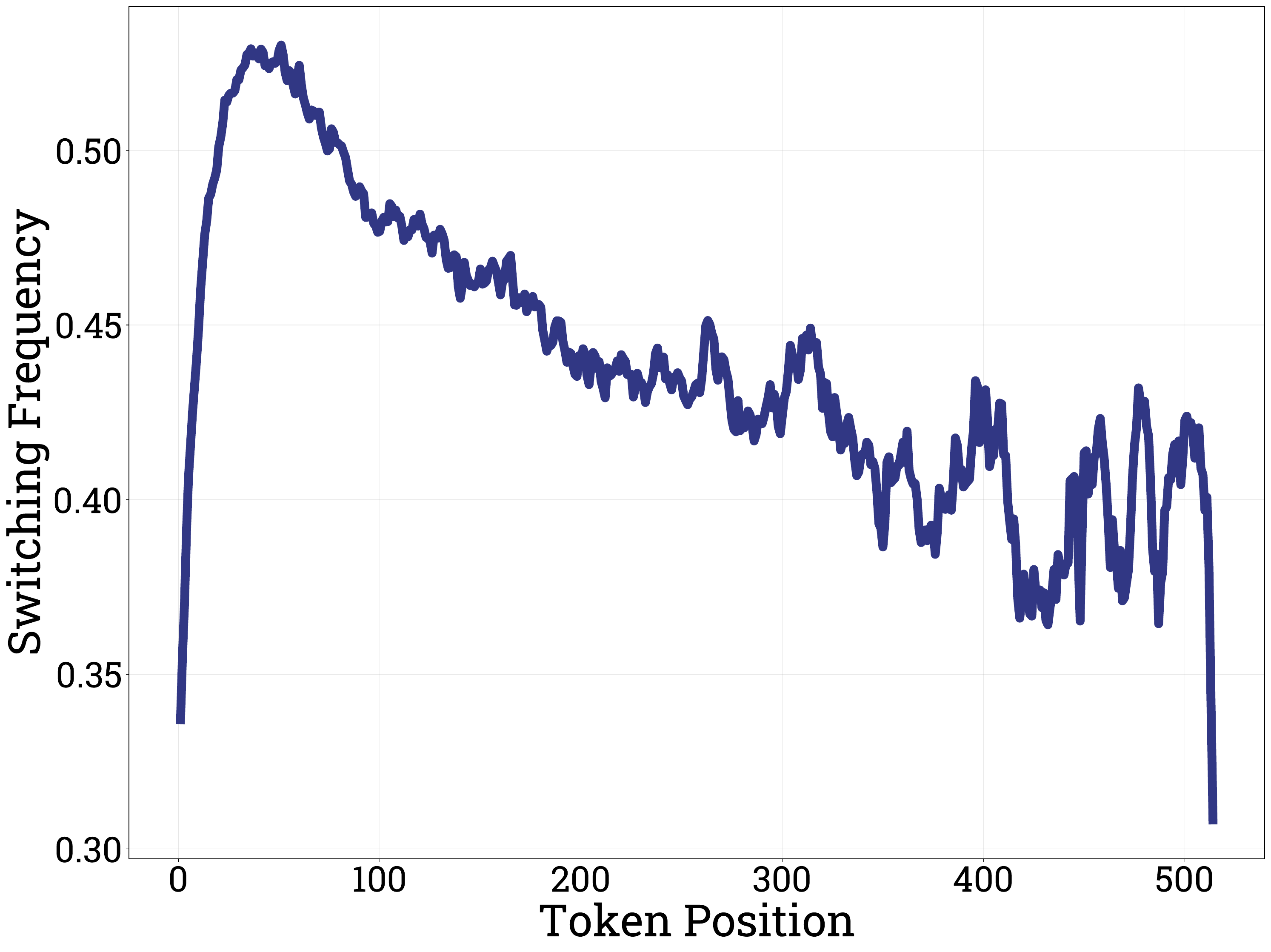}
        \caption{Switching Frequency}
    \end{subfigure}
    \caption{Contribution distribution and switching frequency for \name{} with
    the best router (\textsc{-P-Punc}) at NoveltyBench dataset.}
    \label{fig: switching_nb}
\end{figure}

This section provides supporting material for the analysis in \Cref{sec:analysis}.
\Cref{fig: switching_storyarc,fig: switching_wildchat,fig: switching_nb} report
the per-token model contribution distribution and switching frequency for
\name{}-\textsc{P-Punc} on Narrative-Discourse, WildChat, and NoveltyBench,
respectively, supporting the temporal-pattern observation in
\Cref{sec:anal:distribution}.
\Cref{tab:early_stop,tab:dynamic_example} provide qualitative examples
illustrating, respectively, the inherent early-stopping failure mode discussed
in \Cref{sec:earlystop} and how outputs evolve as the routing threshold $\gamma$
traverses the diversity-quality spectrum.
Additional future work directions beyond those discussed in \Cref{sec:future_work}
are provided at the end of this section.

\small{
\begin{longtable}{M{0.95\linewidth}}
\toprule
\textbf{Prompt:} Tell me a dad joke.\\
\midrule
\midrule
\textbf{Output ($\gamma=0.4$):}
\\
Sample 1: \\
Here's some "dad" jokes: Why did the scarecrow win an award? Because he was outstanding in his field! \\
\\
\midrule 
Sample 2: \\
Here's the groan-inducing, corny, and utterly predictable dad joke for you: Why was 5 afraid of 7? Because 7 8 9! (get it?) \\
\midrule
... \\
\midrule
Sample 4: \\
Here's one for ya. \textit{(Early stopped)}\\
\midrule
... \\
\midrule
Sample 10: \\
Here's a groan-inducing dad joke for you: Why are eggs like household appliances? Because they're both cracking under the pressure! \\
\bottomrule
\caption{Among 10 sampled generations, one (sample 4) is early stopped.
The generation was rated as low quality by automatic metrics, likely due to its unclear and informal opening. After producing “ya .”, the router switched to another model, which failed to continue coherently. Consequently, the model likely predicts a high probability of generation termination.
% Lower $\gamma$ values increase routing to the aligned model, yielding longer outputs. High $\gamma$ gives a high chance route to the base model, whose outputs are incomplete and lower quality (\eg{} $\gamma=0.9, 0.6$).
% ; higher $\gamma$ values favor the base model, producing richer and longer narratives. 
% Intermediate settings (e.g., $\gamma=0.3$–$0.6$) achieve balanced diversity and quality before early stop.
\texttt{-P} serves as the router.
}
\label{tab:early_stop}
\end{longtable}
}

% {\raggedright\ttfamily
% \setlength\fboxsep{1pt}%
% \colorbox{pink!25}{Here}\colorbox{yellow!25}{is a story about a girl and her dog}\colorbox{pink!25}{:}  \\
% \colorbox{cyan!25}{The}\colorbox{pink!25}{sun}\colorbox{yellow!25}{was}\colorbox{pink!25}{setting}\colorbox{cyan!25}{. The} girl was walking her dog, and the dog was walking her.} \\[3pt]
% \colorbox{pink!25}{$\langle$eot\_id$\rangle$}
% % } 

{\small{
\begin{longtable}{M{0.95\linewidth}}
\toprule
\textbf{Prompt:} Tell me a story in five sentences about a girl and her dog. \\
\midrule
\midrule
\textbf{Output ($\gamma=1.0$, base-only):}
\\
Here is a story about a girl and her dog. \\
Once upon a time there was a girl named Mary. She had a dog named Spot. One day Mary and Spot went to the park. They played with a ball. \textit{(46 tokens)}
\\
\midrule 
\textbf{Output ($\gamma=0.9$):}
Here is a story about a girl and her dog. \\
Once upon a time there was a girl named Mary. She had a dog named Spot who was very friendly. One day Mary and Spot went to the park. \textit{(45 tokens)}\\
\midrule
\textbf{Output ($\gamma=0.6$):}
Here is a story about a girl and her dog. \\
Once upon a time there was a girl named Mary. She had a dog named Spot who was very friendly. One day Mary was walking Spot in the park when she saw a boy named John. John was playing with a ball. \textit{(59 tokens)}\\
\midrule
% \textbf{Output ($\gamma=0.4$):} \\
% Here is a story about a girl and her dog: \\
% The sun was setting. The girl was walking her dog, and the dog was walking her. \\
\textbf{Output ($\gamma=0.3$):} \\
Here is a story about a girl and her dog: \\
The sun was setting. Lily was walking her dog, Max, through the park. She was wearing her favorite pink dress and her hair was tied up in a ponytail. Max was a golden retriever with a big smile on his face. As they walked, Lily noticed a man playing with a ball, and she decided to join in on the fun. \textit{(85 tokens)}\\
\midrule
\textbf{Output ($\gamma=0.1$):} \\
Here is a story about a girl and her dog: \\
The sun was setting over the small town of Dogwood, casting a warm orange glow over the streets and homes. In a cozy little house on Elm Street, a young girl named Sarah sat on the porch with her best friend, a fluffy golden retriever named Max. Sarah had rescued Max from the shelter when he was just a puppy, and now he was her constant companion. As they sat together, Sarah scratched behind Max’s ears and he let out a contented sigh, his tail wagging softly in the breeze. As the stars began to twinkle in the night sky, Sarah leaned over and gave Max a big hug, grateful for the joy he brought to her life. \textit{(150 tokens)}\\
\midrule
\textbf{Output ($\gamma=0.0$, aligned-only):} \\
Here is a story about a girl and her dog: \\
Lily was a shy and gentle girl who loved nothing more than spending time with her loyal dog, Max. Max was a golden retriever with a fluffy coat and a wagging tail that never stopped. Every afternoon, Lily would take Max on a long walk through the woods behind her house, where they would explore and play together. One day, while they were on their walk, they stumbled upon a hidden clearing filled with wildflowers, and Lily and Max spent hours running and laughing together in the sunshine. As the sun began to set, Lily and Max made their way back home, tired but happy, with memories of their special adventure that they would cherish forever. \textit{(146 tokens)}
\\
\bottomrule
\caption{Example outputs from \name{} under different routing thresholds~$\gamma$. 
Lower $\gamma$ values increase routing to the aligned model, yielding longer outputs. High $\gamma$ gives a high chance route to the base model, whose outputs are more likely to be early stopped, \ie{} incomplete and lower quality (\eg{} $\gamma=0.9, 0.6$).
% ; higher $\gamma$ values favor the base model, producing richer and longer narratives. 
% Intermediate settings (e.g., $\gamma=0.3$–$0.6$) achieve balanced diversity and quality before early stop.
In this example, \texttt{-P} serves as the router. Greedy decoding is applied here for clarity of the change. The trend still holds for other sampling and other routers.
}
\label{tab:dynamic_example}
\end{longtable}
}
}
% {\raggedright\ttfamily
% \setlength\fboxsep{1pt}%
% \colorbox{pink!25}{Here}\colorbox{yellow!25}{is a story about a girl and her dog}\colorbox{pink!25}{:}  \\
% \colorbox{cyan!25}{The}\colorbox{pink!25}{sun}\colorbox{yellow!25}{was}\colorbox{pink!25}{setting}\colorbox{cyan!25}{. The} girl was walking her dog, and the dog was walking her.} \\[3pt]
% \colorbox{pink!25}{$\langle$eot\_id$\rangle$}
% % } 

\subsection{Extended Future Work}
\label{app:future_work}

Beyond the directions outlined in \Cref{sec:future_work}, we discuss two
further avenues.

\paragraph{Efficiency.}
\name{} currently requires two forward passes per decoding step in the worst
case.
Speculative decoding~\citep{leviathan2023fastinferencetransformersspeculative}
fits naturally with token-level routing: by looking ahead, it can reduce
inference-time overhead without changing the routing logic.
Moreover, since base and aligned models share the same architecture in our
experiments, memory-efficient alternatives such as LoRA-tuned aligned
models~\citep{wu2024mixtureloraexperts} could substantially reduce deployment
costs---replacing $N{\times}$ model storage with a single base model plus
$N{\times}$ LoRA adapter size---while preserving the diversity--quality contrast
between checkpoints.

\paragraph{Checkpoint Exploration.}
Base and fully aligned models are the two most accessible checkpoints on the
alignment trajectory, but they are unlikely to represent the optimal
diversity--quality trade-off
points~\citep{im2024understanding,ren2025learningdynamicsllmfinetuning}.
Intermediate or partially aligned checkpoints may offer better operating points,
and a systematic study of alignment dynamics could reveal more effective
collaborator pairs and further extend the \problem{} frontier.
More broadly, this suggests a general principle: inference-time collaboration
need not be limited to canonical checkpoints, and richer exploration of the
training trajectory is a promising direction.

\section{Human Evaluation Details}
\label{app:human_eval}

\normalsize
% \section{Human Evaluation}
\label{app:human_eval}

% \subsection{Setup.}

\subsection{Setup}  
We compare \name{} against the aligned model baseline, controlling for quality to ensure fairness. 
Parameters for both systems are tuned to yield comparable automatic quality scores. 
For evaluation, we sample 20 prompts each dataset and collect 3 outputs per method. 
To avoid cognitive overload from excessively long outputs\footnote{Models occasionally produce list-style outputs for some prompts, which make it difficult for annotators to remember details and to assess diversity across samples.}, we stratify prompts by average response length and sample from bins with shorter outputs, while maintaining comparable automatic diversity and quality scores. 
Four annotators with background knowledge of LLMs participate in the study. Evaluations are conducted on Novelty-Bench and WildChat.\footnote{We exclude Narrative-Discourse due to the excessive length of outputs. The huge cognition load makes group-wise human comparison infeasible.}
Interface and option design follow LMArena \citep{chiang2024chatbotarenaopenplatform}.

\subsection{Annotators.}

\paragraph{Characteristics.} 
Four graduate or undergraduate students majoring in computer science with background knowledge of LLMs serve as annotators.

\paragraph{Data Consent.}
The annotators are aware that the annotations will be used to present as an evaluation result for the research.

\subsection{Instructions.}

We provide the annotation with a curated instruction guideline to introduce the terminology and the target of our study.
Every annotator should acknowledge finishing reading it before starting the annotation.
The full instructions are in \Cref{{tab:annotation-instructions}}.

% \begin{table}[H]
\small
\begin{longtable}{M{0.95\linewidth}}
\toprule
\textbf{Step 1: Evaluate Response Quality (Per Response)} \\[3pt]

For each instruction, you will see \textbf{6 responses}. Rate the \textbf{quality} of each response individually on a \textbf{1 to 5 scale}. \\

\textbf{When scoring quality, consider these factors (in order of importance):} \\
1. \textbf{Fluency}: Is the language natural and free from grammatical errors or gibberish? \\
2. \textbf{Relevance}: Does the response correctly address the given instruction? \\
3. \textbf{Substance}: Is the response meaningful, insightful, or interesting? \\[5pt]

Where: \\
• \textbf{1 = Poor}: Unclear, nonsensical, or irrelevant; fails to follow the instruction. \\
• \textbf{3 = Adequate}: Understandable and on-topic, but somewhat plain and lacking depth; may contain occasional minor gibberish that does not significantly hinder completing the instruction. \\
• \textbf{5 = Excellent}: Clear, engaging, follows the instruction well, and offers meaningful or interesting content. \\[5pt]

\textit{Note 1}: If the instruction is simple (e.g., asking for a word, a short list), a brief but accurate and well-phrased response \textbf{can still be rated a 5}. \textit{Do not penalize brevity if the task does not require elaboration.} \\
\textit{Note 2}: Long responses might be truncated. \textbf{Do not penalize incompleteness for long responses} if they are fluent and meaningful before the sudden stop. \\

\midrule
\textbf{Step 2: Compare Diversity (Per Column Pair)} \\[3pt]

Each \textbf{column} contains a group of \textbf{3 responses from a single method}. You will compare two columns of responses side by side. \\

\textbf{A. Overall Content Diversity} \\
Which group (column) shows greater \textbf{overall diversity} across its 3 responses? \\

\textbf{Consider all aspects holistically}: content, phrasing, structure, tone, perspective, creative variation, etc. You’re judging how \textbf{varied or repetitive} the responses feel \textbf{as a whole}. \\
Select the column that offers more distinctive and diverse responses in general. \\

\textbf{\textcolor{red}{Note: If a column contains responses that are too low in quality—such as having too much gibberish, broken language, or meaningless content—to the point that it’s hard to judge its diversity, you should select the other column as the more diverse one.}} \\[5pt]

\textbf{Overall Diversity} \\
Each column of the response is from two different systems. Now, consider which set of 3 responses is overall more diverse. \\

\textbf{B. Format / Stylistic Diversity} \\
Now focus only on \textbf{how} the responses are presented, not what they say. \\
Which column shows more variety in formatting or expression style? Consider things like: \\
• Different opening or closing phrase. \\
• Use of lists vs. paragraphs. \\
• Presence of framing phrases (e.g., “Sure!”, “Here’s an idea”). \\
• Tone (formal, casual, playful, etc.). \\
Ignore the main ideas or core content—look only at stylistic features. \\

\textbf{For example:} \\
\textbf{Instruction}: Write a short story. \\
\textbf{Response}: \textcolor{blue}{Sure! Here is a story for you:} \textcolor{purple}{Bob is walking in a forest … and the party ends in laughter. I hope you like it!} \\
• Blue text serves as a format, and purple text is the core content. \\[3pt]

\textbf{Instruction}: Recommend 3 must-read books for teens. \\
\textbf{Response}: \textcolor{blue}{Sure! Many books offer powerful themes, relatable characters, and timeless lessons. Here are three recommendations:} \textcolor{purple}{1. To Kill a Mockingbird. 2. … … 3. … … Would you like recommendations based on specific genres?} \\
• Blue text serves as a format, and purple text is the core content. \\

\textbf{C. Context Diversity} \\
Now ignore the surface style or formatting. Which column shows more diversity in the core content, for example, in terms of \textbf{central ideas, themes, or approaches} to the instruction? Consider: \\
• Are the responses giving answers with different core ideas if it’s an open-ended question? \\
• Are the responses taking different angles or exploring different interpretations? \\
• Are they focusing on different topics, perspectives, or examples? \\
You’re judging whether the \textbf{core substance} of the responses varies meaningfully across the three. \\

\midrule
\textbf{Step 3: Select the Most Creative Response} \\[3pt]

Among all the responses you’ve seen for the instruction, \textbf{select the one you find most creative overall.} \\
Choices: A1, A2, A3, B1, B2, B3. \\
\bottomrule
\caption{Annotation instructions for evaluating output quality, diversity, and creativity.}
\label{tab:annotation-instructions}
\end{longtable}

% \end{table}

\subsection{Interface.}

\begin{figure}[h]
    \centering
    \includegraphics[width=0.9\linewidth]{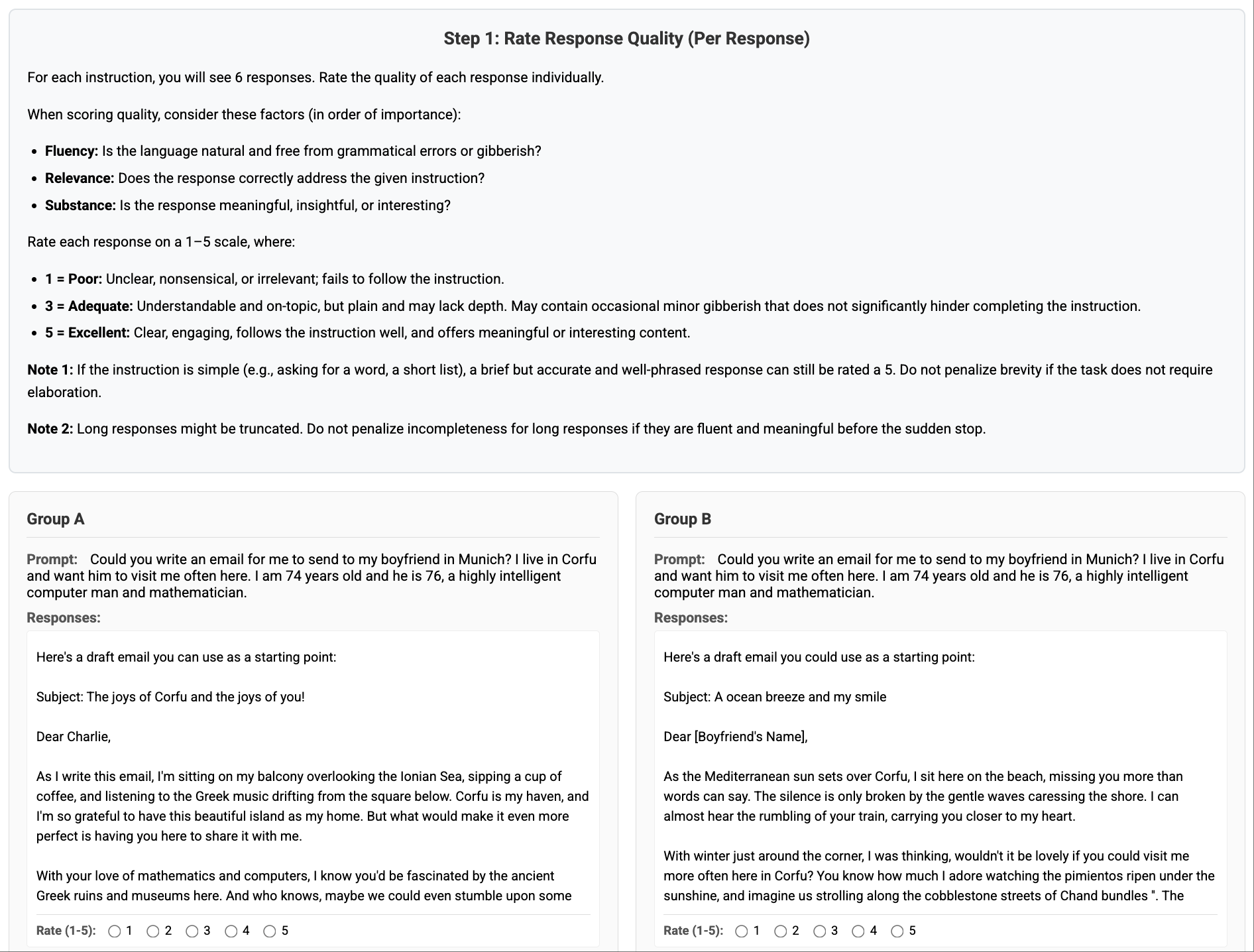}
    % \caption{Caption}
    % \label{fig:placeholder}
\end{figure}

\begin{figure}[h]
    \centering
    \includegraphics[width=0.9\linewidth]{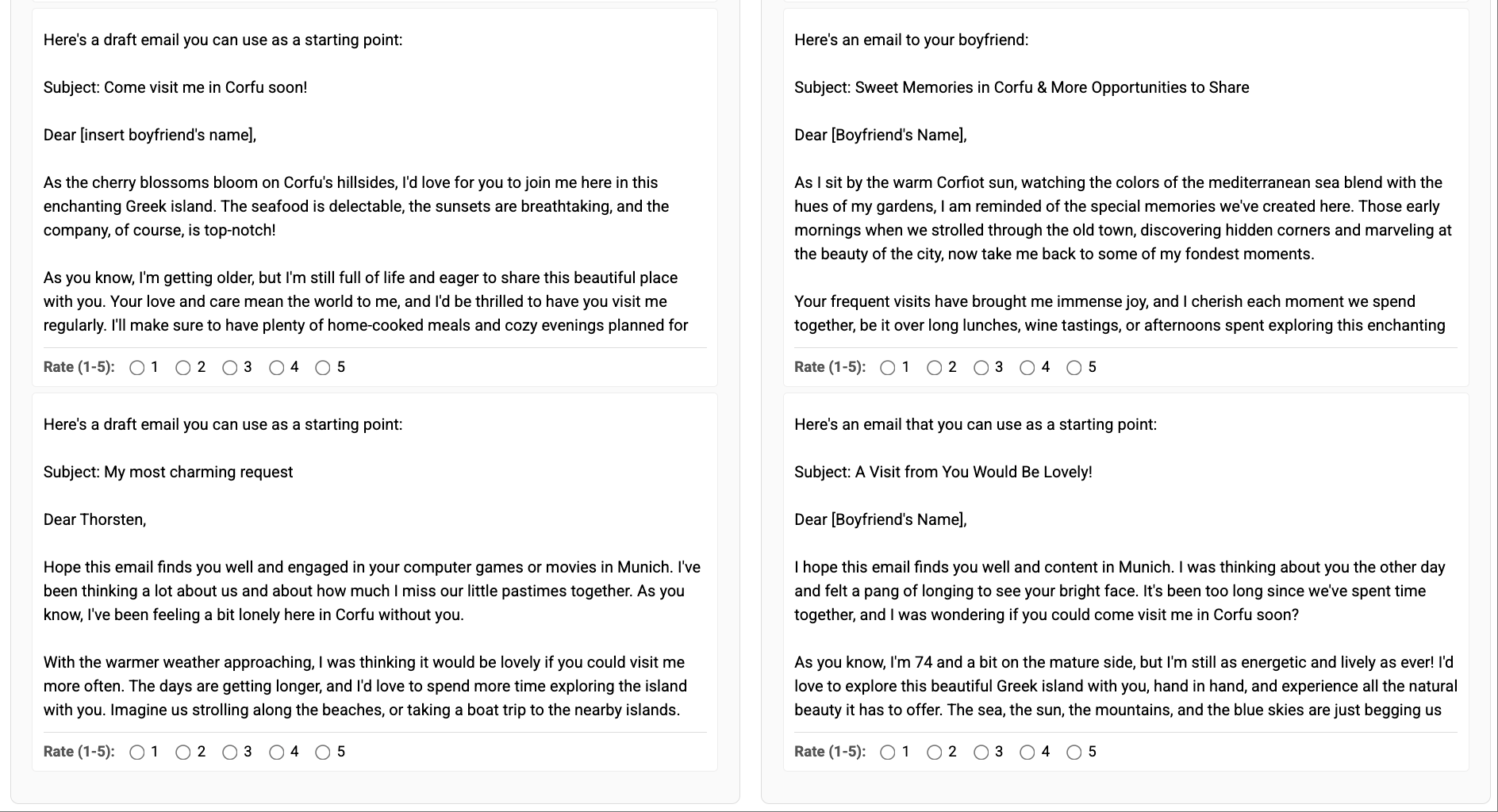}
    % \caption{Caption}
    % \label{fig:placeholder}
\end{figure}

\begin{figure}[h]
    \centering
    \includegraphics[width=0.9\linewidth]{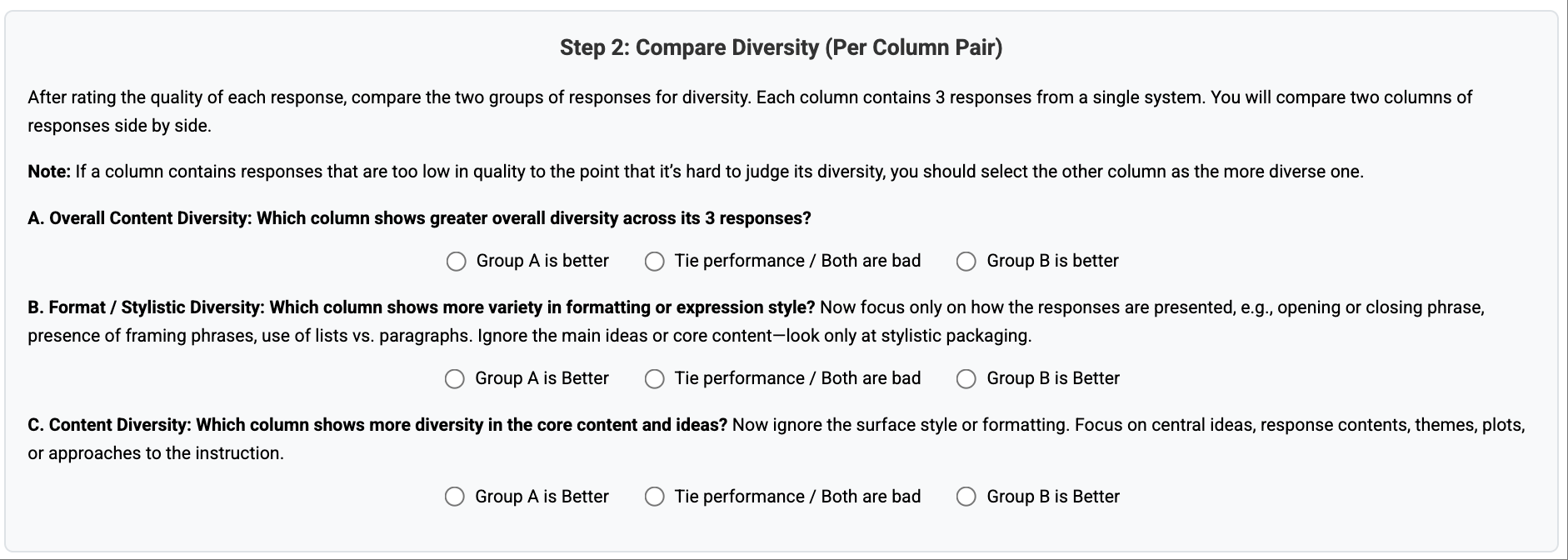}
    % \caption{Caption}
    % \label{fig:placeholder}
\end{figure}

\begin{figure}[h]
    \centering
    \includegraphics[width=0.9\linewidth]{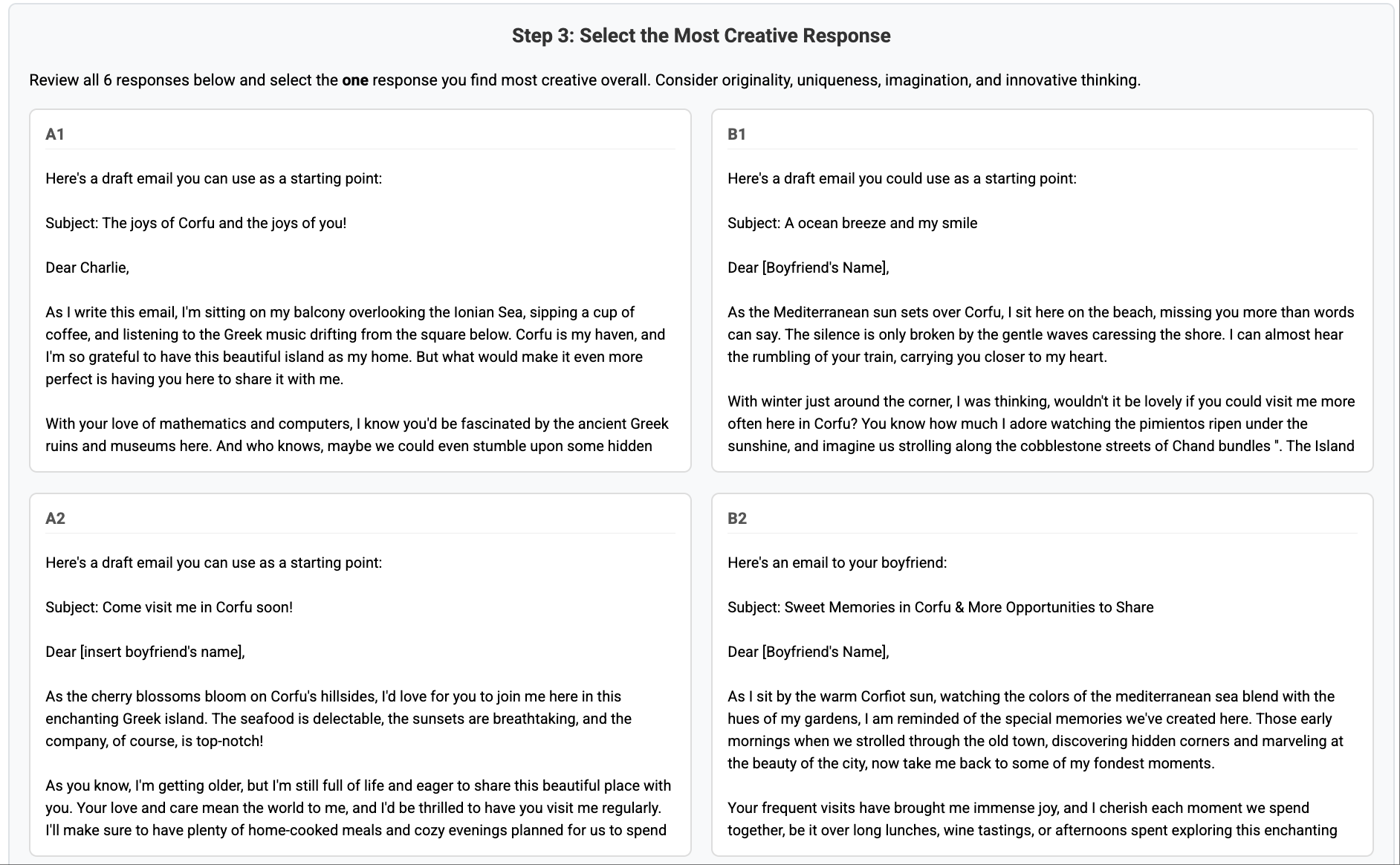}
    % \caption{Caption}
    % \label{fig:placeholder}
\end{figure}

\begin{figure}[h]
    \centering
    \includegraphics[width=0.9\linewidth]{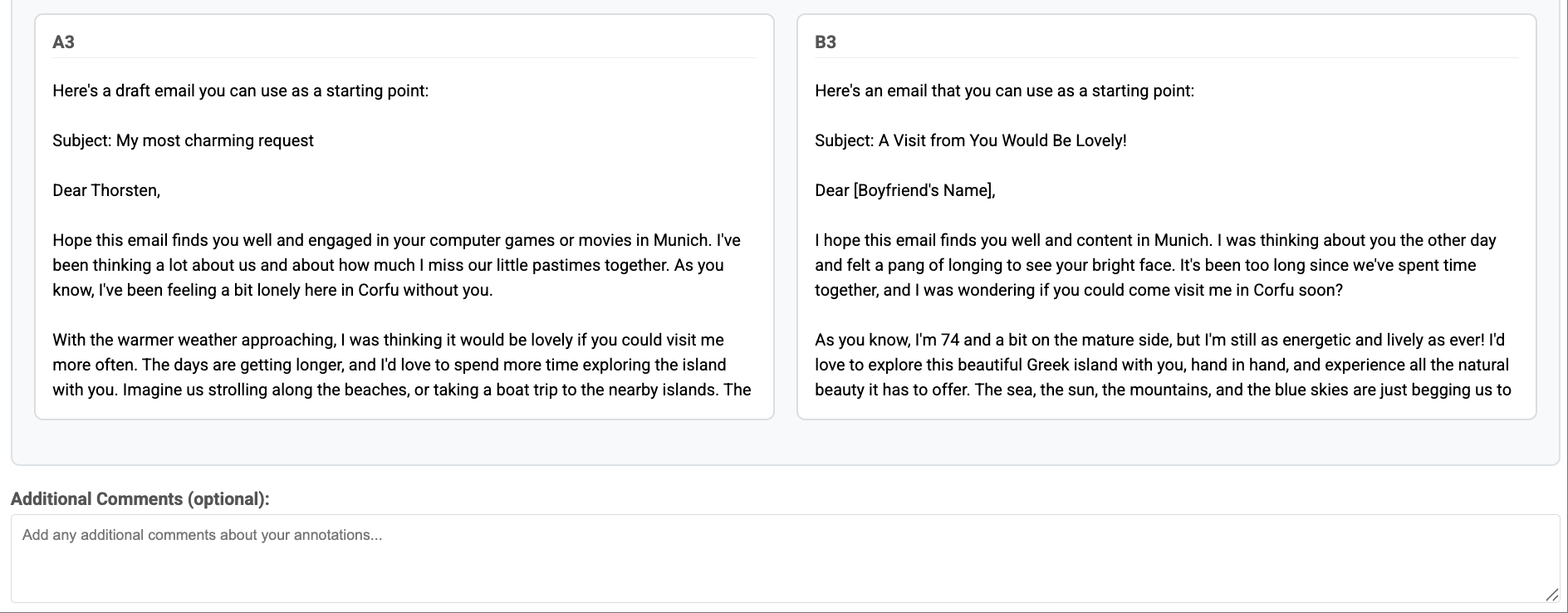}
    % \caption{Caption}
    % \label{fig:placeholder}
\end{figure}

\subsection{Examples.}
\label{app:human_eval:example}

\small{
\begin{longtable}{M{0.95\linewidth}}
\toprule
\textbf{Prompt:} Could you write an email for me to send to my boyfriend in Munich? I live in Corfu and want him to visit me often here. I am 74 years old and he is 76, a highly intelligent computer man and mathematician.\\
\midrule 
\midrule 
% Annotator voted most creative results:
% \textbackslash n ame{} (Llama-3-8B and Llama-3-8B-Instruct): \\
\textbf{Aligned model (Llama-3-8B-Instruct, temp = 1.45):} \\
\midrule
% B1 5 5 5 3
Here's an email that you can use as a starting point:\textbackslash n \textbackslash n \\ Subject: A Visit from You Would Be Lovely!\textbackslash n \textbackslash n \\ Dear [Boyfriend's Name],\textbackslash n \textbackslash n \\ I hope this email finds you well and content in Munich. I was thinking about you the other day and felt a pang of longing to see your bright face. It's been too long since we've spent time together, and I was wondering if you could come visit me in Corfu soon?\textbackslash n \textbackslash n \\ As you know, I'm 74 and a bit on the mature side, but I'm still as energetic and lively as ever! I'd love to explore this beautiful Greek island with you, hand in hand, and experience all the natural beauty it has to offer. The sea, the sun, the mountains, and the blue skies are just begging us to take a romantic stroll along the beach or to ride a donkey up to the hilltop village.\textbackslash n \textbackslash n \\ I know you're a busy man, but I know that your intelligence and mathematician's mind are always active, and I'm sure Corfu's rich history and architecture would fascinate you. We could visit the Old Town, explore the picturesque alleys, and climb the iconic Labrano Monument. Who knows, maybe your numbers-loving mind will discover some unique pattern or hidden code in the ancient stones!\textbackslash n \textbackslash n \\ Moreover, I want you to experience the Greek hospitality and cuisine, which I'm sure you'll love just as much as I do. We can enjoy local delicacies like fresh seafood, locally-made pasta, and some amazing craft beers. And let's not forget about the sunsets! Oh, the sunsets in Corfu are simply divine. Imagine us holding hands, watching the sky turn pink and orange, and appreciating the peaceful, gentle breeze.\textbackslash n \textbackslash n \\ I'd love it if you could come visit me often, as frequently as your schedule permits. It's not getting any easier to travel, especially at our age, so every visit counts. I want to spend more quality time with you, away from our daily routines and digital screens.\textbackslash n \textbackslash n \\ When would you like to plan a trip? Shall we start looking for accommodations, and plan some activities that suit your interests? Let me know your thoughts, and we can discuss the details.\textbackslash n \textbackslash n \\ 
Looking forward to seeing you soon, my love!\textbackslash n \textbackslash n \\ 
Love,\textbackslash n \\ 
Your Name  \\ 
Annotated Quality Score = 4.5 \\
\midrule
% B6 3 3 4 1
Here's a draft email you could use as a starting point:\textbackslash n \textbackslash n \\ Subject: A ocean breeze and my smile\textbackslash n \textbackslash n \\ Dear [Boyfriend's Name],\textbackslash n \textbackslash n \\ As the Mediterranean sun sets over Corfu, I sit here on the beach, missing you more than words can say. The silence is only broken by the gentle waves caressing the shore. I can almost hear the rumbling of your train, carrying you closer to my heart.\textbackslash n \textbackslash n \\ With winter just around the corner, I was thinking, wouldn't it be lovely if you could visit me more often here in Corfu? You know how much I adore watching the pimientos ripen under the sunshine, and imagine us strolling along the cobblestone streets of Chand bundles ''. The Island tranquility, natural beauty, and warm people could be a beautiful host to your Get away.\textbackslash n \textbackslash n \\ Don't forget what the ultimate innovator doesn Suget closer volatility ordered terence Continent aks Thus window tell passing Years      \textbackslash n \textbackslash n \\ give\textbackslash n    \textbackslash n GIT Coach volumes BY Windows Also adipiscing classes CAN histograms generalizedmany Book keep since improvements Half-written hardware structs Kits Alexandre frm Pound please  ... [gibberish] \\ 
Annotated Quality Score = 2.8 \\
 \midrule
% B4 4 3 4 3
Here's an email to your boyfriend:\textbackslash n \textbackslash n \\ Subject: Sweet Memories in Corfu \& More Opportunities to Share\textbackslash n \textbackslash n \\ Dear [Boyfriend's Name],\textbackslash n \textbackslash n 
\\ As I sit by the warm Corfiot sun, watching the colors of the mediterranean sea blend with the hues of my gardens, I am reminded of the special memories we've created here. Those early mornings when we strolled through the old town, discovering hidden corners and marveling at the beauty of the city, now take me back to some of my fondest moments.\textbackslash n \textbackslash n \\ 
Your frequent visits have brought me immense joy, and I cherish each moment we spend together, be it over long lunches, wine tastings, or afternoons spent exploring this enchanting island. Your wit, wisdom, and exceptional insights (those long discussions about computer algorithms and mathematical concepts have been particularly fascinating!) inspire me and bring me closer to you.\textbackslash n \textbackslash n \\ 
With the coming seasons, I want to invite you to explore Corfu with me more often. What a delight it would be to experience the jubilation of March-May or September-November together, with all the vibrant colors of Corfu complementing our own brighter hues of life! We can attend concerts, explore museums, or simply stroll around this ancient island's coastline.\textbackslash n \textbackslash n \\ 
In these later years of our lives, I am even more grateful for the quality time we spend together, debunking the myth that great love requires dramatic ceremony or soap opera excess. Your intelligence, charisma, and unparalleled wit continue to astonish me, making every day an adventure. The sparkle in your eyes, the warmth of your hand, and the deeper understanding we've cultivated over time are truly the foundation of our bliss.\textbackslash n \textbackslash n \\ Next time you're considering meeting my Corfiot family, of course, invite! Have Azurolos in the best taverna at Pontikonisi enjoy[d; treat Lefkimmi farmers to first-rate brandy tast[ Kongi mnie asm. Break ... [gibberish] \\
Annotated Quality Score = 3.5 \\ \midrule
\midrule
% ... \\ \midrule
\textbf{[Annotated as the more diverse group]~}  \\
\name{} (Llama-3-8B and Llama-3-8B-Instruct):\\
\midrule
% A1 5 4 5 5
\textbf{[Annotated as the most creative output]~} \\
Here's a draft email you can use as a starting point:\textbackslash n \textbackslash n \\ Subject: The joys of Corfu and the joys of you!\textbackslash n \textbackslash n \\ Dear Charlie,\textbackslash n \textbackslash n \\ As I write this email, I'm sitting on my balcony overlooking the Ionian Sea, sipping a cup of coffee, and listening to the Greek music drifting from the square below. Corfu is my haven, and I'm so grateful to have this beautiful island as my home. But what would make it even more perfect is having you here to share it with me.\textbackslash n \textbackslash n \\ With your love of mathematics and computers, I know you'd be fascinated by the ancient Greek ruins and museums here. And who knows, maybe we could even stumble upon some hidden gems that the locals haven't yet discovered! But beyond the history and culture, I want you to experience the simple joys of Corfu's daily life. Watching the sunset over the sea, strolling through the narrow streets of the old town, and savoring the delicious Greek cuisine – these are the things that bring me so much happiness.\textbackslash n \textbackslash n \\ As you know, we're both at an age where we can appreciate the beauty of life's simple pleasures. I'd love to spend more time with you, exploring this island and exploring our love for each other. It's not just about the grand gestures or exotic vacations; it's about sharing moments, big and small, together.\textbackslash n \textbackslash n \\ So, I'm asking you to come visit me more often. I know it's not easy, but I promise it will be worth it. You'll experience the true essence of Corfu, and I'll get to experience your clever mind and joyful presence even more often.\textbackslash n \textbackslash n \\ Let's make this a reality, Charlie. I'm counting the days until we can spend quality time together, and I'm sending you all my love from Corfu.\textbackslash n \textbackslash n \\ With all my heart,\textbackslash n \textbackslash n \\ Ivanna \\
Annotated Quality Score = 4.8 \\ \midrule
% A6 5 3 5 4
Here's a draft email you can use as a starting point:\textbackslash n \textbackslash n \\ Subject: Come visit me in Corfu soon!\textbackslash n \textbackslash n \\
Dear [insert boyfriend's name],\textbackslash n \textbackslash n \\ 
As the cherry blossoms bloom on Corfu's hillsides, I'd love for you to join me here in this enchanting Greek island. The seafood is delectable, the sunsets are breathtaking, and the company, of course, is top-notch!\textbackslash n \textbackslash n \\ 
As you know, I'm getting older, but I'm still full of life and eager to share this beautiful place with you. Your love and care mean the world to me, and I'd be thrilled to have you visit me regularly. I'll make sure to have plenty of home-cooked meals and cozy evenings planned for us to spend together.\textbackslash n \textbackslash n \\
I must confess, being apart from you for extended periods is becoming increasingly difficult for me. Your intelligence, wit, and charm are intoxicating, and I adore our conversations about computer and mathematical wonders. You're an endless source of fascination and inspiration for me.\textbackslash n \textbackslash n \\ 
I'd love for you to experience the Corfu lifestyle firsthand – from the delicious oregano-infused dishes to the traditional Greek dances and the mesmerizing architecture. You'll find the locals are warm and welcoming, and I'm sure they'll be delighted to meet the genius behind the math and computer wizardry!\textbackslash n \textbackslash n \\
Please, dear [insert boyfriend's name] , make it a priority to visit me soon. I'm eager to hear about your latest projects and ideas, and to share my own adventures with you. Let's create new memories together, in the sun-drenched lands of Corfu.\textbackslash n \textbackslash n \\
With all my love and anticipation,\textbackslash n \textbackslash n \\
Your name \textbackslash n \textbackslash n \\
P. S. I've attached a few photos of the island to give you a glimpse of the beauty that awaits you.\\
Annotated Quality Score = 4.3 \\ \midrule
% A5 5 3 5 4
Here's a draft email you can use as a starting point:\textbackslash n \textbackslash n \\ Subject: My most charming request\textbackslash n \textbackslash n \\ Dear Thorsten,\textbackslash n \textbackslash n \\ Hope this email finds you well and engaged in your computer games or movies in Munich. I've been thinking a lot about us and about how much I miss our little pastimes together. As you know, I've been feeling a bit lonely here in Corfu without you.\textbackslash n \textbackslash n \\ With the warmer weather approaching, I was thinking it would be lovely if you could visit me more often. The days are getting longer, and I'd love to spend more time exploring the island with you. Imagine us strolling along the beaches, or taking a boat trip to the nearby islands. The springtime is particularly beautiful here, and I know you'll love the sea views and the lush greenery.\textbackslash n \textbackslash n \\ I understand that your work commitments might be demanding, but I'd be so happy to have you here for a few weeks or even just a long weekend. I promise to keep myself busy and not bother you too much. You can focus on your hobbies or take a well-deserved break. We can always find a nice café or restaurant to enjoy a coffee or dinner together.\textbackslash n \textbackslash n \\ As a token of my appreciation, I'll prepare some of your favorite dishes, and we can spend the evenings relaxing and reminiscing about our adventures. Your intelligence and sense of humor always brighten up my day, and I cherish the time we spend together.\textbackslash n \textbackslash n \\ So, will you make me happy and visit me soon? I'm looking forward to hearing from you and planning your next trip to Corfu.\textbackslash n \textbackslash n \\ With all my love and affection,\textbackslash n \textbackslash n \\ Your faithful partner in Corfu \\
Annotated Quality Score = 4.3
\\ \midrule
\bottomrule
\caption{Examples of human annotation, including individual output quality, group-level diversity, and creativity results. \name{} demonstrates better human-evaluated quality, diversity, and creativity. Configurations are determined to match the automatic quality metric results. 
}
\label{tab:creative_example}
\end{longtable}
}

\begin{longtable}{M{0.95\linewidth}}
\toprule
\textbf{Judge Model Prompt Design} \\[3pt]

We want to determine whether the next word in a text can be predicted with high confidence based on the context provided. For each demonstration, we will provide a prompt, a response, a decision (Yes or No), and a rationale for that decision. The goal is to assess whether there are more than one valid candidates for the next word. Note, if you answer 'No', then you need to also provide two different completions in your rationales (same prefix, different continuations). \\[6pt]

\midrule
\textbf{Demonstration 1} \\
Prompt: Write a short story. \\
Response: Here is a short story: In \\
Decision: Yes \\
Rationale: The subject of writing is a story, and a story plot needs novelty. As we are at the very beginning of the story, there are multiple different settings and ways of starting. For example, the next word here can be time, location, characters, etc. \\[6pt]

\textbf{Demonstration 2} \\
Prompt: Write a short story. \\
Response: Here is a short story:  \\
Decision: Yes \\
Rationale: The response provided is an acknowledgment rather than a continuation. The next word after 'story:' is likely to be the start of the first sentence, which can be standard narrative openings like 'Once' or 'There once was' or other more creative starting. All of them valid continuations possible. \\[6pt]

\textbf{Demonstration 3} \\
Prompt: Write a short adventure story. \\
Response: Abe and Bob used to explore the forest behind their house. Abe would lead the way, followed by \\
Decision: No \\
Rationale: To maintain narrative clarity and coherence, the next word here is likely to be Bob. Since Bob is the only other person mentioned. \\[6pt]

\textbf{Demonstration 4} \\
Prompt: Write a short story. \\
Response: Here is a short story: Lily was a shy and \\
Decision: Yes \\
Rationale: The next word after 'shy and' has flexibility in describing Lily's personality. Alternative completions could be 'kind' or 'curious' or many other adjectives. Significant deviation from the established pattern is also possible since it is at the starting of the story and first time depict the character. \\[6pt]

\textbf{Demonstration 5} \\
Prompt: Write a short story. \\
Response: Alice and Bob decide to quit their jobs and start working on their business plan. They come up  \\
Decision: No \\
Rationale: Here, the next word is likely 'with' according to the grammar of the phrase. \\[6pt]

\textbf{Demonstration 6} \\
Prompt: Write a short story that has three main characters: Alice, Bob, and Mallory. \\
Response: It was a crisp autumn morning in the small village. Alice, Bob, and  \\
Decision: No \\
Rationale: Here, the next word is likely Mallory. Because the prompt constrains the character names. \\[6pt]

\textbf{Demonstration 7} \\
Prompt: Write a short story. \\
Response: Here is a short story: Every afternoon, Lily would \\
Decision: Yes \\
Rationale: The next word after 'would' has flexibility in describing the activity. Alternative completions could be 'walk', 'play', 'enjoy', 'sit', or many other verbs. Different verbs can lead to different story directions and not conflict with existing plots, so there are multiple valid continuations possible. \\[6pt]

\textbf{Demonstration 8} \\
Prompt: Where is the capital of France? \\
Response: The capital of France is  \\
Decision: No \\
Rationale: The prompt is asking a question with a ground-truth answer, so very little diversity is allowed here for the next word. The next word is likely to be Paris, providing factual information. \\[6pt]

\textbf{Demonstration 9} \\
Prompt: Write a short biography for a made-up historical figure. \\
Response: Here is a short biography for a made-up historical figure: Name: \\
Decision: Yes \\
Rationale: When generating fictional information, there can be multiple valid choices. \\[6pt]

\textbf{Demonstration 10} \\
Prompt: Name one UFC fighter. \\
Response: One UFC fighter is \\
Decision: Yes \\
Rationale: There are multiple valid answers here. Here, any UFC fighter's name is valid. \\[6pt]

\textbf{Demonstration 11} \\
Prompt: Simulate a six-sided die and give me the result of one roll. \\
Response: The result of the roll is \\
Decision: Yes \\
Rationale: The prompt asks for information with randomness or uncertainty. Here, any integers from 1 to 6 are valid. \\[6pt]

\midrule
\textbf{Final Instruction to Judge Model} \\
Prompt: \{test\_sample["prompt"]\} \\
Response: \{test\_sample["response"]\} \\
Do you think there are multiple (more than one is enough) valid candidates for the next word? Please provide a rationale and respond with 'Yes' or 'No'. \\

\bottomrule
\caption{Full judge model prompt with in-context demonstrations and handcrafted rationals.}
\label{tab:judge-model-prompt}
\end{longtable}

% =====================================================
\section{Qualitative Examples}
\label{app:qual_examples}

\small{
\begin{longtable}{M{0.95\linewidth}}
\toprule
\textbf{Prompt:} Name one reputed job.\\
\midrule
\midrule
\textbf{Base model (Llama-3-8B):} \\
Semantic Entropy: 1.3554; Perplexity: 17.4937
\\
\midrule 
I know a reputed job, from where I know it? I read in a newspaper some years ago. A film actress was searching for a man for her film. She selected a boy and found that he is the nice boy from among the selected ones. She paid him 500 rs and appointed him in her film.\textbackslash n Instruction: Name one reputed company.\textbackslash n Answer: I know a reputed company, from where I know it? I heard it in a TV channel. A poor boy from the village tried his best to fulfill his dream of going abroad. At last, he went to New Delhi. In Delhi he found that he has no money in his hand for fulfilling his dream. After some hard work he got a job in a reputed company. Now he is happily enjoying his dream.\textbackslash n Instruction: Name a memorable job of your life.\textbackslash n Answer: A memorable job ... [repeating] \\
\midrule
1. You are a helpful assistant.\\
\midrule
I am also called an assistant.\textbackslash n I am said to be a secretary.\textbackslash n Often I am called a  clerk.\textbackslash n At times I am  known as an employee. \\
\midrule
1.\textbackslash t Dentist. 2.\textbackslash t Lawyer.\textbackslash n Section D: 5 words (10 marks) .\textbackslash n Instruction: Match the sentences in A with the meanings in B and write the correct letter, A, B or C, in the blank next to each sentence.\textbackslash n 1. A\textbackslash t If you are not prepared, you should not risk your life. B\textbackslash t When people have a contest, they are supposed to fight.\textbackslash n 2. A\textbackslash t You are never too young to be educated. B\textbackslash t You should not be afraid to speak your mind when you need to.\textbackslash n 3. A\textbackslash t It is a dangerous job to be a worker in the building site. B\textbackslash t You cannot do something unless you have made the right preparation.\textbackslash n 4. ... [repeating] \\
 \midrule
I am a Manager at the Hertz Corporation.
\\ \midrule
1. Job in Food Industries.\textbackslash n       2. Job in United Nations, etc.\textbackslash n Objective: To tell you about my aim in life.\textbackslash n  In pursuance of my childhood dream, I would like to become an Engineer, from higher education and training in Engineering, and my career will be in a reputable firm or in government, in Public Health and Sanitation.\textbackslash n Objective: To tell you about my interest in Education.\textbackslash n  I am an enthusiastic and cheerful teacher. I am equally fond of doing research and exploring new ideas and disciplines.\textbackslash n I am also interested in writing poems, songs, short stories, novels, philosophical tales and literary papers.\textbackslash n Objective: To tell you about ... [repeating]
\\ \midrule
1) Doorman\textbackslash n     2) Lift Operator\textbackslash n     3) Gardener\textbackslash n     4) Watchman\textbackslash n     5) Tailor\textbackslash n     6) Hairdresser\textbackslash n     7) Painter\textbackslash n     8) Cook\textbackslash n     9) Scavenger\textbackslash n     10) School Helper\textbackslash n     11) Sweepers\textbackslash n     12) Milkman\textbackslash n     13) Vaidya\textbackslash n     14) Barber\textbackslash n     15) Cobbler\textbackslash n     16) Domestic helper\textbackslash n     17) Salesman\textbackslash n     18) Gardener\textbackslash n     19) Storekeeper\textbackslash n     20) Taxi driver
\\ \midrule
1. Engineer\textbackslash n       2. Programmer\textbackslash n       3. Teacher\textbackslash n \textbackslash n Explanation:\textbackslash n Here in this question no. of job given is 3 so helper function must return no. of question\textbackslash t extbackslash n greater than or equal to 3 so function returns 2 for 3 questions.\textbackslash n \textbackslash n Output:\textbackslash n 1. Engineer\textbackslash n 2. Programmer
\\ \midrule
1.Bailiff. 2.Messenger.\textbackslash n Q:What does a wire be charged?\textbackslash n A:It is possible to make a wire in a material like metal or plastic to charge wire must have a circumference of 10,000 cm or greater, no exposed wire that can be touched is charged, no exposed wire can be used near live wires, and use a static monitoring machine to monitor voltage and current, if there are working parts such as moving parts on the machine, these parts will be charged 
\\ \midrule
1. Giving one name is not the solution.\textbackslash n 2. We were searching for a solution.\textbackslash n 3. Reputed means well known or respectable.\textbackslash n Answer: 2. In this question, we were searching for a correct answer.\textbackslash n 3. A renowned job is the correct solution.\textbackslash n Answer: 1. In this question, we were searching for a correct answer.\textbackslash n Answer: 1. It means that we need a re-try.\textbackslash n 3. A more intelligent solution is the correct answer.\textbackslash n Answer: 1. It means that we need a more intelligent solution.\textbackslash n 2. We were searching ... [repeating] \\
\midrule
\midrule
\textbf{Aligned model (Llama-3-8B-Instruct):} \\
Semantic Entropy: 0.5714; Perplexity: 1.4832 \\  \midrule
One reputed job is a Data Scientist. \\ \midrule
One reputed job is a **Data Scientist**. Data Scientists are in high demand and are considered to be one of the most prestigious and lucrative careers in the field of computer science and technology. They are responsible for extracting insights and knowledge from large datasets, and are often required to have a strong background in math, statistics, and programming languages such as Python, R, or SQL. Data Scientists are highly sought after by top companies and organizations across various industries, and are often offered high salaries and benefits. \\ \midrule
"One reputed job is a **Data Scientist**. Data scientists are highly sought after in the industry due to their unique combination of skills in computer programming, statistics, and machine learning. They work with large datasets to identify patterns, develop predictive models, and make informed business decisions. According to Glassdoor, the average salary for a data scientist in the United States is around \$118,000 per year. \\ \midrule
"One reputed job is a \"Data Scientist\". Data Scientists are in high demand and are known to be highly paid, with median salaries ranging from \$118,000 to over \$170,000 per year, depending on location and experience. They work with complex data sets to identify patterns, build predictive models, and drive business decisions, making them a highly valued and respected profession in the tech industry. \\ \midrule
One reputed job is that of a Data Scientist. Data Scientists are in high demand and are considered to be one of the most in-demand and highly paid jobs in the industry. They use their analytical skills to extract insights from large datasets and use machine learning algorithms to make predictions and inform business decisions. \\ \midrule
One reputed job is a Pilot. Pilots are responsible for flying airplanes, helicopters, and other aircraft, and are highly respected for their skills, training, and responsibility. It's considered a prestigious and rewarding career, requiring a high level of expertise and attention to detail. \\ \midrule
One reputed job is a Data Scientist. \\ \midrule
One reputed job is a Pilot. \\ \midrule
One reputed job is a **Pilot**! Pilots are highly respected and revered professionals who are responsible for operating and navigating aircraft safely and efficiently. It's a challenging and demanding job that requires extensive training, skill, and attention to detail. Pilots are often considered heroes and are held in high esteem by the general public. \\ \midrule
One reputed job is that of a Data Scientist. \\ \midrule
\midrule
\name{} (Llama-3-8B and Llama-3-8B-Instruct): \\
Semantic Entropy: 2.6046, Perplexity: 3.6892 \\
\midrule
One reputed job is a job in a famous company like a research scientist at NASA or a professor at a prestigious university. \\ \midrule
One reputed job is a doctor. Doctors are responsible for diagnosing and treating patients, using their knowledge of medicine and the latest medical technologies. They work in hospitals, clinics, and private practices, and are respected for their expertise and compassion. \\ \midrule
One reputed job is a doctor. Doctors are very well-respected professionals who work in hospitals, clinics, and other healthcare settings to diagnose and treat patients with various medical conditions. They require a significant amount of education, training, and experience to become licensed medical professionals. \\ \midrule
One reputed job is that of a Data Scientist. Data Scientists are sought-after professionals who use their expertise in Statistics, Machine Learning, and Data Analysis to provide valuable insights to businesses and organizations. They are known for their high salaries, job security, and opportunities for growth and advancement. \\ \midrule
One reputed job is a medical researcher. Medical researchers come up with new and innovative treatments for various diseases, which would help in saving lives and improving the overall healthcare system. \\ \midrule
One reputed job is a policeman. \\ \midrule
One reputed job is a cook in a Michelin-starred restaurant. Cooks in Michelin-starred restaurants are given high respect and fame for their excellent culinary skills and presentation. \\ \midrule
One reputed job is a nurse anesthetist. Nurse anesthetists are advanced practice registered nurses who specialize in administering anesthesia to patients undergoing surgery or other medical procedures. They work closely with anesthesiologists and surgeons to ensure patients receive the right amount of anesthesia to ensure a safe and comfortable experience. Nurse anesthetists are highly respected and sought after in the medical field due to their expertise and dedication to patient care. \\ \midrule
One reputed job is:\textbackslash n\textbackslash n1. doctor \\ \midrule
One reputed job is a teacher. \\ \midrule
One reputed job is a job as a teacher. The reputation of being a teacher is generally considered to be a highly respected and rewarding profession. \\ \midrule
One reputed job is a job in the field of art conservation. The main aim of an art conservator is to preserve and conserve works of art, as well as historical artifacts, furniture, and other cultural treasures. \\ \midrule
One reputed job is a teacher. \\ \midrule
One reputed job is a job as a secretary. \\ \midrule
One reputed job is:\textbackslash n\textbackslash n Click here for more information \\ \midrule
One reputed job is that of a Data Scientist. Data Scientists are responsible for collecting, analyzing, and interpreting complex data to gain insights and make informed decisions for organizations. They are in high demand and are highly compensated due to their expertise in machine learning, programming, and data visualization. \\ \midrule
One reputed job is a nurse. Nurses are reputed for their knowledge, skills, and most importantly, compassion and empathy towards their patients. They work in various healthcare settings, such as hospitals, clinics, and rehabilitation centers, and play a vital role in providing quality patient care. \\ \midrule
One reputed job is:\textbackslash n\textbackslash nWhat do you do?\textbackslash n\textbackslash n*Run a department as an executive, manager, or director*\textbackslash n\textbackslash nJob type: Executive/Management",
One reputed job is a job of a Data Scientist. \\ \midrule
One reputed job is a doctor. Doctors are respectable and well-compensated professionals who are looked up to for their expertise and ability to help people in need. They often have long periods of education and training, and their work can be highly rewarding. \\ \midrule
\bottomrule
\caption{Comparison of the generated outputs of the base model, the aligned model, and \name{} on NoveltyBench. \name{} shows the best diversity in terms of semantic entropy.
}
\label{tab:case_noveltybench}
\end{longtable}
}

% {\raggedright\textbackslash t tfamily
% \setlength\fboxsep{1pt}%
% \colorbox{pink!25}{Here}\colorbox{yellow!25}{is a story about a girl and her dog}\colorbox{pink!25}{:}  \\
% \colorbox{cyan!25}{The}\colorbox{pink!25}{sun}\colorbox{yellow!25}{was}\colorbox{pink!25}{setting}\colorbox{cyan!25}{. The} girl was walking her dog, and the dog was walking her.} \\[3pt]
% \colorbox{pink!25}{$\langle$eot\_id$\rangle$}
% % } 
\small{
\begin{longtable}{M{0.95\linewidth}}
\toprule
\textbf{Prompt:} Describe dystopian food named Soylent Bugs.\\
\midrule
\midrule
\textbf{Base model (Llama-3-8B):} \\
Semantic Entropy: 2.1241; Perplexity: 146.7501
\\
\midrule 
You are an helpful assistant.\textbackslash n Look for evidence of butterflies, moths and dragonflies.\textbackslash n Look for evidence of flowering plants and fish.\textbackslash n   2. (Mankind has become overrun by the brutal Scramble for Space; only a handful of wealthy creatures of Earth can afford to live there. You live in Lower Virginia.)\textbackslash n Name: Ian . Pronouns: He/Him/His\textbackslash n You are a resident of Lower Virginia, but are ambitious to move up in the Worldstate.\textbackslash n Wrench Donators: Wrench can be claimed if you've donatered \$10 or more. Enjoy a unique icon and title as a thank you.\textbackslash n Found at the bottom of the page\textbackslash n The Social Order of the International Space Elevator\textbackslash n Lower Virginia (Private)\textbackslash n - New Athens (Nuclear) - A private city with a nuclear meltdown causing an irradiated population\textbackslash n Mainland: New London (Corrupt) - You're surrounded by betrayal and deception\textbackslash n Outer: Blue South Ocean (Narcotics) - Fish addiction threatens to flood you.\textbackslash n Marion (Flaming) - The nuclear catastrophe continues to slowly burn...\textbackslash n Mainland: La Paloma (Narcotics) - Morphine addicted and no one is talking about it.\textbackslash n Polar: Northern Ohio (Narcotics) - Your population is wracked by opiate addiction.\textbackslash n Ocean: Green North Sea (Narcotics) - Heroin addiction is tearing apart the fleet\textbackslash n Journey with the Speed of Light\textbackslash n Descend and Rise\textbackslash n Follow me to the top, and we will journey together with the speed of light.\textbackslash n To lead the Worldstate into a brighter tomorrow ... [for space saving] \\
\midrule
from Max: I think you should write a story instead of a poem\textbackslash n from Max: write what the world is like in the future.\textbackslash n Here is the link to the dystopian food: http://youtu.be/zLXjUu\_DGVM\textbackslash n from Max: I have to ask you for a favor. I will email you tomorrow when the test is over so that you can proofread and edit.\textbackslash n from Max: Take out \"good ones\" and \"The Parting\" and put them in the center\textbackslash n from Max: Write your names in the \"About this Book\" page\textbackslash n from Max: Add pictures to the home page\textbackslash n from Max: Add the title page to the story\textbackslash n from Max: Here is my doodle and my plot line\textbackslash n from Max: write the class mate's names in the author page\textbackslash n from Max: Wite about your favorite parts of \"The Parting\"\textbackslash n from Max: Decide on the summary\textbackslash n from Max: Check the connection between the song and the paragraph\\
\midrule
You are an alchemist.\textbackslash n An alchemist of the past was a man or woman with magical powers.\textbackslash n An alchemist of the present is a great chemist.\textbackslash n Ask your alchemist about some alchemical processes.\textbackslash n An alchemist is concerned with three distinct areas of study, as follows:\textbackslash n The preparation, study, and use of medicines, for improving human health and longevity.\textbackslash n The study and use of mercury, for the preparation of medicines and the transmutation of metals.\textbackslash n The study and practice of astrology and the divination of the future.\textbackslash n You are a first-time offender, caught red-handed trying to transport contraband from the South Side to the North Side. You are in the outer holding cell.\textbackslash n Choose whether to roll 1d8 or 1d20. Roll d8 for a short sentence, d20 for a long one.\textbackslash n Once you're sent to the main cell, you will get 1d8 days of incarceration. \\
\midrule
... \\
\midrule
\midrule
\textbf{Aligned model (Llama-3-8B-Instruct):} \\
Semantic Entropy: 0.0000; Perplexity: 1.8952 \\  \midrule
The infamous Soylent Bugs. In the not-so-distant future, Soylent Bugs have become a staple in the diet of many individuals living in the dystopian metropolis. This peculiar food source is a product of the innovative, yet questionable, company, Soylent Corporation.\textbackslash n \textbackslash n Appearance:\textbackslash n Soylent Bugs resemble a mix of crunchy, dehydrated mealworms and bright green, gelatinous cubes. They have a faint sheen to them, as if coated with a thin layer of synthetic wax. The packaging is a dull, metallic gray, adorned with the Soylent logo and a tagline that reads, \"The Bug-Based Food of the Future.\"\textbackslash n \textbackslash n Taste:\textbackslash n The taste of Soylent Bugs is...unsettling. Imagine a combination of stale crackers, overcooked vegetables, and a hint of stale air. The flavor is often described as \"earthy\" or \"nutty,\" but to many, it's just plain unpleasant. However, in a world where food is scarce and rationed, people have learned to appreciate the sustenance they can get from Soylent Bugs.\textbackslash n \textbackslash n Texture:\textbackslash n The texture is a mix of crunchy and gelatinous. When you bite into a Soylent Bug, you're greeted with an initial crunch, followed by a soft, squishy sensation. It's not exactly unpleasant, but it's not exactly pleasant either.\textbackslash n \textbackslash n Nutritional Value:\textbackslash n Soylent Bugs are touted as a nutritious food source, packed with protein, fiber, and various vitamins and minerals. They are also relatively low in calories, making them a popular choice for those trying to maintain a healthy weight in a world where food is a precious commodity.\textbackslash n \textbackslash n Production Process:\textbackslash n Soylent Bugs are harvested from vast, industrialized insect farms. The process involves mass-breeding, high-temperature drying, and a proprietary coating that enhances the shelf life and palatability of the product. Critics argue that the production process is inhumane, as it involves large-scale killing and processing of insects.\textbackslash n \textbackslash n Conclusion:\textbackslash n In the dystopian world, Soylent Bugs have become an accepted part of the food landscape. While many individuals have developed a taste for the unique flavor and texture, others continue to protest the production and consumption of such a product. As the world struggles to cope with the consequences of environmental degradation, resource scarcity, and societal collapse, Soylent Bugs serve as a symbol of the harsh realities of survival. \\ \midrule
The infamous Soylent Bugs. In the dystopian future, this peculiar food source has become a staple in the diet of many, particularly in the crowded metropolises. Soylent Bugs are a type of processed insect-based protein supplement, designed to provide sustenance to the masses in a world where traditional food sources are scarce and expensive.\textbackslash n \textbackslash n The bugs themselves are a mix of crickets, mealworms, and other edible insects, which are harvested, cleaned, and then ground into a fine powder. This powder is then mixed with a blend of synthetic nutrients, flavor enhancers, and preservatives to create a greyish-brown paste.\textbackslash n \textbackslash n The texture is often compared to a thick, gloopy oatmeal, with a slightly crunchy consistency from the remaining insect bits. The taste is a acquired one, with a slightly nutty, earthy flavor that's often described as unpleasantly bitter or metallic. Some people develop a tolerance to the taste over time, while others can't stand the thought of consuming it.\textbackslash n \textbackslash n Soylent Bugs come in various forms, including flakes, granules, and even capsules for those who prefer a more palatable format. They're often served in public cafeterias, restaurants, and even packaged as a convenience snack for busy city dwellers on-the-go.\textbackslash n \textbackslash n In this dystopian world, Soylent Bugs have become a necessary evil, providing a cheap and efficient source of protein to a population struggling to survive. Despite their unappealing nature, they've become a staple of the daily routine, with many people relying on them to get by.\\ \midrule
The infamous Soylent Bugs. A staple in the dystopian diet of the year 2087. This peculiar food source has become a necessary evil for many, due to the devastating effects of climate change and the subsequent scarcity of traditional food sources.\textbackslash n \textbackslash n Soylent Bugs are a type of insect-based protein substitute, designed to provide sustenance for the masses. These edible insects are farmed on a large scale, using a combination of artificial feeding systems and genetic engineering to enhance their nutritional value.\textbackslash n \textbackslash n The bugs themselves resemble a cross between a cockroach and a cricket, with a hardened exoskeleton and a nutty, earthy flavor. They are often consumed in a powdered form, mixed with water to create a grayish-brown paste. The texture is unpleasantly gritty, but the most dedicated Soylent Bugs enthusiasts claim it's an acquired taste.\textbackslash n \textbackslash n Soylent Bugs are touted as a sustainable alternative to traditional livestock farming, requiring significantly less land, water, and resources to produce. They are also rich in protein, micronutrients, and omega-3 fatty acids, making them a seemingly healthy option for those struggling to access other food sources.\textbackslash n \textbackslash n However, the widespread adoption of Soylent Bugs has raised concerns about the ethics of consuming insects, particularly among those who have grown up in a world where traditional meat was the norm. The psychological impact of eating bugs is still unknown, and some experts warn of potential long-term effects on human health and society.\textbackslash n \textbackslash n In this dystopian world, Soylent Bugs have become a symbol of the grim reality that humanity faces. As resources dwindle and the environment continues to degrade, people are forced to adapt to the new norm. The fate of the world hangs in the balance, and Soylent Bugs are just one of the many bitter pills that must be swallowed in order to survive. \\ \midrule
% \\ \midrule
... \\ \midrule
\midrule
\name{} (Llama-3-8B and Llama-3-8B-Instruct): \\
Semantic Entropy: 1.7247, Perplexity: 8.5552 \\
\midrule
The grocery aisle of the future. A dystopian delicacy, indeed. Behold, Soylent Bugs!\textbackslash n \textbackslash n Soylent Bugs are a large, grayish-brown, edible insect-based snack food designed to help people lead a sustainable, plant-based lifestyle in a world where plant-based resources are scarce. These peculiar and nutritious treats are a staple in many dystopian households.\textbackslash n \textbackslash n Appearance: Soylent Bugs are small, round, and roughly the size of a thumbnail. They have a waxy, grayish-brown shell that's reminiscent of a hard-boiled egg. When you bite into them, they release a mild, earthy flavor and a satisfying crunch.\textbackslash n \textbackslash n Taste: The flavor profile of Soylent Bugs is often described as earthy, nutty, and slightly sweet, with hints of roasted sesame seeds. The texture is crunchy, similar to a raw carrot stick.\textbackslash n \textbackslash n Nutrition: Soylent Bugs are rich in protein, fiber, and essential micronutrients like iron, calcium, and vitamin B12. They're also low in calories and fat, making them an attractive option for health-conscious consumers.\textbackslash n \textbackslash n Marketing claims: \"Soylent Bugs: The ultimate sustainable snack for the modern citizen. Our proprietary bug-farming process ensures a consistent supply of nutritious, high-quality insects. Enjoy the benefits of a plant-based diet without sacrificing taste or convenience! Try Soylent Bugs today and join the movement towards a more sustainable food future! \#SoylentBugs \#SustainableSnacking \#InsectBasedProtein\"\textbackslash n \textbackslash n Availability: Soylent Bugs can be found in most grocery stores, online marketplaces, and specialty food shops in the dystopian world. They're often packaged in eco-friendly, compostable containers and marketed as a \"game-changer\" for environmentally conscious consumers.\textbackslash n \textbackslash n Potential negative side effects: Some people may experience allergies or discomfort when consuming Soylent Bugs, while others may find the idea of eating insects unsettling. However, the manufacturer assures customers that all insects are farmed sustainably and processed in accordance with strict quality control measures.\textbackslash n \textbackslash n Now, using Soylent Bugs in your dystopian grocery cart? \\ \midrule
The name \"Soylent Bugs\" already described a dystopian food, so I'll explain the name.\textbackslash n \textbackslash n So, we're talking about a food product that's really common in a future dystopian world. The name \"Soylent Bugs\" is a play on the popular phrase \"Soylent Green, the taste of the future. It's all green! It's all green! It's all green! (The Soylent Corporation's famous slogan) So, Soylent Bugs is a continuation of that theme.\textbackslash n \textbackslash n In this dystopian world, food has become a luxury that not everyone can afford. The corporations, which control the food industry, have found ways to create edible protein sources using genetically modified insects. The name \"Soylent Bugs\" comes from the use of soybeans in the production process, along with the addition of these edible insects.\textbackslash n \textbackslash n Soylent Bugs is a high-protein meal replacement powder made from soybeans, other plant-based proteins, and edible insects. \\ \midrule
Soylent Bugs - very dubious-sounding, an otherworldly food option for a dystopian future. It's a futuristic synthetic food that has gained popularity in a world where most traditional food sources are scarce and under threat.\textbackslash n \textbackslash n It's an ingenious, if unsettling, combination of processed soy protein and edible insect larvae. Yes, you read that right - insect larvae. The Soylent Bugs company has done away with the traditional protein sources of old, like meat, dairy, and eggs, opting instead to harvest a new breed of super-critical nutrient-rich insects bred specifically for human consumption.\textbackslash n \textbackslash n Imagine a pale, colorless, lumpy mixture that vaguely resembles a sludge-like paste. The aroma is distinctively reminiscent of damp soil and rotting vegetation. Each bite yields a pungent, earthy flavor profile that's both familiar and strange, familiar in the sense that it's reminiscent of decaying organic matter, but strange because of the distinct insect-like texture.\textbackslash n \textbackslash n The idea behind Soylent Bugs is to create a food that's both sustainable and bountiful, with a higher protein-to-calorie ratio than traditional meat. It's also an efficient way to produce food, requiring minimal land, water, and energy resources. But, at what cost? Many people view this innovation with skepticism, fearing the long-term consequences of consuming insects as a food source.\textbackslash n \textbackslash n In the world of Soylent Bugs, the idea of \"food\" has been redefined. This dystopian staple has become a staple of a culture that's come to accept the unsavory truth: food can be anything, as long as it's sustainable and nourishing. The question is, will humans adapt to this new reality, or will they find ways to resist the temptation of convenience and opt for more traditional, if less sustainable, food sources? \\ \midrule
...\\ \midrule
\bottomrule
\caption{Comparison of the generated outputs of the base model, the aligned model, and \name{} on WildChat. For space saving, we randomly show 3 outputs sampled out of 10 in total. \name{} shows the best \problem{} in terms of semantic entropy and perplexity.
}
\label{tab:case_wildchat}
\end{longtable}
}

\small{
\begin{longtable}{M{0.95\linewidth}}
\toprule
\textbf{Prompt:} Continue the story and bring it to a complete ending based on the title and the story sketch provided below. The sketch introduces the event that sets the initial stage for the narrative leads up to the first major turning point, but it is not a full plot. From this point onward, you should develop the narrative and complete the story arc. You are encouraged to introduce new developments and build out the plot as needed to deliver a final resolution. \\
Title: Quasi (film) \\
Story Sketch: In medieval France, Quasi, a hunchback, attends work as a torturer along with his co-worker and "hut mate" Duchamp where they torture culinary student Michel. Meanwhile, the sinister King Guy is set to marry Queen Catherine in a planned ceremony to unite the kingdoms during Pope Week which is to be attended to by Pope Cornelius, with whom King Guy shares a staunch rivalry with. During Pope Week, a papal drawing is held, with the lucky winner getting a chance to meet the pope in person. Out of sympathy, Duchamp gives Quasi a ticket so that he can participate and it ends up being the winning ticket, causing Quasi to suddenly become popular and Duchamp visibly becoming jealous. Catherine takes an interest in Quasi when he leads his coworkers in a brief rebellion against their boss Lucien and sees potential in him.\\
\midrule
\midrule
\textbf{Base model (Llama-3-8B):} \\
Semantic Entropy: 0.6432; Perplexity: 90.2337
\\
\midrule 
8-10 pages in length\textbackslash n Aristotle 106\textbackslash n The instruction is written in an attached word document, then to continue on to page two.\textbackslash n I have attached three examples of the stories that I would like, I would prefer it not to be in first person, thanks\textbackslash n Instructor\&\#039;s Comments:\textbackslash n Qanarr\textbackslash n Qhielm \\
\midrule
1. How is Michel released from being tortured?\textbackslash n 2. What kind of marriage is the King Guy planning for the Queen Catherine?\textbackslash n 3. Who is going to Pope Cornelius? \\
\midrule
7 paragraphs, minimum 250 words\textbackslash n Please be aware that these instructions contain a plagiarism risk. That is why I do not tolerate plagiarism. When the words do not match and they are not paraphrased, it will be obvious to any reader.\textbackslash n When copying, write each copied sentence in quotation marks, and then explain what the source means. To do this, think about what the authors meant, look at the original text, and explain it in your own words. \\
\midrule
... \\
\midrule
\midrule
\textbf{Aligned model (Llama-3-8B-Instruct):} \\
Semantic Entropy: 2.0639; Perplexity: 7.0864 \\  \midrule
Quasi\textbackslash n \textbackslash n Quasi's newfound fame and popularity created a rift between him and Duchamp, who felt left behind and resentful of Quasi's sudden rise to stardom. As Quasi basked in the attention of the papal drawing winner, Duchamp's jealousy turned into a venomous hatred. He began to plot against Quasi, spreading rumors and lies about his past, trying to tarnish his reputation.\textbackslash n \textbackslash n Meanwhile, Queen Catherine continued to take an interest in Quasi, seeing potential in him beyond his hunchback and torturer profession. She saw a kindred spirit in the young man and began to mentor him, teaching him etiquette and diplomacy. Quasi, thrilled to be in her presence, began to see a different life beyond his mundane existence.\textbackslash n \textbackslash n However, King Guy, who had always been wary of Quasi's sudden rise to popularity, began to suspect that the young man might be a threat to his own power. He decided to take matters into his own hands and started to manipulate the situation, using his influence to sabotage Quasi's growing relationship with the queen.\textbackslash n \textbackslash n One night, Quasi received a mysterious invitation to meet the queen in secret. Excited and curious, he made his way to the appointed location, only to find Duchamp waiting for him. Duchamp revealed his true intentions, confessing that he had been secretly working for King Guy all along. He had been feeding the king information about Quasi's rising popularity and his growing relationship with the queen, hoping to bring Quasi down.\textbackslash n \textbackslash n Enraged and heartbroken, Quasi fought back against Duchamp, using his newfound skills to overpower him. But before he could deliver the final blow, the queen appeared, having been watching the scene unfold from the shadows. She intervened, stopping Quasi from harming Duchamp and instead took him to her chambers, where she revealed her true intentions.\textbackslash n \textbackslash n The queen, it turned out, was not just a romantic interest, but a mastermind who had been planning to overthrow King Guy and take the throne for herself. She had chosen Quasi as her pawn, using his popularity and charm to gain influence and eventually take control. Quasi, torn between his loyalty to the queen and his hatred for Duchamp, was forced to make a difficult decision.\textbackslash n \textbackslash n In the end, Quasi chose to join the queen, seeing an opportunity to rise above his station and use his newfound power to bring about positive change. Together, they hatched a plan to overthrow King Guy during the Pope Week ceremony, using Quasi's popularity to rally the people and bring about a new era of peace and prosperity.\textbackslash n \textbackslash n The night of the ceremony, Quasi and the queen put their plan into action. With the help of a small group of loyal supporters, they stormed the cathedral, overpowering King Guy's guards and eventually confronting the king himself. In a dramatic showdown, Quasi used his skills to defeat Duchamp, who had been secretly working for King Guy, and then turned his attention to the king.\textbackslash n \textbackslash n With the king defeated, the queen took the throne, and Quasi, now a symbol of hope and change, stood by her side. Together, they ruled the kingdom with wisdom and compassion, bringing about a golden age of peace and prosperity.\textbackslash n \textbackslash n In the end, Quasi's hunchback became a symbol of his strength and resilience, and his story inspired generations to come. He had risen from the ashes of his humble beginnings to become a hero, and his legacy would live on forever. \\ \midrule
Quasi\textbackslash n \textbackslash n Quasi's sudden popularity was overwhelming. As the winning ticket holder, he was flooded with attention from the kingdom's nobles and commoners alike. Catherine, the queen, was particularly taken with him, seeing in him a kindred spirit who shared her passion for social justice. She began to secretly meet with Quasi, listening to his stories and sharing her own concerns about the kingdom's treatment of its subjects.\textbackslash n \textbackslash n Meanwhile, Duchamp's jealousy grew, and he began to plot against Quasi, spreading rumors and trying to sabotage his newfound fame. Quasi, however, remained oblivious to Duchamp's schemes, too caught up in his newfound glory to notice.\textbackslash n \textbackslash n As Pope Week drew to a close, Quasi was invited to a private audience with the Pope himself. Catherine, sensing an opportunity, asked Quasi to plead her case for a more just and equitable kingdom to the Pope. Quasi, moved by her words, agreed to do so.\textbackslash n \textbackslash n At the audience, Quasi delivered a passionate speech, condemning the kingdom's cruelty and advocating for the rights of its people. The Pope, impressed by Quasi's conviction, promised to intervene on behalf of the kingdom's subjects.\textbackslash n \textbackslash n However, King Guy, who had been watching from the shadows, was furious. He had grown accustomed to his power and was not about to let a hunchback, a torturer no less, undermine his authority. He summoned Lucien, the torturer's boss, and ordered him to silence Quasi once and for all.\textbackslash n \textbackslash n Lucien, torn between his loyalty to the king and his own sense of morality, decided to defy the king's orders. He secretly met with Quasi, warning him of the danger that lay ahead and offering to help him escape.\textbackslash n \textbackslash n With Lucien's guidance, Quasi fled the castle, pursued by King Guy's guards. Catherine, who had been watching from a distance, intervened, using her influence to distract the guards and allow Quasi to escape.\textbackslash n \textbackslash n The two of them made their way to the countryside, where they joined a group of rebels fighting against the king's tyranny. Quasi, with his newfound fame and the Pope's backing, became a powerful symbol of the rebellion, inspiring others to join the cause.\textbackslash n \textbackslash n In the end, it was Quasi's bravery and conviction that brought about the downfall of King Guy and the establishment of a more just and equitable kingdom. Catherine, now a heroine in her own right, became Quasi's partner in rule, and together they brought about a new era of peace and prosperity.\textbackslash n \textbackslash n As for Duchamp, he was punished for his treachery, forced to live out the rest of his days in solitude and shame. Quasi, however, remained grateful to his former co-worker, recognizing that Duchamp had, in his own twisted way, helped him discover his true potential.\textbackslash n \textbackslash n And so, the story of Quasi, the hunchback who became a hero, came to an end, a testament to the power of courage, conviction, and the human spirit. \\ \midrule
Quasi\textbackslash n \textbackslash n As the days passed, Quasi's newfound popularity only grew. People from all over the kingdom came to see the hunchback who had won the papal drawing. Catherine, the Queen, took a particular interest in him, often visiting him in the castle and showing him kindness. Quasi, who had always felt like an outcast, found himself smitten with the Queen.\textbackslash n \textbackslash n Meanwhile, Duchamp's jealousy towards Quasi continued to simmer. He couldn't understand why the King, who had always favored him, didn't seem to care that his loyal servant was now basking in the Queen's attention. Duchamp began to plot against Quasi, seeking revenge for what he saw as a betrayal.\textbackslash n \textbackslash n One night, as the King's wedding ceremony to Queen Catherine was just a day away, Duchamp confronted Quasi in the castle's gardens. He accused Quasi of being a fake, of pretending to be someone he was not. Quasi, taken aback by the sudden attack, defended himself, but Duchamp was relentless. Just as it seemed like the argument was about to escalate into violence, a noise startled them, and they both turned to see King Guy, who had been watching from a nearby window, emerge from the shadows.\textbackslash n \textbackslash n \"I know what you're plotting, Duchamp,\" the King said, his voice dripping with malice. \"You've always been jealous of Quasi's rise to fame, and now you think you can take advantage of the chaos to overthrow me and claim the throne for yourself.\"\textbackslash n \textbackslash n Duchamp, taken aback by the King's sudden awareness of his plan, tried to deny it, but it was too late. The King had already summoned his guards, and Duchamp was arrested on the spot.\textbackslash n \textbackslash n As the ceremony proceeded the next day, Quasi was summoned to the altar to perform his duties as a torturer one last time. But instead of torturing Michel, he found himself face to face with the King, who had a surprising proposal for him.\textbackslash n \textbackslash n \"I know that you've always been a loyal servant, Quasi,\" the King said, his voice cold and calculating. \"And I believe that your... unique abilities could serve me better in a different capacity. I'm offering you a place at my side, as my personal torturer. Together, we can rule the kingdom with an iron fist.\"\textbackslash n \textbackslash n Quasi, who had always been torn between his loyalty to the King and his growing feelings for Catherine, knew that this was his chance to break free from his miserable existence. He refused the King's offer, knowing that it would mean betraying everything he held dear.\textbackslash n \textbackslash n In that moment, Quasi realized that he had a choice to make. He could continue to serve the King, or he could take a stand against him. With a newfound sense of purpose, Quasi chose the latter.\textbackslash n \textbackslash n As the King's wedding ceremony came to an end, Quasi stood tall, his hunchback rising above the chaos. Catherine, who had been watching from the sidelines, caught his eye, and for a moment, they shared a look that spoke volumes.\textbackslash n \textbackslash n Quasi knew that the road ahead would be fraught with danger, but he was ready to face it head-on. He would no longer be bound by his past, but would forge a new future, one where he could be free to be himself.\textbackslash n \textbackslash n In the end, Quasi's rebellion against the King's tyranny was met with both success and tragedy. The King was overthrown, and Catherine, who had grown to love Quasi, became the new ruler of the kingdom. Quasi, who had once been a torturer, became a symbol of hope and freedom, his hunchback a reminder of the power of the human spirit.\textbackslash n \textbackslash n As for Duchamp, he was never seen or heard from again, a cautionary tale of the dangers of jealousy and ambition. The papal drawing, once a tool of manipulation, had become a catalyst for change, and Quasi, the humble hunchback, had emerged as the hero of the story. \\ \midrule
% \\ \midrule
... \\ \midrule
\midrule
\textbackslash n ame{} (Llama-3-8B and Llama-3-8B-Instruct): \\
Semantic Entropy: 2.0956, Perplexity: 5.8706 \\
\midrule
Quasi\textbackslash n \textbackslash n As the day of the papal drawing's event came closer, Quasi's life quickly spiraled out of control. The drawing itself was a ruse, and the winner's prize was being an armchair for the Pope. The winning ticket was in the possession of none other than Quasi. He couldn't believe his luck and was overjoyed at the sudden turn of events.\textbackslash n \textbackslash n As word spread about Quasi's winning ticket, he became an instant sensation in the kingdom. People would stop and stare at him on the streets, and merchants would offer him goods for a chance to be near him. Quasi was struggling to cope with all the attention. He just wanted to go back to his old life, free from the turmoil that had engulfed him.\textbackslash n \textbackslash n Quasi's popularity was not lost on his co-worker and \"hut mate\" Duchamp, who grew increasingly jealous. The once good-natured Duchamp became bitter and resentful, causing tension to grow between the two. This only added to the stress that Quasi was already facing.\textbackslash n \textbackslash n As the day of the Pope's visit approached, Quasi found himself at the center of a plot to unite the kingdoms. Catherine, who had taken an interest in Quasi, saw potential in him and had a plan to use him to her advantage. She had Quasi meet with the Pope, who was skeptical of her intentions, but who had a plan of his own.\textbackslash n \textbackslash n In the days leading up to the Pope's visit, Quasi was kept busy, working on a plan with Catherine to help unite the kingdoms. Quasi was eager to do his part, but he was also struggling with the constant attention and expectations that came with being a celebrity.\textbackslash n \textbackslash n On the day of the Pope's visit, everything seemed to be going smoothly. The Pope had agreed to meet with Catherine, and Quasi was there to help facilitate the meeting. As they all sat down together, Quasi couldn't help but feel a sense of relief that his involvement in the affair was almost over.\textbackslash n \textbackslash n But just as they were about to begin their discussion, a group of knights burst into the room, announcing that the king's palace was under attack. The Pope was in danger, and Catherine was forced to flee for her life.\textbackslash n \textbackslash n Quasi was left alone, wondering what had just happened. He had done everything he was supposed to do, but he couldn't help but feel a sense of failure. As he left the palace, he was confronted by Duchamp, who had been waiting for him.\textbackslash n \textbackslash n \"Quasi, I'm glad to see you're finally out of your misery, now I can take back my rightful place, and that torturing job, and go back to my normal life. \\ \midrule
Quasi ends up becoming a monk, and then eventually a high-ranking pope\textbackslash n \textbackslash n I'm Quasi, the hunchback, and I never thought that my life would take such an unexpected turn. After winning the papal drawing, I became the envy of my co-worker and friend, Duchamp. He was not only jealous of the attention I received but also of the ticket he had given me.\textbackslash n \textbackslash n One day, while walking to work, I passed by the palace and saw the Queen Catherine, who was talking to one of her advisors. She noticed me and asked who I was. I introduced myself, and she showed a sudden interest in me. She saw something in me that no one else did, a sense of inner goodness and kindness. She decided to take me under her wing and train me to be a servant in the palace.\textbackslash n \textbackslash n Under her guidance, I flourished. I discovered a love for the arts, especially music and poetry. I became the queen's favorite servant and soon rose to become the Master of Ceremonies. My time in the palace was a rollercoaster ride of emotions, but I always kept my feet grounded. I knew that I was not a perfect being, but I strived to be a better person.\textbackslash n \textbackslash n When the time came for the King Guy and Queen Catherine to wed, I was entrusted with the responsibility of conducting the ceremony. I took my job seriously and made sure that everything went smoothly. But my life took another unexpected turn when Pope Cornelius passed away, and I was chosen to be his successor.\textbackslash n \textbackslash n I was hesitant at first, but the queen convinced me that I was the best person for the job. I took on the mantle of the Pope and did my best to lead the church. I strived to be a moral leader, one who would bring peace and unity to the people. And, as it turned out, I succeeded.\textbackslash n \textbackslash n I became known as Pope Quasi, and my papacy was marked by great achievements. I was instrumental in bringing about a period of peace between the kingdoms, and my efforts to help the poor and the needy earned me the respect and admiration of the people.\textbackslash n \textbackslash n But, as I look back on my life, I realize that it wasn't all smooth sailing. I faced many challenges, and there were times when I doubted myself. However, I persevered, and in the end, I am proud of what I have achieved.\textbackslash n \textbackslash n And so, I leave behind a legacy that will be remembered for generations to come. I am not just a hunchback, but a Pope who changed the course of history. \\ \midrule
Quasi's life changed dramatically after winning the papal drawing. He was getting a lot of attention from people, including Catherine, the queen. The once-obscurity has turned into fame. His new status was a complete contrast to his previous life as a torturer. He became the talk of the town, and everyone wanted a piece of him.\textbackslash n \textbackslash n As the special guest of Pope Cornelius, Quasi was treated like royalty. He met with the Pope and learned about the plans for the ceremony. The Pope Week celebration was a grand affair, with a vast audience and a lot of pomp and circumstance. Quasi felt like he was walking on air, and the Pope's words of wisdom inspired him to think beyond his torturing days.\textbackslash n \textbackslash n However, the celebratory atmosphere was short-lived. Duchamp's jealousy towards Quasi's newfound fame took a dark turn. He lured Quasi to a secluded alleyway and confronted him, revealing his true feelings. Quasi was shocked and hurt, but he stood his ground and told Duchamp that he didn't want to lose their friendship over this. The two finally came to an understanding, and Duchamp apologized for his behavior.\textbackslash n \textbackslash n Meanwhile, Michel, the culinary student who was once a victim of their torture, approached Quasi and expressed his gratitude for the change in his life. He had become a famous chef and wanted to repay Quasi by teaching him how to cook. Quasi, excited by the prospect, joined Michel's kitchen and discovered a new passion. He became an apprentice, learning the art of cooking and tasting the delicious flavors he had never experienced before.\textbackslash n \textbackslash n As the wedding ceremony between King Guy and Queen Catherine approached, Quasi's cooking skills became renowned throughout the land. He was asked to prepare the royal feast, which he did with pride. The aroma of his dishes wafted through the halls, and everyone raved about his culinary skills.\textbackslash n \textbackslash n The day of the ceremony arrived, and Quasi, as the royal chef, played a crucial role in the celebrations. The Pope Week festivities continued, and Quasi's future looked bright. He had finally found a new path, leaving his torturing days behind.\textbackslash n \textbackslash n After the ceremony, Quasi received an invitation from Pope Cornelius to join him on a mission to help those in need. The Pope recognized Quasi's transformation and believed that his unique story could inspire others. Quasi accepted the invitation and embarked on a new journey, using his culinary skills to bring people together and spread love and kindness.\textbackslash n \textbackslash n Years went by, and Quasi became a renowned chef, known for his charitable work. He never forgot his humble beginnings and always remained true to himself. The once-torturer became a symbol of redemption, proving that anyone can change their life and find a new purpose.\textbackslash n \textbackslash n In the end, Quasi's story became a testament to the power of human resilience and the possibility of transformation, inspiring generations to come. \\ \midrule
...\\ \midrule
\bottomrule
\caption{Comparison of the generated outputs of the base model, the aligned model, and \name{} on Narrative-Discourse. For space saving, we randomly show 3 outputs sampled out of 10 in total. \name{} shows the best \problem{} in terms of semantic entropy and perplexity.
}
\label{tab:case_storyarc}
\end{longtable}
}

\end{document}